\documentclass{article}
\pdfoutput=1
    \PassOptionsToPackage{numbers, compress}{natbib}

    \usepackage[final]{neurips_2020}

\usepackage[utf8]{inputenc} 
\usepackage[T1]{fontenc}    
\usepackage{hyperref}       
\usepackage{url}            
\usepackage{booktabs}       
\usepackage{amsfonts}       
\usepackage{nicefrac}       
\usepackage{microtype}      
\usepackage{graphicx}
\usepackage{color}
\usepackage{amsmath,bm}
\usepackage{amsmath}
\usepackage{amsbsy}
\usepackage{amssymb}
\usepackage{dsfont}
\usepackage{mathrsfs}
\usepackage{tcolorbox,multicol}
\usepackage[ruled, vlined, linesnumbered]{algorithm2e}%
\usepackage{threeparttable}
\usepackage{wrapfig}
\usepackage{multirow}
\usepackage{cleveref}
\usepackage{caption}
\usepackage{subcaption}
\newcommand{\rulesep}{\unskip\ \vrule\ }
\usepackage{paralist}
\usepackage{makecell}
\usepackage{pifont}
\usepackage{paralist}
\usepackage{diagbox}
\usepackage{float}

\let\oldemptyset\emptyset
\let\emptyset\varnothing

\newtheorem{definition}{Definition}
\newtheorem{lemma}{Lemma}
\newtheorem{theorem}{Theorem}
\newtheorem{prop}{Proposition}

\def\aorr{\texttt{AoRR}}
\def\sorr{\texttt{SoRR}}
\def\tkml{\texttt{TKML}}

\def\X{\mathcal{X}}

\def\cE{\mathcal{E}}
\def\EX{\mathbb{E}}
\def\cR{\mathcal{R}}
\newcommand{\sgn}{\text{sign}}
\def\mbI{\mathbb{I}}
\def\R{\mathbb{R}}
\def\gl{\lambda}
\def\hgl{\hat{\lambda}}

\usepackage{enumerate}

\newenvironment{proof}{{\bf Proof:}}{\hfill\rule{2mm}{2mm}}
\def\ie{{\em i.e.}}

\def\L{\mathcal{L}}

\title{Learning by Minimizing the Sum of Ranked Range}

\author{%
  Shu Hu\\
  University at Buffalo, SUNY\\
  \texttt{shuhu@buffalo.edu} \\
  \And
  Yiming Ying \\
  University at Albany, SUNY \\
  \texttt{yying@albany.edu} \\
  \AND
  Xin Wang \\
  CuraCloud Corporation \\
  \texttt{xinw@curacloudcorp.com} \\
  \And
  Siwei Lyu \\
  University at Buffalo, SUNY \\
  \texttt{siweilyu@buffalo.edu} \\
}

\begin{document}

\maketitle

\begin{abstract}
In forming learning objectives, one oftentimes needs to aggregate a set of individual values to a single output. Such cases occur in the aggregate loss, which  combines individual losses of a learning model over each training sample, and in the individual loss for multi-label learning, which combines prediction scores over all class labels. In this work, we introduce the \underline{s}um \underline{o}f \underline{r}anked \underline{r}ange (\sorr) as a general approach to form learning objectives. A ranked range is a consecutive sequence of sorted values of a set of real numbers. The minimization of \sorr~is solved with the difference of convex algorithm (DCA). We explore two applications in machine learning of the minimization of the \sorr~framework, namely the \aorr~aggregate loss for binary classification and the \tkml~individual loss for multi-label/multi-class classification. 
Our empirical results highlight the effectiveness of the proposed optimization framework and demonstrate the applicability of proposed losses using synthetic and real datasets.  
\end{abstract}

\section{Introduction}

\begin{wrapfigure}{r}{.4\textwidth}
\includegraphics[width=.4\textwidth]{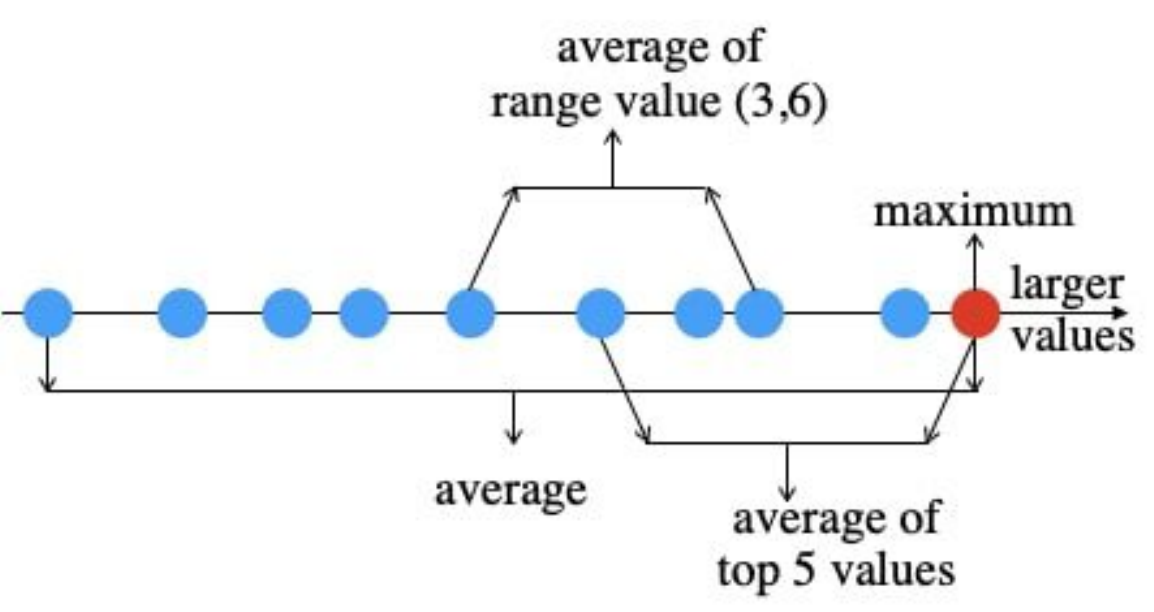}
\caption{\small \em Illustrative examples of different approaches to aggregate individual values to form learning objectives in machine learning. The red dot corresponds to a potential outlier.}
\label{fig:illustrate}
~\vspace{-2em}
\end{wrapfigure}

Learning objective is a fundamental component in any machine learning system. In forming learning objectives, we often need to aggregate a set of individual values to a single numerical value. Such cases occur in the aggregate loss, which combines individual losses of a learning model over each training sample, and in the individual loss for multi-label learning, which combines prediction scores over all class labels. For a set of real numbers representing individual values, the ranking order reflects the most basic relation among them.  Therefore, designing learning objectives can be achieved by choosing operations defined based on the ranking order of the individual values.

Straightforward choices for such operations are the average and the maximum. Both are widely used in forming aggregate losses \cite{vapnik1992principles,shalev2016minimizing} and multi-label losses \cite{madjarov2012extensive}, yet each has its own drawbacks. The average is {\em insensitive} to minority sub-groups while the maximum is {\em sensitive} to outliers, which usually appear as the top individual values. The average top-$k$ loss is introduced as a compromise between the average and the maximum for aggregate loss \cite{fan2017learning} and multi-label individual loss \cite{fan2020groupwise}. However, it dilutes but not exclude the influences of the outliers. The situation is graphically illustrated in Fig.\ref{fig:illustrate}.

In this work, we introduce the \underline{s}um \underline{o}f \underline{r}anked \underline{r}ange (\sorr) as a new form learning objectives that aggregate a set of individual values to a single value.  A ranked range is a consecutive sequence of sorted values of a set of real numbers. The \sorr~can be expressed as the difference between two sums of the top ranked values, which are convex functions themselves. As such, the \sorr~is the difference of two convex functions and its optimization is an instance of the difference-of-convex (DC) programming problems \cite{le2018dc}. The non-smoothness of the \sorr~function and the non-convex nature of the DC programming problem can be efficiently solved with the DC algorithm (DCA).

We explore two applications in machine learning of the minimization of the \sorr~framework. The first is to use the {\em average} of ranked range (\aorr) as an aggregate loss for binary classification. Unlike previous aggregate losses, the \aorr~aggregate loss can completely eliminate the influence of outliers if their proportion in training data is known. Second, we use a special case of \sorr~as a new type of individual loss for multi-label, the \tkml~loss, which explicitly encourages the true labels in the top $k$ range. The new learning objectives are tested and compared experimentally on several synthetic and real datasets\footnote{Code available at \url{https://github.com/discovershu/SoRR}.}. 
The main contributions of this work can be summarized as follows:
\setlength{\parindent}{0em}
\setdefaultleftmargin{0em}{2em}{}{}{}{}
\begin{compactitem}
\item We introduce \sorr~as a general learning objective and show that it can be formulated as the difference of two convex functions, which leads to an efficient solution based on the DC algorithm. 
\item Based on \sorr, we introduce the \aorr~aggregate loss for binary classification, and establish its classification calibration with regards to the optimal Bayes classifier.
\item We also introduce the \tkml~individual loss for multi-label learning, which is another special case of \sorr, and show that the \tkml~loss is a lower bound of the conventional multi-label loss.
\item We empirically demonstrate the robustness and effectiveness of the proposed \aorr, \tkml, and their optimization framework on both synthetic and real datasets.
\end{compactitem} 

\section{Sum of Ranked Range}
\label{sec:dca}

For a set of real numbers $S = \{s_1, \cdots, s_n\}$, we use $s_{[k]}$ to denote the {\em top-$k$ value}, which is the $k$-th largest value after sorting the elements in $S$ (ties can be broken in any consistent way). Correspondingly, we define $\phi_k(S) = \sum_{i=1}^k s_{[i]}$ as the {\em sum of the top-$k$} values of $S$. 
For two integers $k$ and $m$, $1 \le m < k \le n$, the $(m,k)$-ranked range is the set of sorted values $\{s_{[m+1]},\cdots,s_{[k]}\}$. The sum of $(m,k)$-ranked range ($(m,k)$-\sorr) is defined as $\psi_{m,k}(S) =\sum_{i=m+1}^k s_{[i]}$, and the average of $(m,k)$-ranked range ($(m,k)$-\aorr) is ${1 \over k-m}\psi_{m,k}(S)$. It is easy to see that the sum of ranked range (\sorr) is the difference between two sum of top values as, $\psi_{m,k}(S) =\phi_k(S) - \phi_m(S)$. Also, the top-$k$ value corresponds to the $(k-1,k)$-\sorr, as  $\psi_{k-1,k}(S)=s_{[k]}$. Similarly, the median can also be obtained from \aorr, as $\frac{1}{\lceil \frac{n+1}{2} \rceil - \lfloor \frac{n+1}{2}\rfloor +1 }\psi_{\lfloor \frac{n+1}{2}\rfloor-1, \lceil \frac{n+1}{2} \rceil}(S)$.

In machine learning problems, we are interested in the set $S(\theta) = \{s_1(\theta),\cdots,s_n(\theta)\}$ formed from a family of functions where each $s_i(\theta)$ is a convex function of parameter $\theta$. We can use \sorr~to form learning objectives. In particular, we can eliminate the ranking operation and use the equivalent form of \sorr~in the following result. Denote $[a]_+=\max\{0,a\}$ as the hinge function. 
\vspace*{-1mm}
\begin{theorem}\label{theorem:sorr} Suppose $s_i(\theta)$ is convex with respect to $\theta$ for any $i\in [1,n]$, then 
\begin{equation}
\min_{\theta}\psi_{m,k}(S(\theta)) = \min_{\theta}\bigg[ \min_{\lambda\in \mathbb{R}}\Big\{k\lambda+\sum_{i=1}^n[s_i(\theta)-\lambda]_+\Big\}-\min_{\hat{\lambda}\in \mathbb{R}}\Big\{m\hat{\lambda}+\sum_{i=1}^n[s_i(\theta)-\hat{\lambda}]_+\Big\}\bigg].
\label{eq:0}    
\end{equation}
Furthermore, $\hat{\lambda}>\lambda$, when the optimal solution is achieved. 
\end{theorem} 
The proof of Theorem \ref{theorem:sorr} is in the Appendix \ref{proof_theorem_sorr}. Note that $\psi_{m,k}(S(\theta))$ is not a convex function of $\theta$. But its equivalence to the difference between $\phi_k(S(\theta))$ and $\phi_m(S(\theta))$ suggests that $\psi_{m,k}(S(\theta))$ is a difference-of-convex (DC) function, because $\phi_k(S(\theta))$ and $\phi_m(S(\theta))$ are convex functions of $\theta$ in this setting. As such, a natural choice for its optimization is the DC algorithm (DCA) \cite{phan2016dca}. 

To be specific, for a general DC problem formed from two convex functions $g(\theta), h(\theta)$ as $s(\theta)=g(\theta)- h(\theta)$, DCA iteratively search for a critical point of $s(\theta)$ \cite{thi2017stochastic}. At each iteration of DCA, we first form an affine majorization of function $h$ using  its sub-gradient at $\theta^{(t)}$, \ie, $\hat{\theta}^{(t)} \in \partial h(\theta^{(t)})$, and then update $\theta^{(t+1)} \in \arg\!\min_\theta \left \{ g(\theta) -\theta^\top \hat{\theta}^{ (t)} \right \}$. DCA is a descent method without line search, which means the objective function is monotonically decreased at each iteration \cite{tao1997convex}. It does not require the differentiability of $g(\theta)$ and $h(\theta)$ to assure its convergence. Moreover, it is known that DCA converges from an arbitrary initial point and often converges to a global solution \cite{le2018dc}. While a DC problem can be solved based on standard (sub-)gradient descent methods, DCA seems to be more amenable to our task because of  its appealing properties  and the natural DC structure of our objective function. In addition, as shown in  \cite{piot2016difference} with extensive experiments,  DCA empirically outperforms the gradient descent method on various problems.


\begin{wrapfigure}{R}{0.45\textwidth}
\vspace{-0.5cm}
\begin{algorithm}[H]
    \caption{DCA for Minimizing \sorr}\label{Alg0}
    \SetAlgoLined
    \begin{flushleft}
    \textbf{Initialization:} $\theta^{(0)}$, $\lambda^{(0)}$, $\eta_l$, and two hyperparameters $k$ and $m$ \\

    \end{flushleft}
    
    \For{$t=0,1,...$}{
    Compute $\hat{\theta}^{(t)}$ with Eq.(\ref{eq:subgradient_phi_m})

    \For{$l=0,1,...$}{

    Compute $\theta^{(l+1)}$ and $\lambda^{(l+1)}$ with Eq.(\ref{eq:SGD})
    }
    
    Update $\theta^{(t+1)} \leftarrow \theta^{(l+1)}$ 

    }
    
\end{algorithm}
\vspace{-10mm}
\end{wrapfigure}

To use DCA to optimize \sorr, we need to solve the convex sub-optimization problem 
\[
\min_{\theta}\bigg[\min_{\lambda}\Big\{k\lambda+\sum_{i=1}^n[s_i(\theta)-\lambda]_+\Big\}- \theta^T\hat{\theta} \bigg].
\]
This problem can be solved using a stochastic sub-gradient method \cite{bottou2008tradeoffs, rakhlin2011making, srebro2010stochastic}. We first randomly sample $s_{i_l}(\theta^{(l)})$ from the collection of $\{s_i(\theta^{(l)})\}_{i=1}^n$ and then perform the following steps:
\begin{equation}\small
    \begin{aligned}
        &\theta^{(l+1)} \leftarrow \theta^{(l)}-\eta_l\left(\partial s_{i_l}(\theta^{(l)})\cdot \mathbb{I}_{[s_{i_l}(\theta^{(l)})>\lambda^{(l)}]} - \hat{\theta}^{ (t)}\right), \\
      &\lambda^{(l+1)}\leftarrow \lambda^{(l)}- \eta_l\left(k - \mathbb{I}_{[s_{i_l}(\theta^{(l)})>\lambda^{(l)}]}\right)
    \end{aligned}
\label{eq:SGD}
\end{equation}
where $\eta_l$ is the step size. In Eq.\eqref{eq:SGD}, we use the fact that the sub-gradient of $\phi_m(S(\theta))$ is computed, as
\begin{equation}
\hat{\theta} \in \partial \phi_m(S(\theta)) = \sum_{i=1}^n \partial s_i(\theta)\cdot \mathbb{I}_{[s_i(\theta)>s_{[m]}(\theta)]},
\label{eq:subgradient_phi_m} 
\end{equation}
where $\partial s_i(\theta)$ is the gradient or a sub-gradient of convex function $s_i(\theta)$ (Proof can be found in the Appendix \ref{prooftheorem1})\footnote{For large datasets, we can use a stochastic
version of DCA, which is more efficient with a provable convergence to a critical point \cite{thi2019stochastic}.}.
The pseudo-code of minimizing \sorr~is described in Algorithm \ref{Alg0}. 

\section{\aorr~Aggregate Loss} 

\sorr~provides a general framework to aggregate individual values to form learning objective. Here we study in detail of its use as an aggregate loss in supervised learning problems and optimizing it with the DC algorithm. Specifically, we aim to find a parametric function $f_\theta$ with parameter $\theta$ that can predict a target $y$ from the input data or features $x$ using a set of labeled training samples $\{(x_i,y_i)\}_{i=1}^n$.  We assume that the individual loss for a sample $(x,y)$ as $s_i(\theta) = s(f(x;\theta),y) \ge 0$. The learning objective for supervised learning problem is constructed from the aggregate loss $\mathcal{L}(S(\theta))$ that accumulates all individual losses over training samples, $S(\theta)=\{s_i(\theta)\}_{i=1}^n$. Specifically, we define the \aorr~ aggregate loss~as
\[
\L_{aorr}(S(\theta)) = {1 \over k-m} \psi_{m,k}(S(\theta)) = {1 \over k-m} \sum_{i=m+1}^{k} s_{[i]}(\theta).
\]
If we choose the $\ell_2$ individual loss or the hinge individual loss, we get the learning objectives in \cite{ortis2019predicting} and \cite{kanamori2017robustness}, respectively. For $m\ge 1$, we can optimize \aorr~using the DCA as described in Section \ref{sec:dca}.

The \aorr~aggregate loss is related with previous aggregate losses that are widely used to form learning objectives.
\begin{compactitem}
    \item the {\em average loss} \cite{vapnik2013nature}:   $\L_{avg}(S(\theta))=\frac{1}{n}\sum_{i=1}^n s_i(\theta)$;
    \item the {\em maximum loss} \cite{shalev2016minimizing}: $\L_{max}(S(\theta)) = \mbox{max}_{1\leq i \leq n}s_i(\theta)$;
    \item the {\em median loss} \cite{ma2011robust}:  $\L_{med}(S(\theta)) = {1 \over 2}
    \left(s_{\left[\lfloor \frac{n+1}{2}\rfloor\right]}(\theta) + s_{\left[\lceil \frac{n+1}{2} \rceil\right]}(\theta)\right)$;
    \item the {\em average top-$k$ loss} (AT$_k$) \cite{fan2017learning}: $\L_{avt-k}(S(\theta)) = \frac{1}{k} \sum_{i=1}^k s_{[i]}(\theta)$, for $1\leq k \leq n$.
\end{compactitem}
The \aorr~aggregate loss generalizes the average loss ($k=n$ and $m=0$), the maximum loss ($k=1$ and $m=0$), the median loss ($k=\lceil \frac{n+1}{2} \rceil$, $m=\lfloor \frac{n+1}{2}\rfloor-1$) , and the average top-$k$ loss ($m=0$). Interestingly, the average of the bottom-$(n-m)$ loss, $\L_{abt-m}(S(\theta)) = \frac{1}{n-m} \sum_{i=m+1}^n s_{[i]}(\theta)$, which is not widely studied in the literature as a learning objective, is an instance of the \aorr~aggregate loss ($k=n$). {In additional, the robust version of the maximum loss \cite{shalev2016minimizing}, which is a maximum loss on a subset of samples of size at least $n-(k-1)$, where the number of outliers is at most $k-1$, is equivalent to the top-$k$ loss, a special case of the \aorr~ aggregate loss ($m=k-1$).}  

\begin{wrapfigure}{R}{0.4\textwidth}
\vspace{-0mm}
\centering
    \includegraphics[trim=1 1 1 1, clip,keepaspectratio, width=0.4\textwidth]{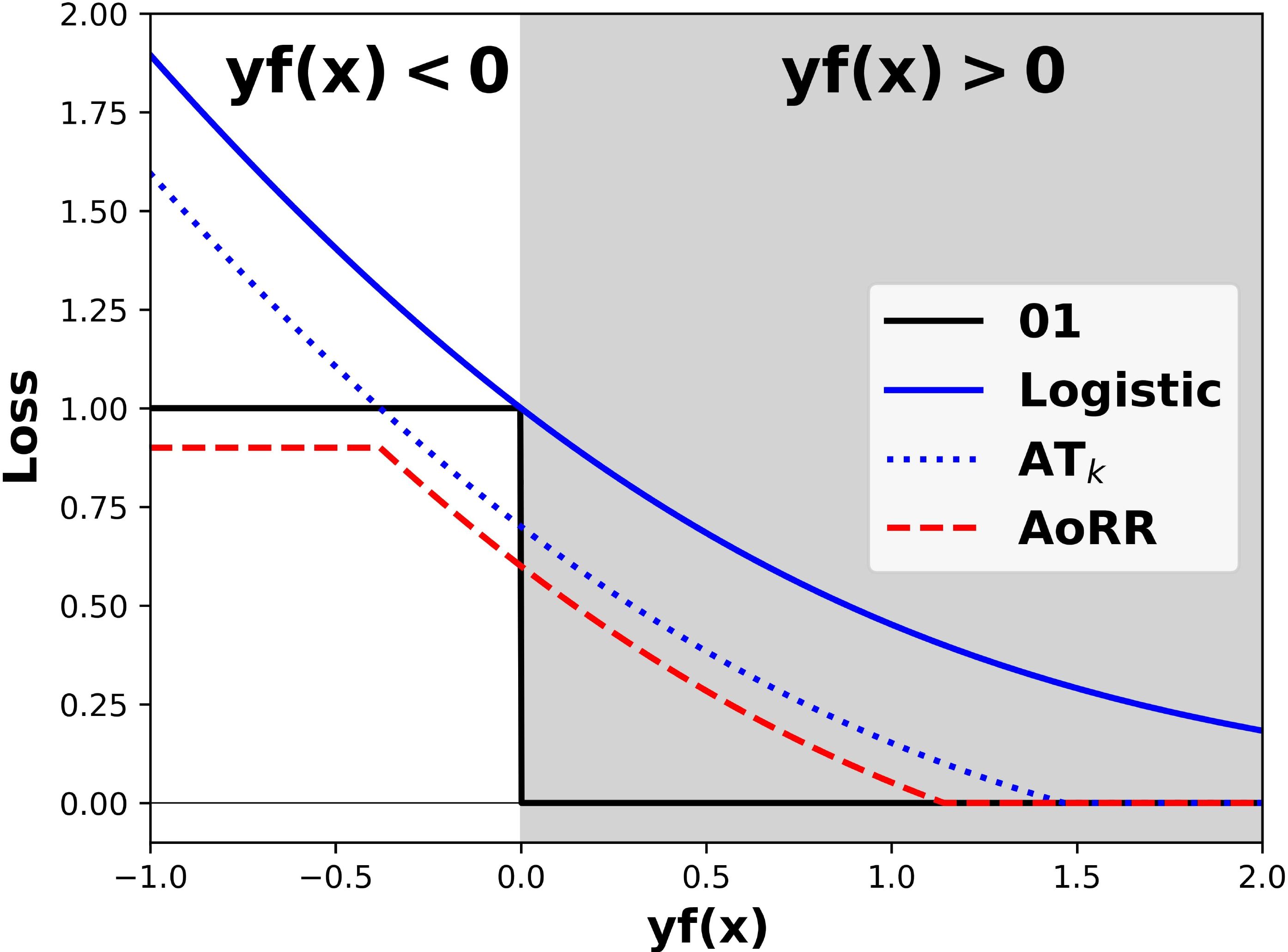}
  \caption{\small \em The \aorr~loss and other losses interpreted at
the individual sample level. The shaded area over 0.00 loss
corresponds to data/target with the correct
classification.}
\vspace{-4mm}
\label{fig:interpretation}
\end{wrapfigure}

Using the \aorr~aggregate loss can bring flexibility in designing learning objectives and alleviate drawbacks of previous aggregate losses. In particular, the average loss, the maximum loss, and the AT$_k$ loss are all influenced by outliers in training data, which correspond to extremely large individual losses. They only differ in the degree of influence, with the maximum loss being the most sensitive to outliers. In comparison, \aorr~loss can completely eliminate the influence of the top individual losses by excluding the top $m$ individual losses from the learning objective.

In addition, {Traditional approaches to handling outliers focus on the design of robust {\em individual losses} over training samples, notable examples include the Huber loss \cite{ friedman2001elements} and the capped hinge loss \cite{nie2017multiclass}. Changing individual losses may not be desirable, as they are usually relevant to the learning problem and application. On the other hand, the AoRR loss introduces robustness to outliers when individual losses are aggregated. The resulting learning algorithm is more flexible.}

The robustness to outliers of the \aorr~loss can be more clearly understood at the individual sample level, with fixed $\gl$ and $\hat{\gl}$. We use binary classification to illustrate with $s_i(\theta) = s(y_if_\theta(x_i))$ where $f_\theta$ is the parametric predictor and $y_i\in \{\pm 1\}.$ In this case,  $y_if_\theta(x_i)>0$ and $y_if_\theta(x_i)<0$ corresponds to the correct and false predictions, respectively. Specifically, noting that $s_i(\theta)\geq 0$, we can rearrange terms in Eq.\eqref{eq:0} to obtain
\begin{equation}
\L_{aorr}(S(\theta)) = {1 \over k-m}\min_{\lambda > 0}\max_{\hat{\lambda} > \lambda} \sum_{i=1}^n \Big\{[s_i(\theta)-\lambda]_+ -[s_i(\theta)-\hat{\lambda}]_+ \Big\}+ k\lambda - m\hat{\lambda}.
\label{eq:aorr-loss}
\end{equation}
We are particularly interested in the term inside the summation in Eq.\eqref{eq:aorr-loss} 
\[
[s(yf_\theta(x))-\lambda]_+ -[s(yf_\theta(x))-\hat{\lambda}]_+ = \left\{
\begin{array}{cc}
     \hat{\lambda}-\lambda & s(yf_\theta(x)) > \hat{\lambda} \\
    s(yf_\theta(x))-\lambda &  \lambda < s(yf_\theta(x)) \le \hat{\lambda} \\
    0 & s(yf_\theta(x)) \le \lambda
\end{array}
\right..
\]
According to this, at the level of individual training samples, the equivalent effect of using the \aorr~loss is to uniformly reduce the individual losses by $\lambda$, but truncate the reduced individual loss at values below zero or above $\hat{\lambda}$. The situation is illustrated in Fig.\ref{fig:interpretation} for the logistic individual  loss $s(yf(x)) = \log_2(1+e^{-yf(x)})$, which is a convex and smooth surrogate to the ideal $01$-loss. The effect of reducing and truncating from below and above has two interesting consequences. First, note that the use of convex and smooth surrogate loss inevitably introduce penalties to samples that are correctly classified but are ``too close'' to the boundary. The reduction of the individual loss alleviate that improper penalty. This property is also shared by the AT$_k$ loss. On the other hand, the ideal $01$-loss exerts the same penalty to all incorrect classified samples regardless of their margin value, while the surrogate has unbounded penalties. This is the exact cause of the sensitivity to outliers of the previous aggregate losses, but the truncation of \aorr~loss is similar to the $01$-loss, and thus is more robust to the outliers. It is worth emphasizing that the above explanation of the \aorr~ loss has been illustrated at the individual sample level with fixed $\gl$ and $\hat{\gl}. $ The aggregate \aorr~loss defined  by \eqref{eq:aorr-loss} as  a whole is not an average sample-based loss because it can not be decomposed into the summation of individual losses over samples.  

\subsection{Classification Calibration} 

A fundamental question in learning theory for classification \cite{bartlett2006convexity,vapnik2013nature} is to investigate when the best possible estimator from a learning objective is consistent with the best possible, \ie, the Bayes rule.  Here we investigate this statistical question for the \aorr~loss by considering its infinite sample case, \ie, $n\to \infty.$ {As mentioned above, the \aorr~loss as a whole is not the average of individual losses over samples, and therefore the analysis for the standard ERM \cite{bartlett2006convexity,lin2004note}  does not apply to our case. }

We assume that the training data $\{(x_i,y_i)\}_{i=1}^n$ are i.i.d. from an unknown distribution $p$ on $\X \times \{\pm 1\}$.  The misclassification error measures the quality of a classifier $f:\X \to \{\pm 1\}$  is denoted by $\cR(f) = \Pr( Y \neq f(X))=  \EX[\mbI_{Yf(X)\le 0}]$. The Bayes error leads to the least expected error, which is defined by $\cR^\ast = \inf_{f} \cR(f) =f_c(x)=\sgn(\eta(x) - {1 \over 2})$ where $\eta(x) = P(Y=1|X=x).$ It is well noted that, in practice, one uses a surrogate loss $\ell: \R \to [0,\infty)$ which is a continuous function and upper-bounds the $01$-loss. Its true risk is given by $\cE_\ell(f) =  \EX[\ell(Yf(X))]$.  Denote the optimal $\ell$-risk by $\cE^\ast_\ell = \inf_f \cE_\ell(f)$, the {\em classification calibration} (point-wise form of Fisher consistency) for loss $\ell$ \cite{bartlett2006convexity,lin2004note} holds true if the minimizer $f^\ast_\ell = \inf_{f}\cE_\ell(f)$ has  the same sign as the Bayes rule $f_c(x)$, \ie, $\sgn(f^\ast_\ell(x)) = \sgn(f_c(x))$ whenever $f_c(x) \neq 0$.

In analogy, we can investigate the classification calibration property of the \aorr~loss. Specifically, we first obtain the population form of the \aorr~loss using the infinite limit of the empirical one given by Eq.\eqref{eq:aorr-loss}. Indeed, 
we know from \cite{bhat2019concentration,brown2007large} that, for any bounded $f$ and $\alpha\in (0,1]$, there holds $ \inf_{\lambda\ge 0}\alpha\lambda+\frac{1}{n}\sum_{i=1}^n[s(y_if_\theta(x_i))-\lambda]_+  \to \inf_{\lambda\ge 0}\alpha\lambda+\EX[s(Yf(X))-\lambda]_+$  as $n\to \infty.$ Consequently, we have the limit case of the \aorr~loss $ \L_{aorr}(S(\theta)) $  restated as follows:
\begin{equation}\small
\begin{aligned}
&\frac{n}{k-m}\bigg[\min_{\lambda}\Big\{\frac{k}{n}\lambda+\frac{1}{n}\sum_{i=1}^n[s(y_if_\theta(x_i))-\lambda]_+\Big\}-\min_{\hat{\lambda}}\Big\{\frac{m}{n}\hat{\lambda}+\frac{1}{n}\sum_{i=1}^n[s(y_if_\theta(x_i))-\hat{\lambda}]_+\Big\}\bigg]\\
&\xrightarrow[n\rightarrow \infty]{\frac{k}{n}\rightarrow \nu, \frac{m}{n} \rightarrow \mu} \frac{n}{k-m}\bigg[\min_{\lambda \ge 0}\Big\{\mathbb{E}[[s(Yf(X))-\lambda]_+]+\nu\lambda\Big\}-\min_{\hat{\lambda} \ge 0}\Big\{\mathbb{E}[[s(Yf(X))-\hat{\lambda}]_+]+\mu\hat{\lambda}\Big\}\bigg].
\end{aligned}
\end{equation}
Throughout the paper, we assume that $\nu>\mu$ which is reasonable as $k>m.$ In particular, we assume that $\mu>0$ since if $\mu=0$ then it will lead to $\hat{\lambda} = \infty$ and this case is reduced to the population version of average top-k case in \cite{fan2017learning}.  As such, the population version of our \aorr~loss \eqref{eq:aorr-loss} is given by 
\begin{equation}
\begin{aligned}
(f^*_0, \lambda^*, \hat{\lambda}^*) = \mbox{arg}\ \underset{f, \lambda\geq 0}{\mbox{inf}} \underset{\hat{\lambda}\geq 0}{\mbox{sup}} \left\{\mathbb{E}[[s(Yf(X))-\lambda]_+ - [s(Yf(X))-\hat{\lambda}]_+]+(\nu\lambda-\mu \hat{\lambda})\right\}.
\end{aligned}
\label{eq:opt_ar_loss}
\end{equation}
It is difficult to directly work on the optima $f^*_0$ since the problem in Eq.\eqref{eq:opt_ar_loss} is a non-convex min-max problem and the standard min-max theorem does not apply here. Instead, we assume the existence of $\gl^*$ and $ \hgl^*$ in \eqref{eq:opt_ar_loss} and work with the minimizer $f^* = \arg\inf_{f} \L(f, \gl^*,\hgl^*)$ where $ \L(f, \gl^*,\hgl^*): = \mathbb{E}[[s(Yf(X))-\lambda^*]_+ - [s(Yf(X))-\hat{\lambda}^*]_+]+(\nu\lambda^*-\mu \hat{\lambda}^*).$ Now we can define the classification calibration for the \aorr~loss. 

\begin{definition} The \aorr~loss is called classification calibrated if there is a minimizer $f^*=\arg\inf_{f} \L(f,\gl^*,\hgl^*)$ such as $f^*(x) >0$ if $\eta(x)>1/2$ and $f^*(x)<0$ if $\eta(x)<1/2.$
\end{definition}
We can then obtain the following theorem. Its proof can be found in the Appendix \ref{appendix:proof_theorem_2}.  
\begin{theorem}
Suppose the individual loss $s: \mathbb{R} \rightarrow \mathbb{R}^+$ is  non-increasing, convex, differentiable at 0 and $s^{\prime}(0)<0.$  If  $0\leq \lambda^*< \hat{\lambda}^*$, then the \aorr~loss is classification calibrated.
\label{theorem:AR_calibration}
\end{theorem}

\subsection{Experiments}

We empirically demonstrate the effectiveness of the \aorr~aggregate loss combined with two types of individual losses for binary classification, namely, the logistic loss and the hinge loss. For simplicity, we consider a linear prediction function $f(x;\theta) = \theta^T x$ with parameter $\theta$, and the $\ell_2$ regularizer $\frac{1}{2C}||\theta||_2^2$ with $C>0$. 

\begin{figure*}[t!]
\captionsetup[subfigure]{justification=centering}
\centering
        \begin{subfigure}[b]{0.246\textwidth}
                \includegraphics[width=\linewidth]{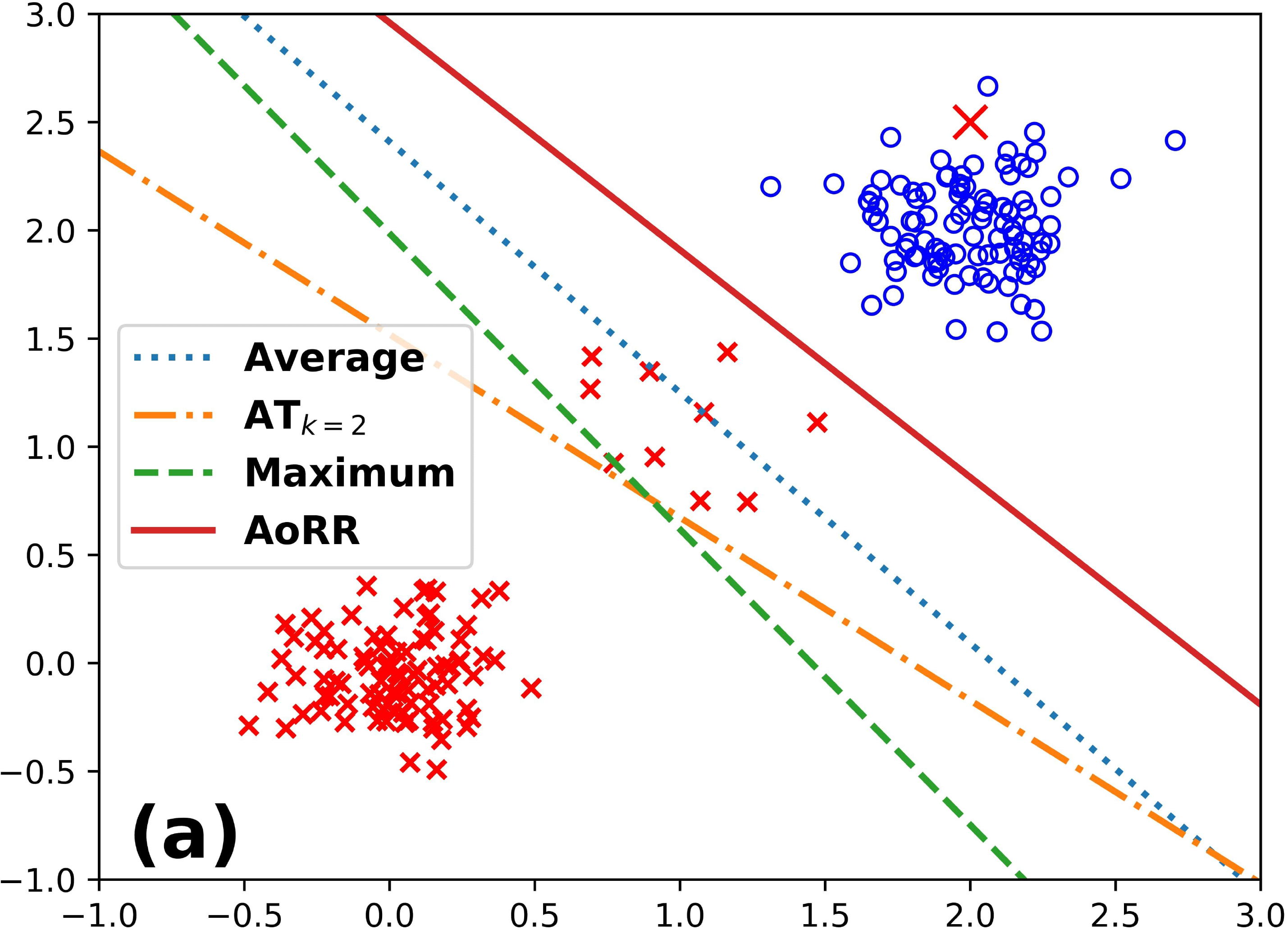}
        \end{subfigure}%
        \begin{subfigure}[b]{0.246\textwidth}
                \includegraphics[width=\linewidth]{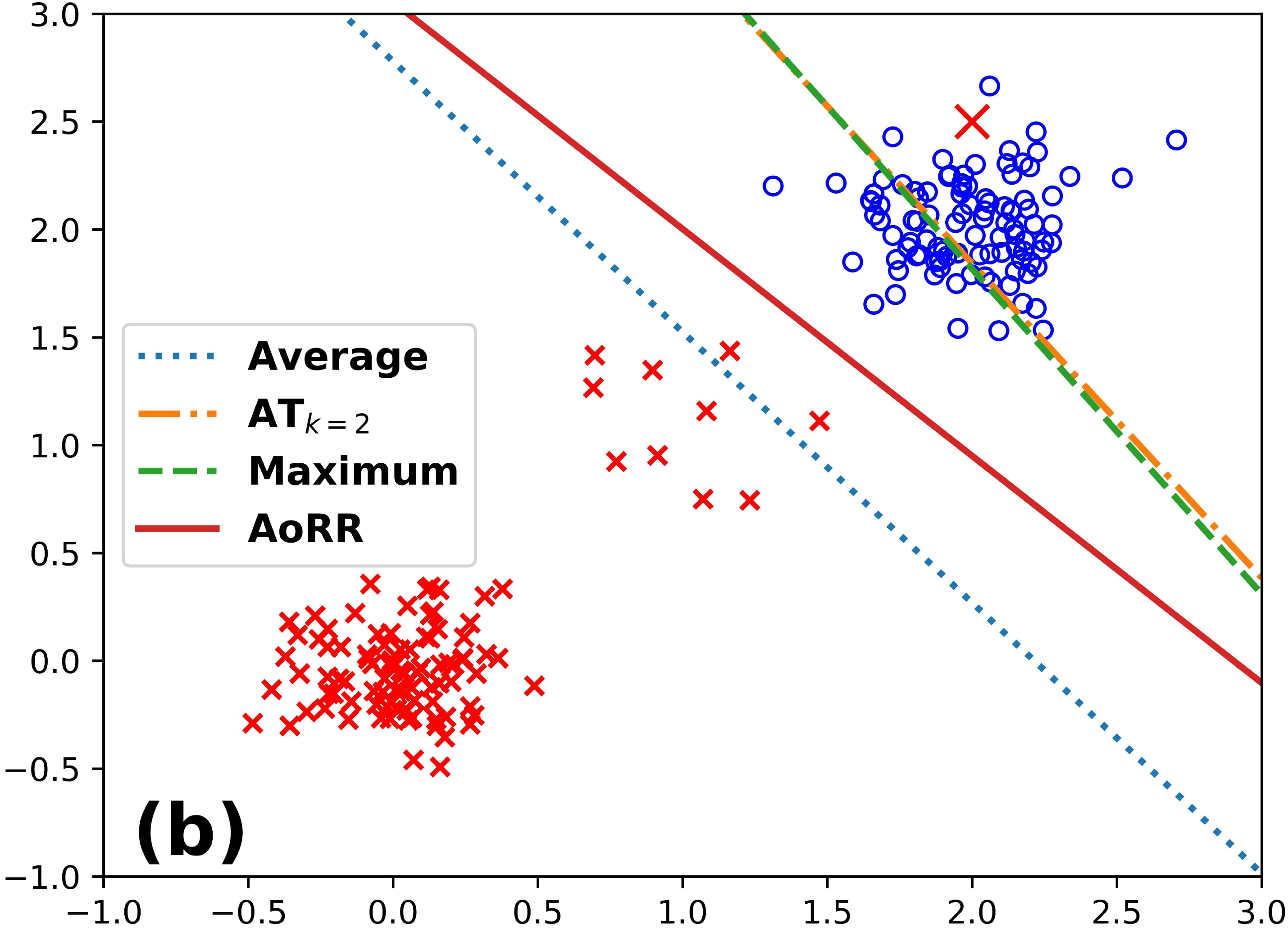}
        \end{subfigure}%
        \rulesep
        \begin{subfigure}[b]{0.246\textwidth}
                \includegraphics[width=\linewidth]{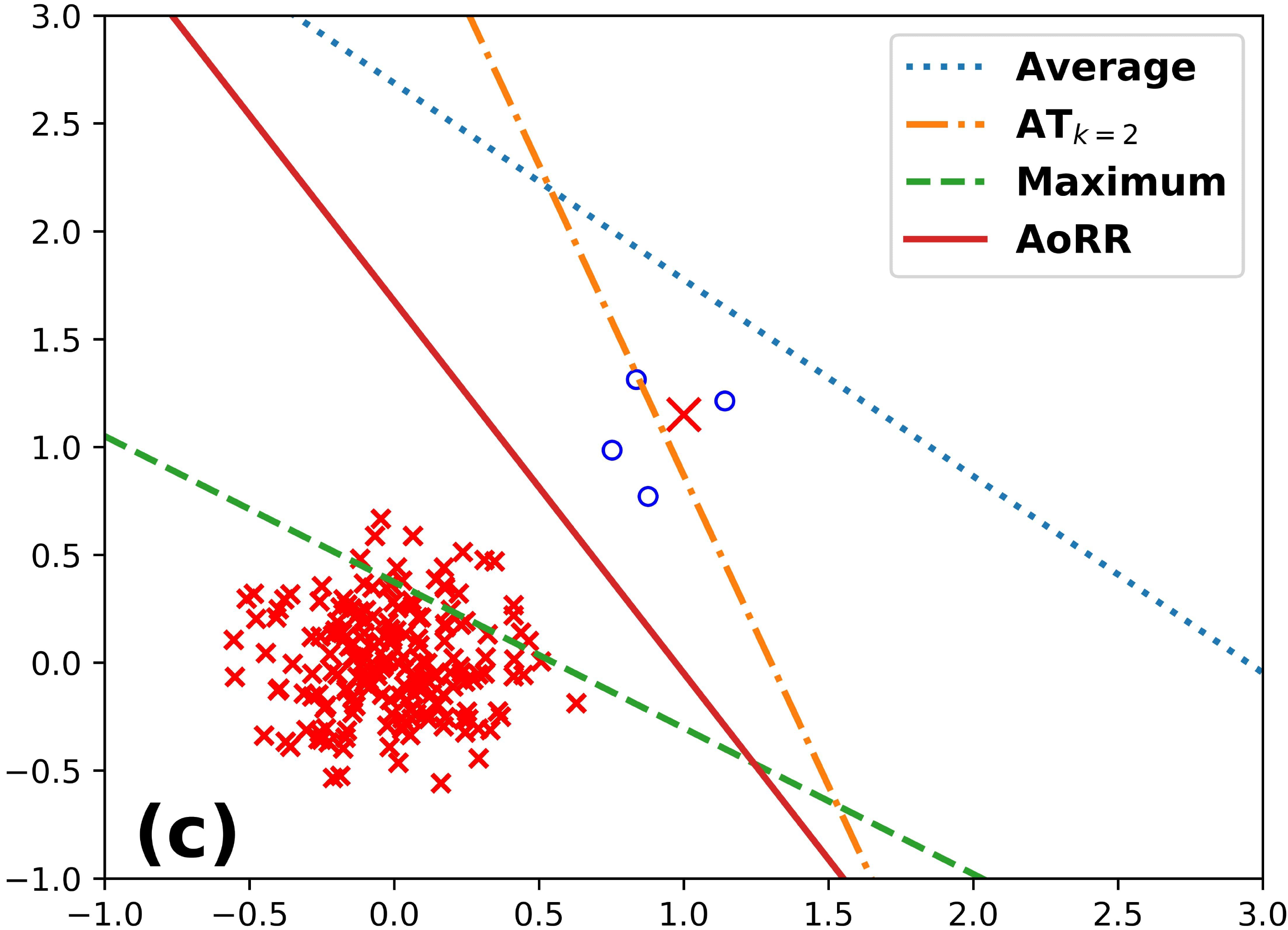}
        \end{subfigure}%
        \begin{subfigure}[b]{0.246\textwidth}
                \includegraphics[width=\linewidth]{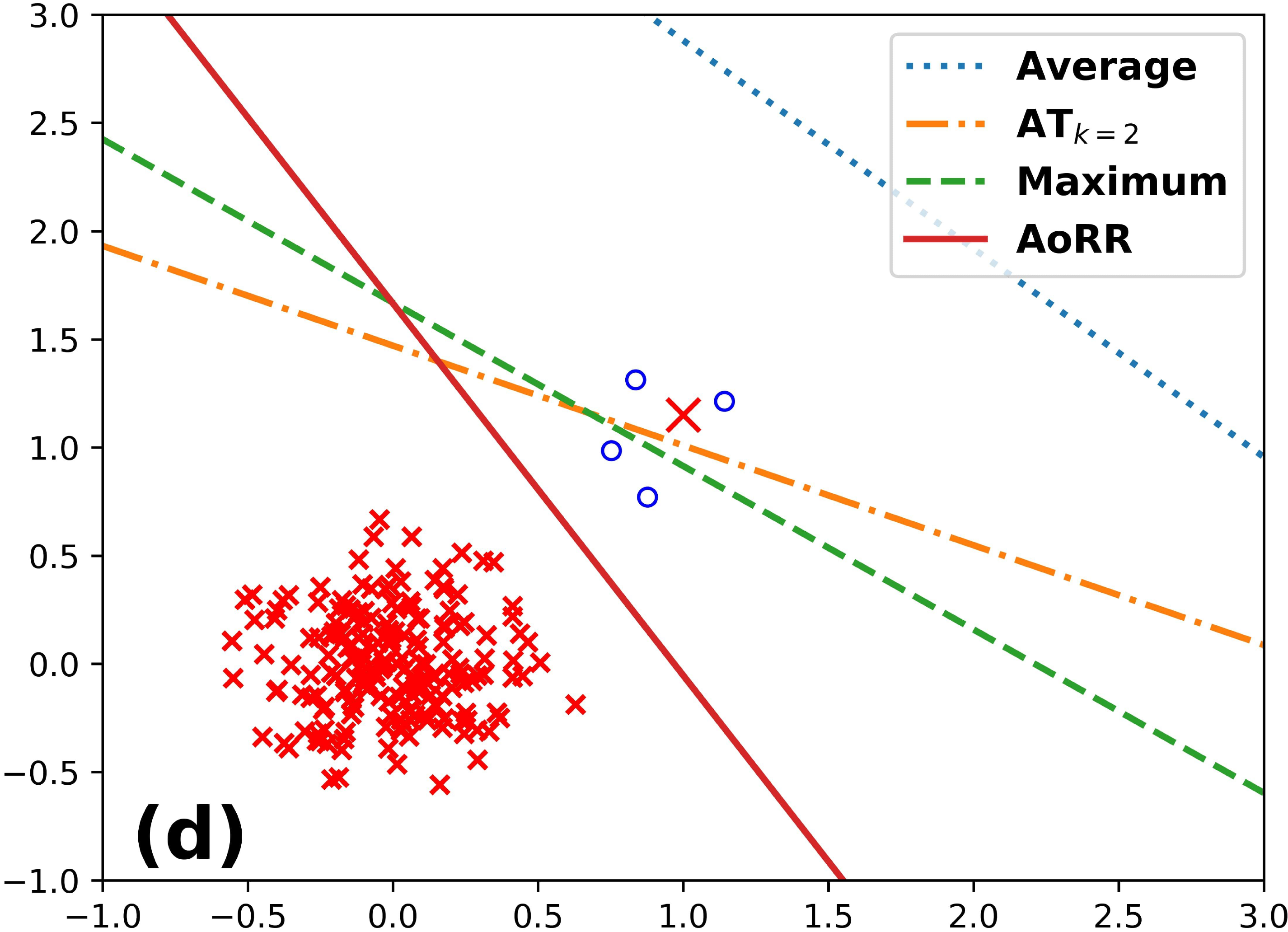}
        \end{subfigure}
        \bigskip
        \begin{subfigure}[b]{0.251\textwidth}            \includegraphics[width=\linewidth]{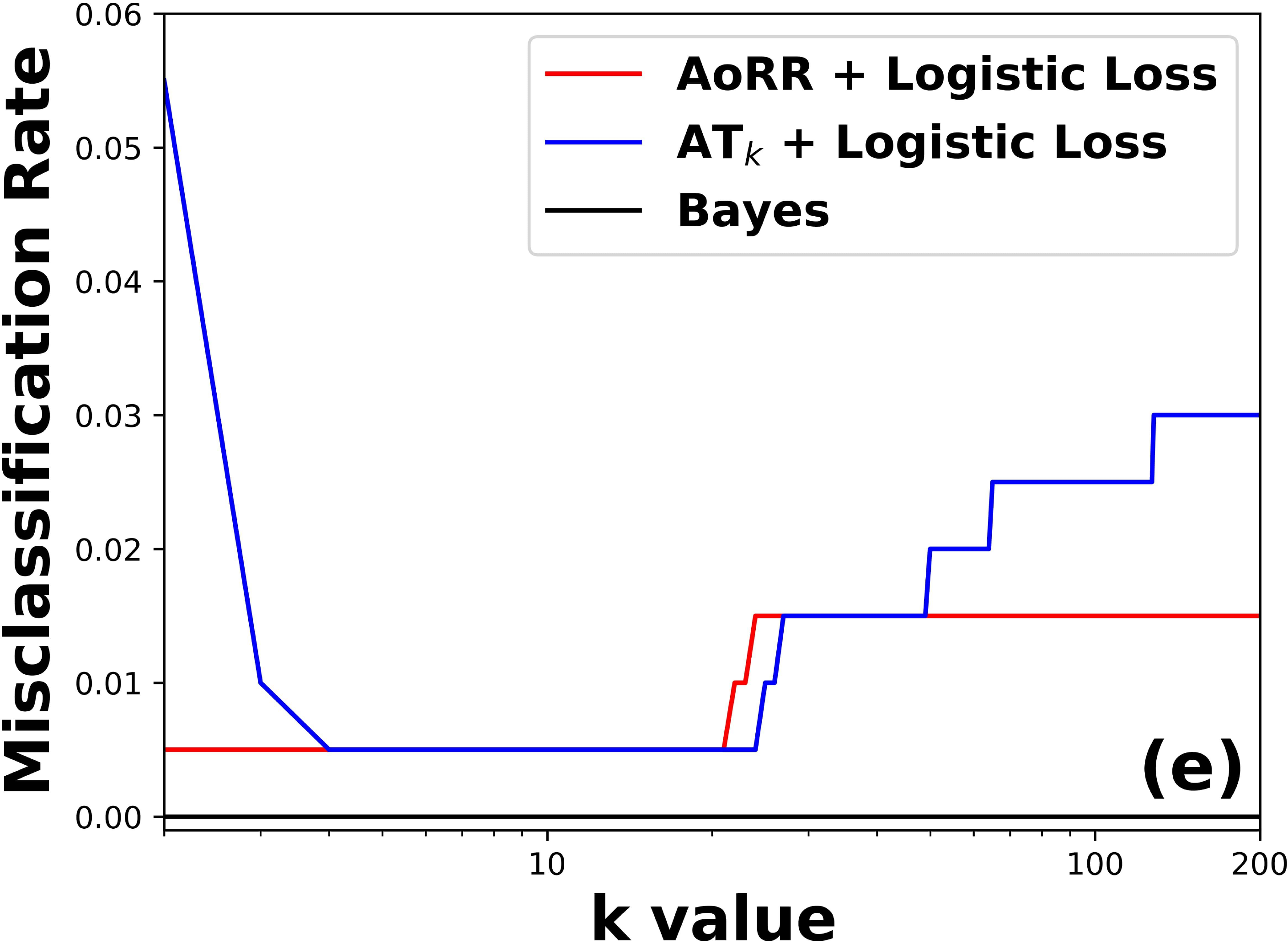}
        \end{subfigure}%
        \begin{subfigure}[b]{0.243\textwidth}
                \includegraphics[width=\linewidth]{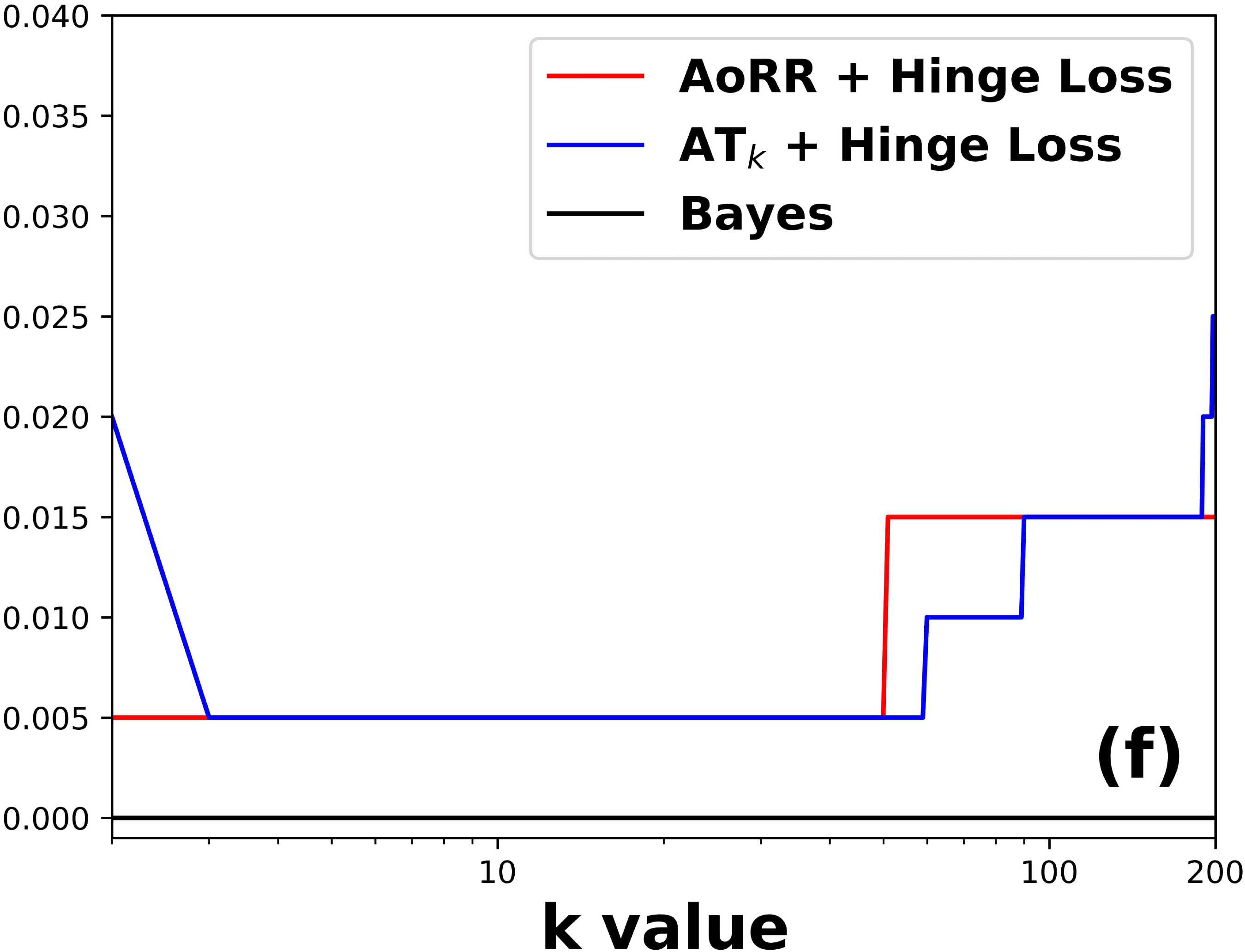}
        \end{subfigure}%
        \rulesep
        \begin{subfigure}[b]{0.243\textwidth}
                \includegraphics[width=\linewidth]{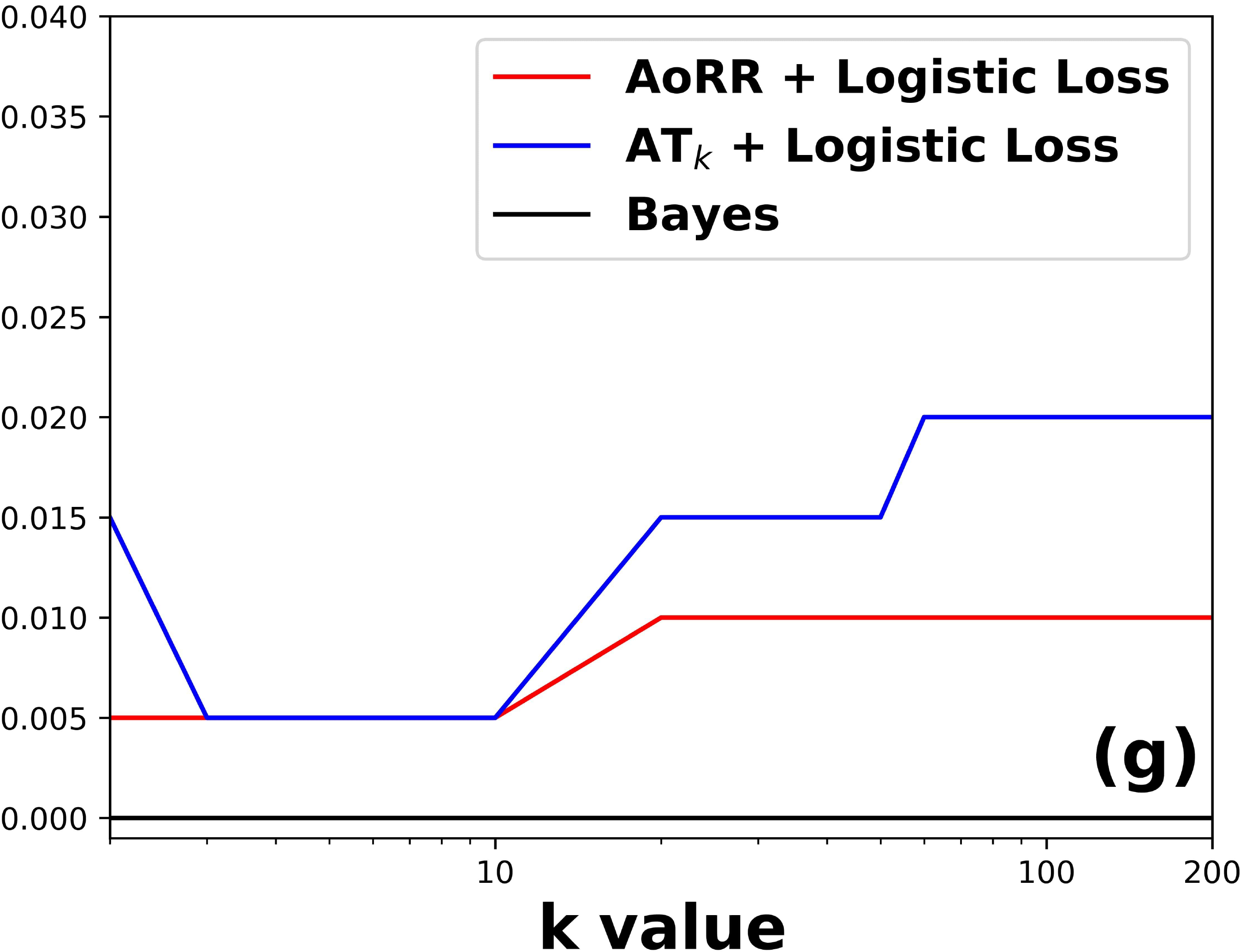}
        \end{subfigure}%
        \begin{subfigure}[b]{0.246\textwidth}
                \includegraphics[width=\linewidth]{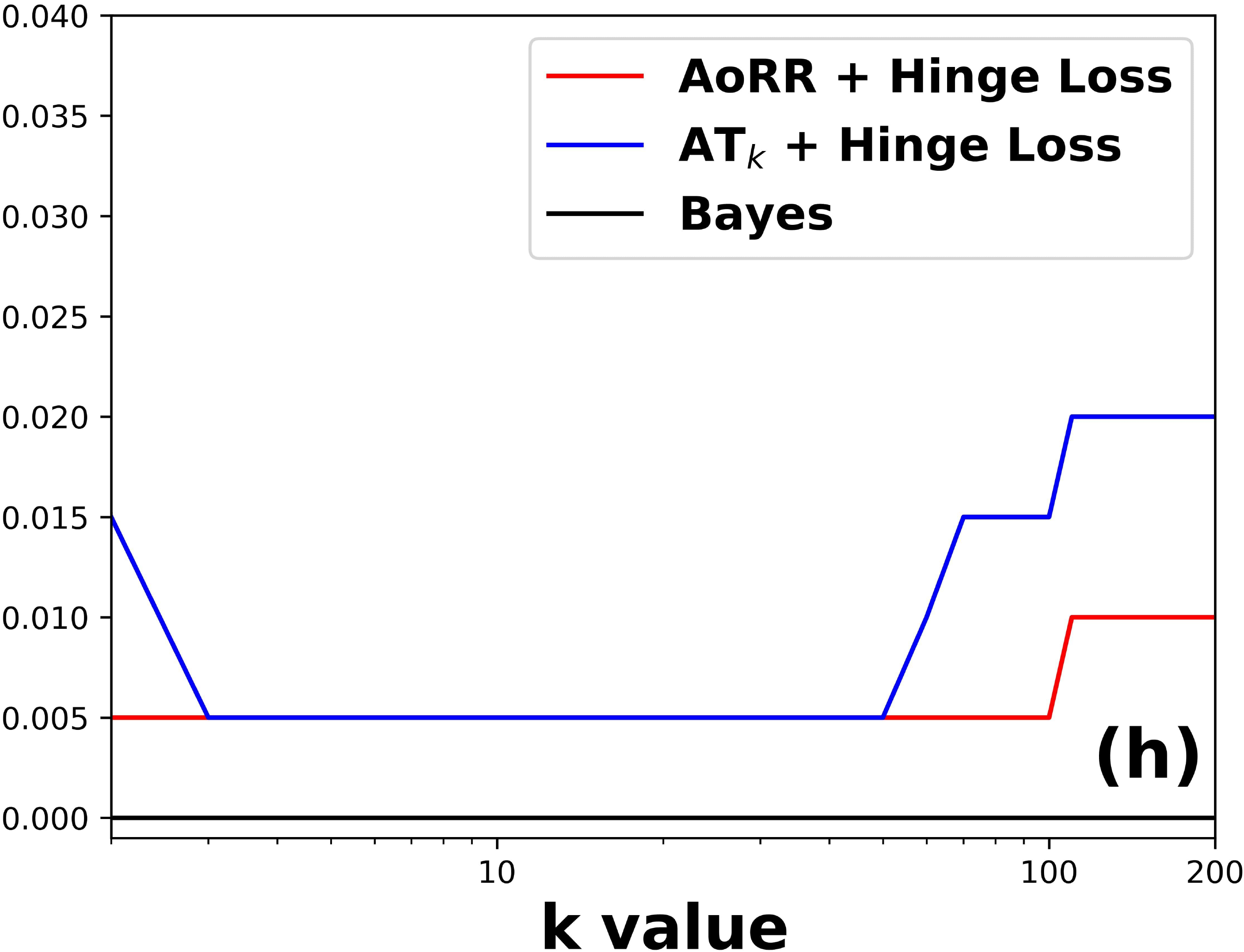}
        \end{subfigure}
        \vspace{-2em}
        \caption{\small \em Comparison of different aggregate losses for binary classification on a balanced but multi-modal synthetic dataset and with outliers with logistic loss (a) and hinge loss (b), and an imbalanced synthetic dataset with outliers with logistic loss (c) and hinge loss (d). Outliers in data are shown as $\times$ in blue class. {The figures (e), (f), (g) and (h) show the misclassification rates of \aorr~w.r.t. different values of $k$ for each case and compare with the AT$_k$ and the optimal Bayes classifier.}}\label{fig: aggregate_loss_toy_example}
\vspace{-2em}
\end{figure*}

\textbf{Synthetic data.} We generate two sets of 2D synthetic data (Fig.\ref{fig: aggregate_loss_toy_example}). Each dataset contains $200$ samples from Gaussian distributions with different means and variances. We consider both the case of the balanced (Fig.\ref{fig: aggregate_loss_toy_example} (a,b)) and the imbalanced  (Fig.\ref{fig: aggregate_loss_toy_example} (c,d)) data distributions, in the former the training data for the two classes are approximately equal while in the latter one class has a dominating number of samples in comparison to the other. The learned linear classifiers with different aggregate losses are shown in Fig.\ref{fig: aggregate_loss_toy_example}. Both datasets have an outlier in the blue class (shown as $\times$). Experiments with more outliers can be found in the Appendix \ref{additional_exp_more_outliers}.

To optimally remove the effect of outliers, we need to set $k$ larger than the number of outliers in the training dataset. Since there is one outlier in this synthetic dataset, we select $k=2$ here as an example. As shown in Fig.\ref{fig: aggregate_loss_toy_example}, neither the maximum loss nor the average loss performs well on the synthetic dataset, due to the existence of outliers and the multi-modal nature of the data. Furthermore, 
Fig.\ref{fig: aggregate_loss_toy_example} also shows that the AT$_k$ loss does not bode well: it is still affected by outliers. The reason can be that the training process with the AT$_k$ loss with $k=2$ will most likely pick up one individual loss from the outlier for optimization.  In contrast, the \aorr~loss with $k$=2 and $m$=1, which is equivalent to the top-$2$ or second largest individual loss, yields better classification results. Intuitively, we avoid the direct effects of the outlier since it has the largest individual loss value. {Furthermore, we perform experiments to show misclassification rates of \aorr~with respect to different values of $k$ in Fig.\ref{fig: aggregate_loss_toy_example} (e), (f), (g), (h) for each case and compare with the AT$_k$ loss and optimal Bayes classifier. The results show that for $k$ values other than 2, the \aorr~ loss still exhibits an advantage over the AT$_k$ loss. Our experiments are based on a grid search for selecting the value of $k$ and $m$ because we found it is simple and often yields comparable performance.} In practice {for large-scale datasets}, we can decide the minimal value of $m$ if we have prior knowledge about the faction of outliers in the dataset. To avoid extra freedom due to the value of $k$, we follow a very popular adaptive setting which has been applied in previous works, (e.g., \cite{kawaguchiordered}). At the beginning of training, $k$ equals to the size ($n$) of training data, $k=\lfloor \frac{n}{2} \rfloor$ once training accuracy $\geq 70\%$,  $k=\lfloor \frac{n}{4} \rfloor$ once training accuracy $\geq 80\%$, $k=\lfloor \frac{n}{8} \rfloor$ once training accuracy $\geq 90\%$, $k=\lfloor \frac{n}{16} \rfloor$ once training accuracy $\geq 95\%$, $k=\lfloor \frac{n}{32} \rfloor$ once training accuracy $\geq 99.5\%$.

\textbf{Real data.} We use five benchmark datasets from the UCI \cite{Dua:2019} and the KEEL \cite{alcala2011keel} data repositories (Statistical information of each dataset is given in the Appendix \ref{aggregate_loss_datasets}). For each dataset, we first randomly select $50\%$ samples for training, and the remaining $50\%$ samples are randomly split for validation and testing (each contains 25\% samples). Hyper-parameters $C$, $k$, and $m$ are selected based on the validation set. Specifically, parameter $C$ is chosen from $\{10^0, 10^1, 10^2, 10^3, 10^4, 10^5\}$, parameter $k\in\{1\}\cup [0.1:0.1:1]n$, where $n$ is the number of training samples, and parameter $m$ are selected in the range of $[1,k)$. The following results are based on the optimal values of $k$ and $m$ obtained based on the validation set. The random splitting of the training/validation/testing sets is repeated $10$ times and the average error rates, as well as the standard derivation on the testing set are reported in Table \ref{tab:aggregate_real_experiments}. {In \cite{shalev2016minimizing}, the authors introduce slack variables to indicate outliers and propose a robust version of the maximum loss. We term it as Robust\_Max loss and compare it to our method as one of the baselines}. As these results show, comparing to the maximum, {Robust\_Max}, average, and AT$_k$ losses, the \aorr~loss achieves the best performance on all five datasets with both individual logistic loss and individual hinge loss. For individual logistic  loss, the \aorr~loss significantly improves the classification performance on {\tt Monk} and {\tt Phoneme} datasets and a slight improvement on datasets {\tt Titanic} and {\tt Splice}. More specifically, the performance of maximum aggregate loss is very poor on all cases due to its high sensitivity to the outliers or noisy data. {The optimization of the Robust\_Max loss uses convex relaxation on the domain of slack variables constraint and using $l_2$ norm to replace the $l_1$ norm in the constraint. Therefore, it can alleviate the sensitivity to outliers, but cannot exclude the influence of them.}  The average aggregate loss is more robust to noise and outliers than the maximum loss { and the Robust\_Max loss} on all datasets. However, as data distributions may be very complicated, the average loss may sacrifice samples from rare distributions to pursue a lower loss on the whole training set and obtains sub-optimal solutions accordingly. The AT$_k$ loss is not completely free from the influence of outliers and noisy data either, which can be observed in particular on the Monk dataset. On the Monk dataset, {in comparison to the AT$_k$ loss, the \aorr~loss reduce the misclassification rates by $4.07\%$ for the individual logistic loss  and $3.87\%$ for the individual hinge loss, respectively.} 

\begin{figure*}[t!]
\captionsetup[subfigure]{justification=centering}
\centering
        \begin{subfigure}[b]{0.256\textwidth}
                \includegraphics[width=\linewidth]{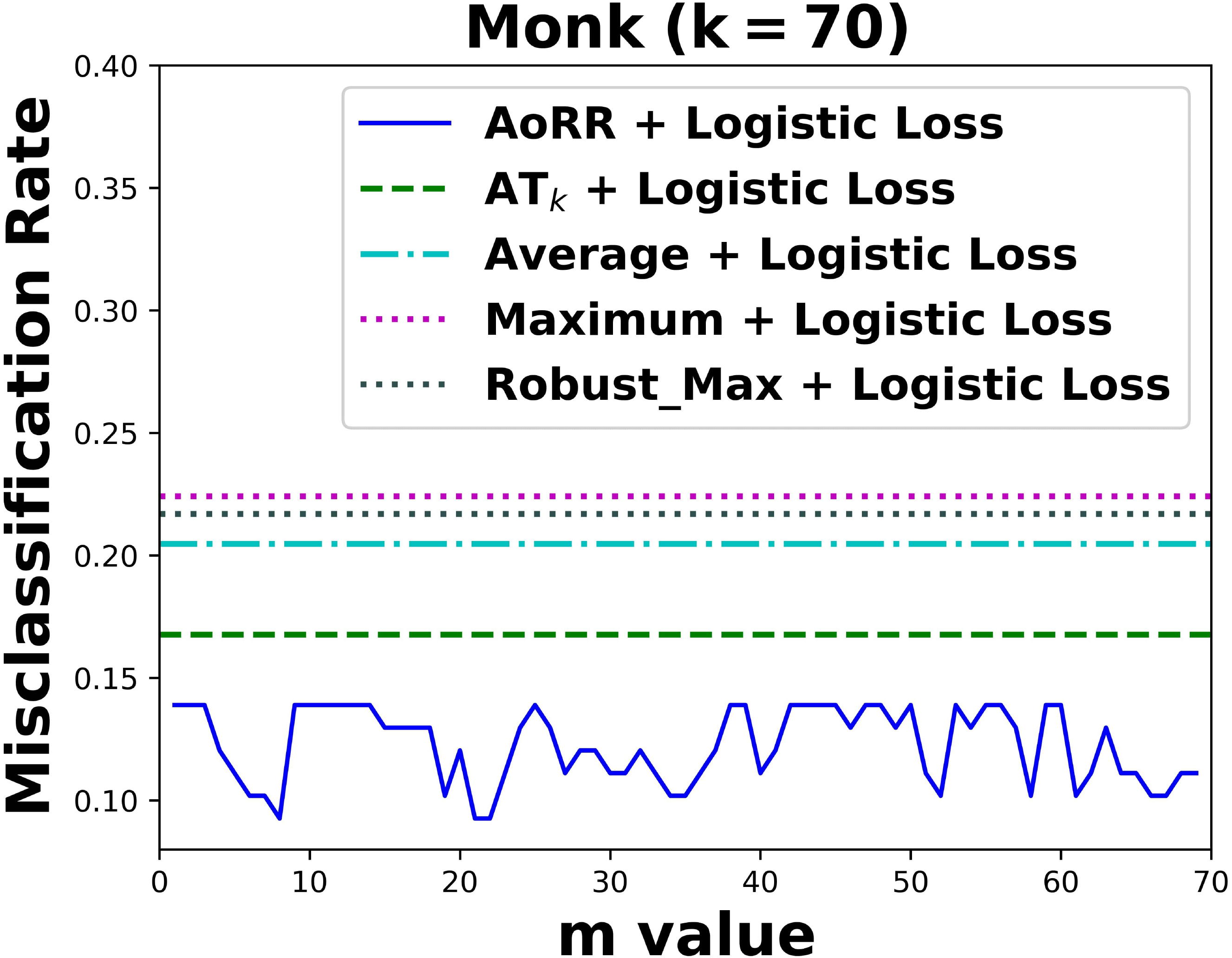}
        \end{subfigure}%
        \begin{subfigure}[b]{0.243\textwidth}
                \includegraphics[width=\linewidth]{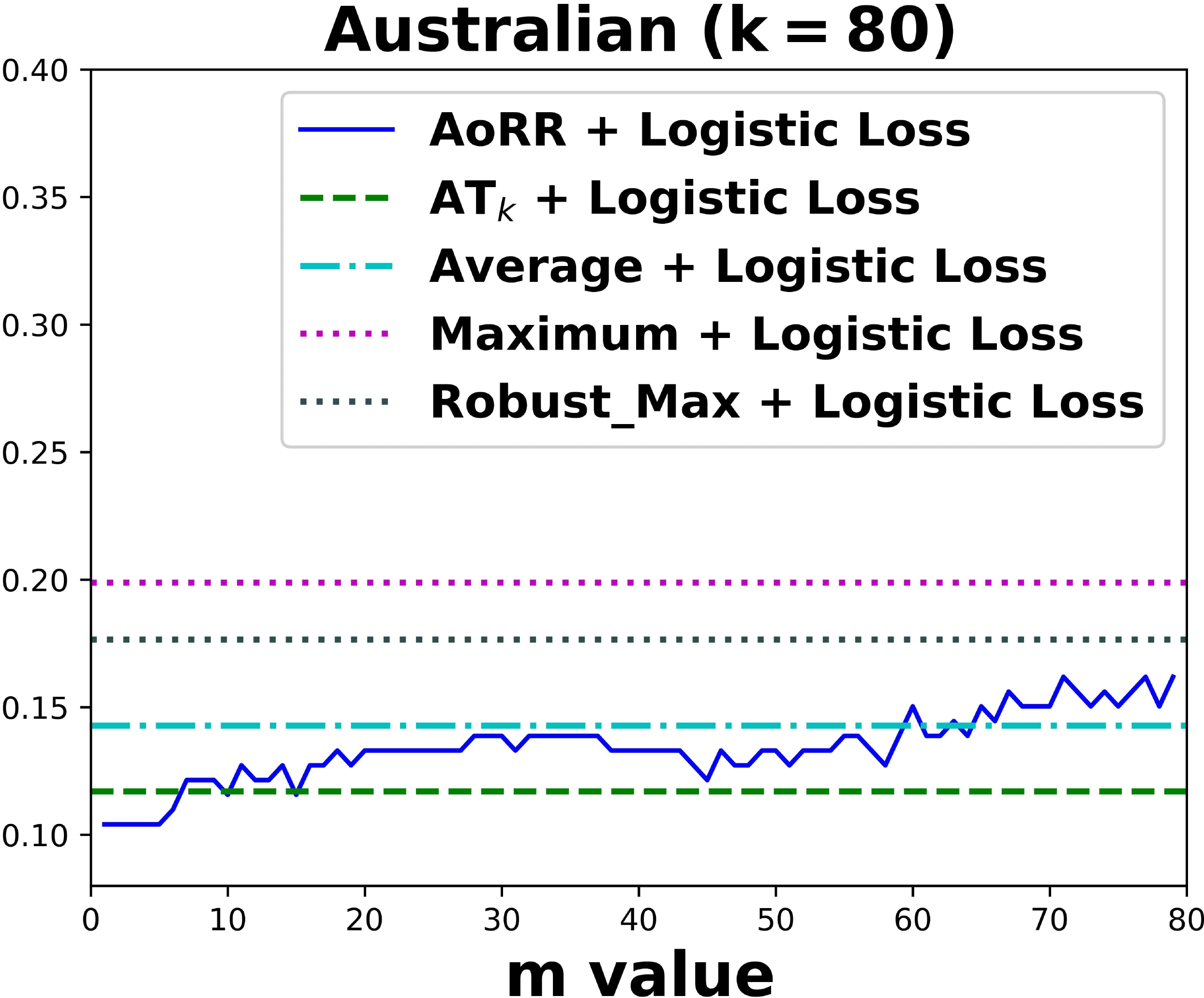}
        \end{subfigure}%
        \begin{subfigure}[b]{0.245\textwidth}
                \includegraphics[width=\linewidth]{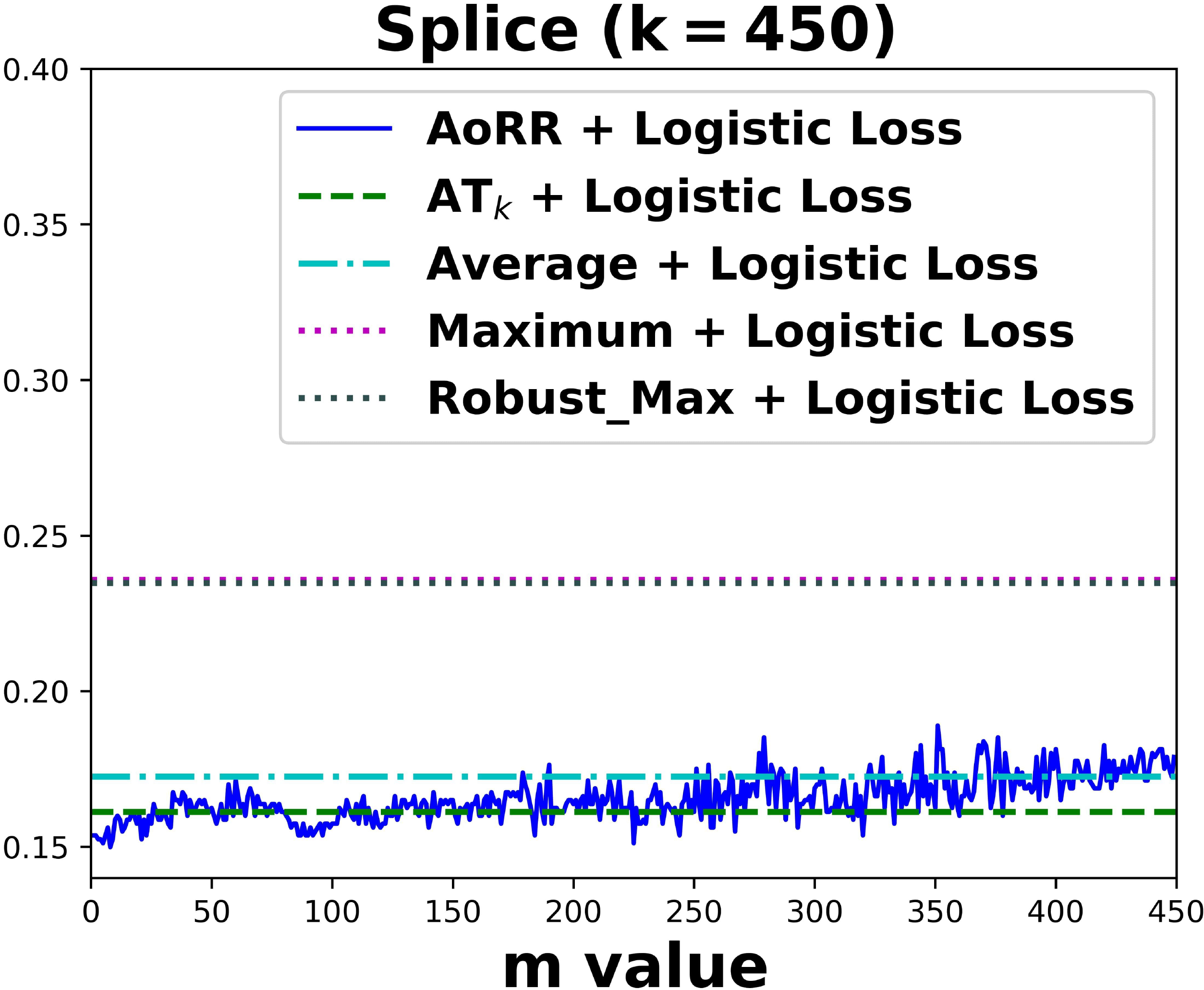}
        \end{subfigure}%
        \begin{subfigure}[b]{0.251\textwidth}
                \includegraphics[width=\linewidth]{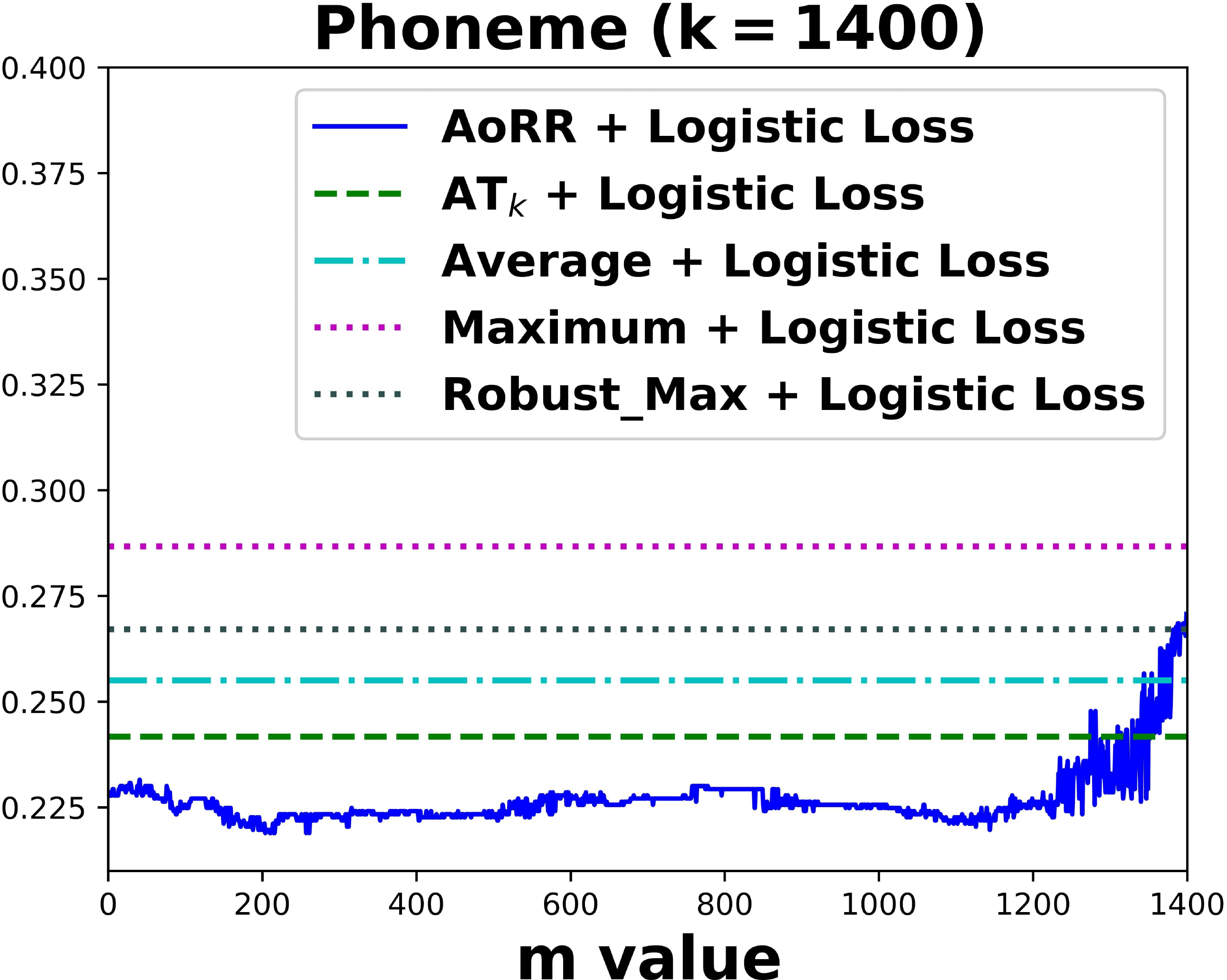}
        \end{subfigure}
        \caption{\small \em Tendency curves of error rate of learning AoRR loss w.r.t. $m$ on four datasets.}\label{fig: aggregate_k_prime_value}
\vspace{-5mm}
\end{figure*}

To further compare with the AT$_k$ loss, we investigate the influence of $m$ in the \aorr~loss. Specifically, we select the best $k$ value based on the AT$_k$ results, and vary $m$ in the range of $[1,k-1]$. We use the individual logistic loss and plot tendency curves of misclassification error rates w.r.t $m$ in Fig. \ref{fig: aggregate_k_prime_value}, together with those from the average, maximum {and Robust\_Max} losses. As these plots show, on all four datasets, there is a clear range of $m$ with better performance than the corresponding AT$_k$ loss. We observe a trend of decreasing error rates with $m$ increasing. This is because outliers correspond to large individual losses, and excluding them from the training loss helps improve the overall performance of the learned classifier. However, when $m$ becomes large, the classification performance is decreasing, as many samples with small losses are included in the \aorr~objective and dominate the training process. The results for the individual hinge loss can be found in the Appendix \ref{Additional Tendency Curves}.

\begin{table*}[]
\captionsetup{font=footnotesize}
\centering
\scriptsize{
\begin{tabular}{c|ccccc||ccccc}
\hline
\multirow{2}{*}{Datasets} & \multicolumn{5}{c||}{Logistic Loss} & \multicolumn{5}{c}{Hinge Loss} \\ \cline{2-11} 
                  &  Maximum  & R\_Max &    Average &  AT$_k$   & \aorr  &    Maximum & R\_Max  &    Average &  AT$_k$   & \aorr    \\ \hline
        \multirow{2}{*}{Monk} &  22.41 & 21.69  &  20.46   &  16.76   &  \textbf{12.69}   & 22.04 & 20.61  & 18.61    &  17.04   & \textbf{13.17}   \\  
                  &  (2.95) & (2.62)  &  (2.02)   & (2.29)   &  \textbf{(2.34)}   & (3.08) & (3.38)  & (3.16)    &  (2.77)   & \textbf{(2.13)}   \\ \hline
\multirow{2}{*}{Australian} &  19.88& 17.65  &  14.27   &  11.7   &  \textbf{11.42}   &  19.82& 15.88  &  14.74   &  12.51   & \textbf{12.5}   \\  
                  &  (6.64)& (1.3)  &  (3.22)   &  (2.82)   &  \textbf{(1.01)}   &  (6.56)& (1.05)  &  (3.10)   &  (4.03)   & \textbf{(1.55)}   \\ \hline
\multirow{2}{*}{Phoneme} &  28.67& 26.71   &  25.50   &  24.17   & \textbf{21.95}    & 28.81&  24.21   &  22.88   &  22.88   &  \textbf{21.95}   \\  
                  &  (0.58)& (1.4)   &  (0.88)   &  (0.89)   & \textbf{5(0.71)}    & (0.62)&  (1.7)   &  (1.01)   &  (1.01)   &  \textbf{(0.68)} \\ \hline
\multirow{2}{*}{Titanic} &  26.50& 24.15 &  22.77   &  22.44   & \textbf{21.69}    &  25.45& 25.08    &  22.82   &  22.02   &  \textbf{21.63}   \\  
                  &  (3.35)& (3.12)   &  (0.82)   &  (0.84)   & \textbf{(0.99)}    &  (2.52)& (1.2)   &  (0.74)   &  (0.77)   &  \textbf{(1.05)} \\\hline 
\multirow{2}{*}{Splice} &  23.57&  23.48&  17.25 &  16.12 & \textbf{15.59}    & 23.40&  22.82& 16.25 &  16.23  & \textbf{15.64}   \\  
                  &  (1.93)&  (0.76)  &  (0.93)   &  (0.97)   & \textbf{(0.9)}    & (2.10)&  (2.63)   & (1.12)    &  (0.97)   & \textbf{(0.89)}   \\ \hline
\end{tabular}
\vspace{1mm}
\caption{\small \em Average error rate (\%) and standard derivation of
different aggregate losses combined with individual logistic loss and hinge loss over 5 datasets. The best results are shown in bold. (R\_Max: Robust\_Max)}
\label{tab:aggregate_real_experiments}
}
\vspace{-2mm}
\end{table*}

\section{Multi-label Learning}

We use \sorr~to construct the individual loss for multi-label/multi-class classification, where a sample $x$ can be associated with a set of labels $\oldemptyset \not = Y \subset \{1,\cdots,l\}$. Our goal is to construct a linear predictor $f_{\Theta}(x) = \Theta^T x$ with $\Theta = (\theta_1,\cdots,\theta_l)$.  The final classifier outputs labels for $x$ with the top $k$ ($1 \le k < l$) prediction scores, \ie, $\theta^\top_{[1]}x \geq  \theta^\top_{[2]}x \geq \cdots \geq  \theta^\top_{[k]}x$. In training, the classifier is expected to include as many true labels as possible in the top $k$ outputs. This can be evaluated by the ``margin'', \ie, the difference between the $(k+1)$-th largest score of all the labels, $\theta^\top_{[k+1]}x$ and the lowest prediction score of all the ground-truth labels, $\min_{y \in Y}\theta^\top_{y}x$. If we have $\theta^\top_{[k+1]}x < \min_{y \in Y}\theta^\top_{y}x$, then all ground-truth labels have prediction scores ranked in the top $k$ positions. If this is not the case, then at least one ground-truth label has a prediction score not ranked in the top $k$. This induces the following metric for multi-label classification as $\mathbb{I}_{[\theta^\top_{[k+1]}x \ge \min_{y \in {Y}}\theta^\top_{y}x]}$. Replacing the indicator function with the hinge function and let $S(\theta)=\{s_{j}(\theta)\}_{j=1}^l$, where $s_{j}(\theta)=\big[1+ \theta^\top_{j}x - \min_{y \in {Y}}\theta^\top_{y}x\big]_+$, we obtain a continuous surrogate loss, as    $\psi_{k,k+1}(S(\theta))= s_{[k+1]}(\theta)$.
We term this loss as the {\em \underline{t}op-$\underline{k}$ \underline{m}ulti-\underline{l}abel} (\tkml) loss. When $k = |Y|$, we have the following proposition and its proof can be found in the Appendix \ref{appendix:proof_proposition1},
\begin{prop}
The \tkml~loss is a lower-bound to the conventional multi-label loss \cite{crammer2003family}, as
$ 
\big[1+ \max_{y \not \in Y}\theta^\top_{y}x -   \min_{y \in {Y}}\theta^\top_{y}x\big]_+ \ge \psi_{|Y|,|Y|+1}(S(\theta)).
$
\end{prop}
The \tkml~loss generalizes the conventional multi-class loss ($|Y| =k=1$) and the top-$k$ consistent $k$-guesses multi-class classification \cite{yang2019consistency} ($1 = |Y| \leq k < l$). A similar learning objective is proposed in \cite{lapin2015top} corresponds to $s_{[k]}(\theta)$, however, as proved in \cite{yang2019consistency}, it is not multi-class top-$k$ consistent. 
{Another work in  \cite{chang2017robust} proposes a robust top-k multi-class SVM based on the convex surrogate  of $s_{[k]}(\theta)$ to address the outliers by using a hyperparameter to cap the values of the individual losses. This approach is different from ours since we directly address the original top-k multi-class SVM problem using our \tkml~loss without introducing its convex surrogate and it is consistent.} For a set of training data $(x_1,Y_1),\cdots,(x_n,Y_n)$, if we denote $\psi_{k,k+1}(S(x,Y;\Theta)) = \psi_{k,k+1}(S(\theta))$, the data loss on \tkml~can be written as $\L_{TKML}(\Theta)=\frac{1}{n}\sum_{i=1}^n \psi_{k,k+1}(S(x_i,Y_i;\Theta))$, which can be optimized using the Algorithm \ref{Alg0}.

\subsection{Experiments}
We use the same $\ell_2$ regularizer, $R(\theta)=\frac{1}{2C}||\theta||_2^2$ and cross-validate hyper-parameter $C$ in the range 10$^0$ to 10$^5$, extending it when the optimal value appears.

\textbf{Multi-label classification.} We use three benchmark datasets (Emotions, Scene, and Yeast) from the KEEL data repository to verify the effectiveness of our \tkml~loss. The average number of positive labels per instance in each dataset is 1.81, 1.06, and 4.22, respectively. For comparison, we compare \tkml~with logistic regression (LR) model (i.e. minimize a surrogate hamming loss \cite{zhang2013review}), and a ranking based method (LSEP \cite{li2017improving}). For these two baseline methods, we use a sigmoid operator on the linear predictor as $f_{\Theta}(x)=1/(1+\exp(-\Theta^Tx)).$  Since \tkml~is based on the value of $k$, we use five different $k$ values ($k\in\{1,2,3,4,5\}$) to evaluate the performance. For each dataset, we randomly partition it to 50\%/25\%/25\% samples for training/validation/testing, respectively. This random partition is repeated 10 times, and the average performance on testing data is reported in Table \ref{tab:multilabel_experiments}. 
We use a metric (top $k$ multi-label accuracy) $\frac{1}{n}\sum_{i=1}^n \mathbb{I}_{[(Z_i\subseteq Y_i) \vee (Y_i\subseteq Z_i)]}$ to evaluate the performance, where $n$ is the size of the sample set. For instance, $(x_i, Y_i)$ with $Y_i$ be its ground-truth set, $f_\Theta(x_i) \in \mathbb{R}^l$ be its predicted scores, and $Z_i$ be a set of top $k$ predictions according to $f_\Theta(x_i)$. This metric reflects the performance of a classifier can get as many true labels as possible in the top $k$ range. More details about these three datasets and settings can be found in the Appendix \ref{Multi-label_datasets} and \ref{sec:multi_label_settings}. 

From Table \ref{tab:multilabel_experiments}, we note that the \tkml~loss in general improves the performance on the Emotions dataset for all different $k$ values. These results illustrate the effectiveness of the \tkml~loss. More specifically, our \tkml~method obtains 4.63\% improvement on $k=2$ and 4.42\% improvement on $k=3$ when comparing to LR. This rate of improvement becomes higher (6.26\% improvement on $k=2$) when compare to LSEP. We also compare the performance of the method based on the \tkml~loss on different $k$ values. If we choose the value of $k$ close to the number of the ground-truth labels, the corresponding classification method outperforms the two baseline methods. For example, in the case of the Emotions dataset, the average number of positive labels per instance is $1.81$, and our method based on the \tkml~loss achieves the best performance for $k=1,2$. As another example, the average number of true labels for the Yeast dataset is 4.22, so the method based on the \tkml~loss achieves the best performance for $k=4,5$. {We provide more experiments in Appendix \ref{sec:Additional_Evaluation_Metric}. }

\textbf{Robustness analysis.} As a special case of \aorr, the \tkml~loss exhibit similar robustness with regards to outliers, which can be elucidated with experiments in the multi-class setting (\ie, $k$=1 and $|Y|$=1). We use the MNIST dataset\cite{lecun1998gradient}, which contains $60,000$ training samples and $10,000$ testing samples that are images of handwritten digits.  To simulate outliers caused by errors occurred when labeling the data, as in the work of \cite{wang2019symmetric}, we use the asymmetric (class-dependent) noise creation method \cite{patrini2017making, zhang2018generalized} to randomly change labels of the training data (2$\rightarrow$7, 3$\rightarrow$8, 5$\leftrightarrow $6, and 7$\rightarrow$1) with a given proportion. The flipping label is chosen at random with probability $p = 0.2, 0.3, 0.4$. As a baseline, we use the top-$k$ multi-class SVM (SVM$_{\alpha}$) \cite{lapin2015top}. The performance is evaluated with the top 1, top 2, $\cdots$, top 5 accuracy on the testing samples.  

\begin{table}[t]
\captionsetup{font=footnotesize}
\centering
\scalebox{0.8} { 
\begin{tabular}{|c|c|c|c|c|c|c|}
\hline
                Datasets  & Methods & $k$=1 & $k$=2 & $k$=3 & $k$=4 & $k$=5 \\ \hline
\multirow{3}{*}{Emotions} & LR & 73.54(3.98) & 57.48(3.35) & 73.20(4.69) & 86.60(3.02) & 96.46(1.71) \\ \cline{2-7} 
                  & LSEP & 72.18(4.56) & 55.85(3.37) & 72.18(3.74) & 85.58(2.92) & 95.85(1.07) \\ \cline{2-7} 
                  & \tkml & \textbf{76.80(2.66)} & \textbf{62.11(2.85)} & \textbf{77.62(2.81)} & \textbf{90.14(2.22)} & \textbf{96.94(0.63)} \\ \hline
\multirow{3}{*}{Scene} & LR & 73.2(0.57) & 85.31(0.47) & \textbf{94.79(0.79)} & \textbf{97.88(0.63)} & \textbf{99.7(0.30)} \\ \cline{2-7} 
                  & LSEP & 69.22(3.43) & 83.83(4.83) & 92.46(4.78) & 96.35(3.5) & 98.56(1.94) \\ \cline{2-7} 
                  & \tkml & \textbf{74.06(0.45)} & \textbf{85.36(0.79)} & 88.92(1.47) & 91.94(0.87) & 95.01(0.61) \\ \hline
\multirow{3}{*}{Yeast} & LR & \textbf{77.57(0.91)} & \textbf{70.59(1.16)} & \textbf{52.65(1.23)} & 43.26(1.16) & 43.49(1.33) \\ \cline{2-7} 
                  & LSEP & 75.5(1.03) & 66.84(2.9) & 49.72(1.26) & 41.90(1.91) & 43.01(1.02) \\ \cline{2-7} 
                  & \tkml & 76.94(0.49) & 67.19(2.79) & 45.41(0.71) & \textbf{43.47(1.06)} & \textbf{44.69(1.14)} \\ \hline
\end{tabular}
}
\vspace{1mm}
\caption{\small \em Top $k$ multi-label accuracy with its standard derivation (\%) on three datasets. The best performance is shown in bold.}
\label{tab:multilabel_experiments}
\vspace{-4mm}
\end{table}

From Table \ref{tab:MSVM_real_experiments}, it is clear that our method \tkml~consistently outperforms the baseline SVM$_{\alpha}$ among all top 1-5 accuracies. The gained improvement in performance is getting more significant as the level of noise increases. Since our flipping method only works between two different labels, we expected the  performance of \tkml~has some significant improvements on top 1 and 2 accuracies. Indeed, this  expectation is correctly verified as Table \ref{tab:MSVM_real_experiments} clearly indicates that the performance of our method is better than SVM$_\alpha$ by nearly 7\% accuracy (see Top-1 accuracy in the noise level 0.4). These results also demonstrate our optimization framework works well. More experiments can be found in the Appendix \ref{additional_exp_kgmc}.

\begin{table}[t]
\captionsetup{font=footnotesize}
\centering
\scalebox{0.8} {
\begin{tabular}{|c|c|c|c|c|c|c|}
\hline
           Noise Level & Methods & Top-1 Accuracy & Top-2 Accuracy & Top-3 Accuracy& Top-4 Accuracy & Top-5 Accuracy \\ \hline
\multirow{2}{*}{0.2} & SVM$_\alpha$ & 78.33(0.18) & 90.66(0.29) & 95.12(0.2) & 97.28(0.09) & 98.49(0.1)  \\  
                  & \tkml~ & \textbf{83.06(0.94)} & \textbf{94.17(0.19)} & \textbf{97.24(0.13)} & \textbf{98.47(0.05)} & \textbf{99.22(0.01)} \\ \hline
\multirow{2}{*}{0.3} & SVM$_\alpha$ & 74.65(0.17) & 89.31(0.24) & 94.14(0.2) & 96.73(0.23) & 98.19(0.07)  \\ 
                  & \tkml~ & \textbf{80.13(1.24)} & \textbf{93.37(0.1)} & \textbf{96.81(0.22)} & \textbf{98.21(0.05)} & \textbf{99.08(0.05)} \\ \hline
\multirow{2}{*}{0.4} & SVM$_\alpha$ & 68.32(0.32) & 86.71(0.42) & 93.14(0.49) & 96.16(0.32) & 97.84(0.18) \\ 
                  & \tkml~ & \textbf{75(1.15)} & \textbf{92.41(0.14)} & \textbf{96.2(0.13)} & \textbf{97.95(0.1)} & \textbf{98.89(0.04)} \\ \hline
\end{tabular}
}
\vspace{1mm}
\caption{\small \em Testing accuracy (\%) of two methods on MNIST with different levels of asymmetric noisy labels. The average accuracy and standard deviation of 5 random runs are reported and the best results are shown in bold.}
\label{tab:MSVM_real_experiments}
\vspace{-2em}
\end{table}

\section{Conclusion}
In this work, we introduce a general approach to form learning objectives, \ie, sum of ranked range, which corresponds to the sum of a consecutive sequence of sorted values of a set of real numbers. We show that \sorr~can be expressed as the difference between two convex problems and optimized with the difference-of-convex algorithm (DCA). 

We explore two applications in machine learning of the minimization of the \sorr~framework, namely the \aorr~aggregate loss for binary classification and the \tkml~individual loss for multi-label/multi-class classification. 
Our empirical results showed the effectiveness of the proposed framework on achieving superior generalization and robust performance on synthetic and real datasets. For future works, we plan to further study the consistency of \tkml~ loss for multi-label learning and incorporate \sorr~ into the learning of deep neural networks.

\section{Broader Impact}

Loss functions are fundamental components in any machine learning system. Our work, by designing new types of loss functions based on the use of \sorr~, is expected to be applicable to a wide range of ML problems. The benefit of using our method is the better handling of potential outliers in the training dataset, which could be the result of gross error or intentional ``poisoning'' of the dataset. However, there is also a risk of resulting a biased learning model when certain training samples are excluded. To mitigate such risks, we encourage further study to understand the impacts of using \sorr~based losses in particular real-world scenarios, focusing on the more contextually meaning choice of the values $m$ and $k$ for better tradeoff of robustness and bias. 


\smallskip
\noindent {\bf Acknowledgments}.
We are grateful to all anonymous reviewers for their constructive comments. This work is supported by NSF research grants (IIS-1816227 and IIS-2008532) as well as an Army Research Office grant (agreement number: W911 NF-18-1-0297).


\bibliography{neurips_2020}

\begin{thebibliography}{10}

\bibitem{alcala2011keel}
J.~Alcal{\'a}-Fdez, A.~Fern{\'a}ndez, J.~Luengo, J.~Derrac, S.~Garc{\'\i}a,
  L.~S{\'a}nchez, and F.~Herrera.
\newblock Keel data-mining software tool: data set repository, integration of
  algorithms and experimental analysis framework.
\newblock {\em Journal of Multiple-Valued Logic \& Soft Computing}, 17, 2011.

\bibitem{bartlett2006convexity}
P.~L. Bartlett, M.~I. Jordan, and J.~D. McAuliffe.
\newblock Convexity, classification, and risk bounds.
\newblock {\em Journal of the American Statistical Association},
  101(473):138--156, 2006.

\bibitem{bertsekas1971control}
D.~P. Bertsekas.
\newblock {\em Control of uncertain systems with a set-membership description
  of the uncertainty.}
\newblock PhD thesis, Massachusetts Institute of Technology, 1971.

\bibitem{bhat2019concentration}
S.~P. Bhat and L.~Prashanth.
\newblock Concentration of risk measures: A wasserstein distance approach.
\newblock In {\em Advances in Neural Information Processing Systems}, pages
  11739--11748, 2019.

\bibitem{bottou2008tradeoffs}
L.~Bottou and O.~Bousquet.
\newblock The tradeoffs of large scale learning.
\newblock In {\em Advances in neural information processing systems}, pages
  161--168, 2008.

\bibitem{brown2007large}
D.~B. Brown.
\newblock Large deviations bounds for estimating conditional value-at-risk.
\newblock {\em Operations Research Letters}, 35(6):722--730, 2007.

\bibitem{chang2017robust}
X.~Chang, Y.-L. Yu, and Y.~Yang.
\newblock Robust top-k multiclass svm for visual category recognition.
\newblock In {\em Proceedings of the 23rd ACM SIGKDD International Conference
  on Knowledge Discovery and Data Mining}, pages 75--83, 2017.

\bibitem{crammer2003family}
K.~Crammer and Y.~Singer.
\newblock A family of additive online algorithms for category ranking.
\newblock {\em Journal of Machine Learning Research}, 3(Feb):1025--1058, 2003.

\bibitem{danskin2012theory}
J.~M. Danskin.
\newblock {\em The theory of max-min and its application to weapons allocation
  problems}, volume~5.
\newblock Springer Science \& Business Media, 2012.

\bibitem{Dua:2019}
D.~Dua and C.~Graff.
\newblock {UCI} machine learning repository, 2017.

\bibitem{fan2017learning}
Y.~Fan, S.~Lyu, Y.~Ying, and B.~Hu.
\newblock Learning with average top-k loss.
\newblock In {\em Advances in neural information processing systems}, pages
  497--505, 2017.

\bibitem{fan2020groupwise}
Y.~Fan, B.~Wu, R.~He, B.-G. Hu, Y.~Zhang, and S.~Lyu.
\newblock Groupwise ranking loss for multi-label learning.
\newblock {\em IEEE Access}, 8:21717--21727, 2020.

\bibitem{friedman2001elements}
J.~Friedman, T.~Hastie, and R.~Tibshirani.
\newblock {\em The elements of statistical learning}, volume~1.
\newblock Springer series in statistics New York, 2001.

\bibitem{kanamori2017robustness}
T.~Kanamori, S.~Fujiwara, and A.~Takeda.
\newblock Robustness of learning algorithms using hinge loss with outlier
  indicators.
\newblock {\em Neural Networks}, 94:173--191, 2017.

\bibitem{kawaguchiordered}
K.~Kawaguchi and H.~Lu.
\newblock Ordered sgd: A new stochastic optimization framework for empirical
  risk minimization.

\bibitem{lapin2015top}
M.~Lapin, M.~Hein, and B.~Schiele.
\newblock Top-k multiclass svm.
\newblock In {\em Advances in Neural Information Processing Systems}, pages
  325--333, 2015.

\bibitem{le2018dc}
H.~A. Le~Thi and T.~P. Dinh.
\newblock Dc programming and dca: thirty years of developments.
\newblock {\em Mathematical Programming}, 169(1):5--68, 2018.

\bibitem{lecun1998gradient}
Y.~LeCun, L.~Bottou, Y.~Bengio, and P.~Haffner.
\newblock Gradient-based learning applied to document recognition.
\newblock {\em Proceedings of the IEEE}, 86(11):2278--2324, 1998.

\bibitem{li2017improving}
Y.~Li, Y.~Song, and J.~Luo.
\newblock Improving pairwise ranking for multi-label image classification.
\newblock In {\em Proceedings of the IEEE conference on computer vision and
  pattern recognition}, pages 3617--3625, 2017.

\bibitem{lin2004note}
Y.~Lin.
\newblock A note on margin-based loss functions in classification.
\newblock {\em Statistics \& probability letters}, 68(1):73--82, 2004.

\bibitem{ma2011robust}
Y.~Ma, L.~Li, X.~Huang, and S.~Wang.
\newblock Robust support vector machine using least median loss penalty.
\newblock {\em IFAC Proceedings Volumes}, 44(1):11208--11213, 2011.

\bibitem{madjarov2012extensive}
G.~Madjarov, D.~Kocev, D.~Gjorgjevikj, and S.~D{\v{z}}eroski.
\newblock An extensive experimental comparison of methods for multi-label
  learning.
\newblock {\em Pattern recognition}, 45(9):3084--3104, 2012.

\bibitem{nie2017multiclass}
F.~Nie, X.~Wang, and H.~Huang.
\newblock Multiclass capped lp-norm svm for robust classifications.
\newblock In {\em Thirty-First AAAI Conference on Artificial Intelligence (AAAI
  2017)}, 2017.

\bibitem{ogryczak2003minimizing}
W.~Ogryczak and A.~Tamir.
\newblock Minimizing the sum of the k largest functions in linear time.
\newblock {\em Information Processing Letters}, 85(3):117--122, 2003.

\bibitem{ortis2019predicting}
A.~Ortis, G.~M. Farinella, and S.~Battiato.
\newblock Predicting social image popularity dynamics at time zero.
\newblock {\em IEEE Access}, 7:171691--171706, 2019.

\bibitem{patrini2017making}
G.~Patrini, A.~Rozza, A.~Krishna~Menon, R.~Nock, and L.~Qu.
\newblock Making deep neural networks robust to label noise: A loss correction
  approach.
\newblock In {\em Proceedings of the IEEE Conference on Computer Vision and
  Pattern Recognition}, pages 1944--1952, 2017.

\bibitem{phan2016dca}
D.~N. Phan.
\newblock {\em DCA based algorithms for learning with sparsity in high
  dimensional setting and stochastical learning}.
\newblock PhD thesis, 2016.

\bibitem{piot2016difference}
B.~Piot, M.~Geist, and O.~Pietquin.
\newblock Difference of convex functions programming applied to control with
  expert data.
\newblock {\em arXiv preprint arXiv:1606.01128}, 2016.

\bibitem{rakhlin2011making}
A.~Rakhlin, O.~Shamir, and K.~Sridharan.
\newblock Making gradient descent optimal for strongly convex stochastic
  optimization.
\newblock {\em arXiv preprint arXiv:1109.5647}, 2011.

\bibitem{shalev2016minimizing}
S.~Shalev-Shwartz and Y.~Wexler.
\newblock Minimizing the maximal loss: How and why.
\newblock In {\em ICML}, pages 793--801, 2016.

\bibitem{srebro2010stochastic}
N.~Srebro and A.~Tewari.
\newblock Stochastic optimization for machine learning.
\newblock {\em ICML Tutorial}, 2010.

\bibitem{tao1997convex}
P.~D. Tao and L.~T.~H. An.
\newblock Convex analysis approach to dc programming: theory, algorithms and
  applications.
\newblock {\em Acta mathematica vietnamica}, 22(1):289--355, 1997.

\bibitem{thi2017stochastic}
H.~A.~L. Thi, H.~M. Le, D.~N. Phan, and B.~Tran.
\newblock Stochastic dca for the large-sum of non-convex functions problem and
  its application to group variable selection in classification.
\newblock In {\em Proceedings of the 34th International Conference on Machine
  Learning-Volume 70}, pages 3394--3403. JMLR. org, 2017.

\bibitem{thi2019stochastic}
H.~A.~L. Thi, H.~M. Le, D.~N. Phan, and B.~Tran.
\newblock Stochastic dca for minimizing a large sum of dc functions with
  application to multi-class logistic regression.
\newblock {\em arXiv preprint arXiv:1911.03992}, 2019.

\bibitem{vapnik1992principles}
V.~Vapnik.
\newblock Principles of risk minimization for learning theory.
\newblock In {\em Advances in neural information processing systems}, pages
  831--838, 1992.

\bibitem{vapnik2013nature}
V.~Vapnik.
\newblock {\em The nature of statistical learning theory}.
\newblock Springer science \& business media, 2013.

\bibitem{wang2019symmetric}
Y.~Wang, X.~Ma, Z.~Chen, Y.~Luo, J.~Yi, and J.~Bailey.
\newblock Symmetric cross entropy for robust learning with noisy labels.
\newblock In {\em Proceedings of the IEEE International Conference on Computer
  Vision}, pages 322--330, 2019.

\bibitem{yang2019consistency}
F.~Yang and S.~Koyejo.
\newblock On the consistency of top-k surrogate losses.
\newblock {\em arXiv preprint arXiv:1901.11141}, 2019.

\bibitem{zhang2013review}
M.-L. Zhang and Z.-H. Zhou.
\newblock A review on multi-label learning algorithms.
\newblock {\em IEEE transactions on knowledge and data engineering},
  26(8):1819--1837, 2013.

\bibitem{zhang2018generalized}
Z.~Zhang and M.~Sabuncu.
\newblock Generalized cross entropy loss for training deep neural networks with
  noisy labels.
\newblock In {\em Advances in neural information processing systems}, pages
  8778--8788, 2018.

\end{thebibliography}

\bibliographystyle{abbrv}

\medskip

\small


\newpage

\appendix
\numberwithin{equation}{section}
\numberwithin{theorem}{section}
\numberwithin{remark}{section}
\renewcommand{\thesection}{{\Alph{section}}}
\renewcommand{\thesubsection}{\Alph{section}.\arabic{subsection}}
\renewcommand{\thesubsubsection}{\Roman{section}.\arabic{subsection}.\arabic{subsubsection}}
\setcounter{secnumdepth}{-1}
\setcounter{secnumdepth}{3}


{\bf \Large  Appendix}

\bigskip 

\section{Proofs}

\subsection{Proof of Theorem 1}
\label{proof_theorem_sorr}
To prove Theorem 1, we need the following lemma.
\begin{lemma}[\cite{ogryczak2003minimizing}]. $\phi_k(S)$ is a convex function of the elements of $S$. Furthermore, for any $i\in [1,n]$, we have $\sum_{i=1}^k s_{[i]}=min_{\lambda\in \mathbb{R}}\{k\lambda+\sum_{i=1}^n[s_i-\lambda]_+\}$, of which $s_{[k]}$ is an optimum solution.
\label{lemma:convex}
\end{lemma}

\textbf{Proof of Theorem 1}

\begin{proof}
From Lemma \ref{lemma:convex}, we have

\begin{equation*}
\begin{aligned}
\min_{\theta}\psi_{m,k}(S(\theta)) 
&= \min_{\theta} \bigl[ \phi_k(S(\theta)) -\phi_m(S(\theta))\bigr]               \\
&= \min_{\theta}\bigg[ \min_{\lambda\in \mathbb{R}}\Big\{k\lambda+\sum_{i=1}^n[s_i(\theta)-\lambda]_+\Big\}-\min_{\hat{\lambda}\in \mathbb{R}}\Big\{m\hat{\lambda}+\sum_{i=1}^n[s_i(\theta)-\hat{\lambda}]_+\Big\}\bigg].
\end{aligned}
\end{equation*}
If the optimal solution $\theta^*$ is achieved, from Lemma \ref{lemma:convex}, we get $\lambda = s_{[k]}$ and $\hat{\lambda}=s_{[m]}$. Therefore, $\hat{\lambda}> \lambda$ because $k>m$.  
\end{proof}

\subsection{Proof of Equation \eqref{eq:subgradient_phi_m}}
\label{prooftheorem1}

Before introducing the sub-gradient of $\phi_m(S(\theta))$, we provide a very useful characterization of differentiable properties of the optimal value function \cite[Proposition A.22]{bertsekas1971control}, which is also an extension of Danskin's theorem \cite{danskin2012theory}.

\begin{lemma}

Let $\phi:\mathbb{R}^n \times \mathbb{R}^m \rightarrow (-\infty, \infty]$ be a function and let $Y$ be a compact subset of $\mathbb{R}^m$. Assume further that for every vector $y\in Y$ the function $\phi(\cdot,y):\mathbb{R}^n \rightarrow (-\infty, \infty]$ is a closed proper convex function. Consider the function $f$ defined as $f(x)=sup_{y\in Y} \phi(x,y)$, then if $f$ is finite somewhere, it is a closed proper convex function. Furthermore, if int$(dom f) \neq \oldemptyset$ and $\phi$ is continuous on the set int$(dom f) \times Y$, then for every $x\in int(dom f)$ we have $\partial f(x)=conv\{\partial \phi(x,\overline{y})|\overline{y}\in\overline{Y}(x)\}$, where $\overline{Y}(x)$ is the set $\overline{Y}(x) = \{\overline{y}\in Y|\phi(x,\overline{y})=max_{y\in Y} \phi(x,y)\}$

\label{lemma:Bertsekas_lemma} 
\end{lemma}

\textbf{Proof of Equation \eqref{eq:subgradient_phi_m}}

\begin{proof}
We apply Lemma \ref{lemma:Bertsekas_lemma} with a new notation $\phi_m(\theta, \hat{\lambda}) = m\hat{\lambda}+\sum_{i=1}^n[s_i(\theta)-\hat{\lambda}]_+$. Suppose $\theta\in \mathbb{R}^n$ and $\hat{\lambda} \in \mathbb{R}$, the function $\phi_m: \mathbb{R}^n \times \mathbb{R} \rightarrow (-\infty, \infty]$. Let Y be a compact subset of $\mathbb{R}$ and for every $\hat{\lambda} \in Y$, it is obviously that the function $\phi_m(\cdot, \hat{\lambda}): \mathbb{R}^n \rightarrow (-\infty, \infty]$ is a closed proper convex function w.r.t $\theta$ from the second term of Eq.(\ref{eq:0}).

Consider a function $f$ defined as $f(\theta)=\sup_{\hat{\lambda}\in Y} \phi(\theta,\lambda)$, since $f$ is finite somewhere, it is a closed proper convex function. The interior of the effective domain of $f$ is nonempty, and that $\phi_m$ is continuous on the set $int(dom f) \times Y$. The condition of lemma \ref{lemma:Bertsekas_lemma} is satisfied. 

$\forall \theta\in int(dom f)$, we have
\[
\partial f(\theta)=conv\{\partial \phi_m(\theta,\overline{\lambda})|\overline{\lambda}\in\overline{Y}(\theta)\},
\]
where 
\[
\overline{Y}(\theta) = \{\overline{\lambda}\in Y|\phi_m(\theta,\overline{\lambda})=max_{\hat{\lambda}\in Y}\phi_m(\theta,\hat{\lambda})\}=\{\overline{\lambda}\in Y|-\phi_m(\theta,\overline{\lambda})=-min_{\hat{\lambda}\in Y}\phi_m(\theta,\hat{\lambda})\}.
\]
As we know $-min_{\hat{\lambda}\in Y}\phi_m(\theta,\hat{\lambda})= -\phi_m(S(\theta))$. This means the subdifferential of $f$ w.r.t $\theta$ exists when we set the optimal value of $\hat{\lambda}$.

From the above and the lemma \ref{lemma:convex}, we can get the sub-gradient $\hat{\theta} \in \partial \phi_m(S(\theta)) = \sum_{i=1}^n \partial s_i(\theta)\cdot \mathbb{I}_{[s_i(\theta)>\hat{\lambda}]}$, where $\hat{\lambda}$ equals to $s_{[m]}(\theta)$.

\end{proof}

\subsection{Proof of Theorem 2}\label{appendix:proof_theorem_2}
\begin{proof} Without loss of generality,  by normalization we can assume $s(0)=1$ which can be satisfied by scaling. 
For any fixed $x\in \X$, by the definition of $f^* = \arg\inf \L(f, \gl^*, \hgl^*)$, we know that  
$$ f^*(x)= t^* = \arg\inf_{t\in \R}  \mathbb{E}\Bigl[[s(Yt)-\lambda^*]_+ - [s(Yt)-\hat{\lambda}^*]_+ \Big|  X = x \Bigr].$$
Notice the assumption $\hat{\lambda}^* >  \lambda^*$ and recall $\eta(x) = P(y=1|  x).$  We need to show that $t^* >0 $ for $\eta(x)>1/2$ and $t^*<0$ if $\eta(x)<1/2.$ Indeed, if $t^*\neq 0 $, then,  
 by the definition of $t^*$,
we have that 
\begin{align*} 
\mathbb{E}\Bigl[[s(Yt^*)-\lambda^*]_+ - [s(Yt^*)-\hat{\lambda^*}]_+ \Big|  X = x \Bigr] < \mathbb{E}\Bigl[[s(-Yt^*)-\lambda^*]_+ - [s(-Yt^*)-\hat{\lambda}^*]_+  \Big|  X = x \Bigr]     
\end{align*}
The above inequality  is identical to 
$$ \bigl[\big((s(t^*) -\lambda^*)_+ - (s(t^*) -\hat{\lambda}^*)_+\big)  - \big((s(-t^*) -\lambda^*)_+ - (s(-t^*) -\hat{\lambda}^*)_+\big) \bigr] \bigl[2\eta(x)- 1\bigr] <0.$$

Since $\hat{\lambda}^*> \lambda^*$, we have that  that $g(s) =  (s-\lambda^*)_+ - (s -\hat{\lambda}^*)_+ $ is a non-decreasing function of variable $s$. Then, if $\eta(x)> {1\over 2}$ we must have $g(s(t^*))<g(s(-t^*))$ which indicates $s(t^*) < s(-t^*).$ From the non-increasing property of $s$ on $\mathbb{R}$, $s(t)$ is also a convex function and $s'(0)<0$ immediately indicates $t^*>0$. Likewise, we can show that $t^*<0$ for $\eta(x)<1/2.$


To prove $t=0$ is not a minimizer, without loss of generality, assume $\eta(x)>\frac{1}{2}$. We need to consider two conditions as follows,

1. If $0\leq \lambda^* < \hat{\lambda}^*\leq 1$ and $s(0)=1$, then
\begin{align*}
A&=\mathbb{E}\Bigl[[s(0)-\lambda^*]_+ - [s(0)-\hat{\lambda}^*]_+ \Big|  X = x \Bigr] \\
&=[1-\lambda^*]_+ - [1-\hat{\lambda}^*]_+\\
&=\hat{\lambda}^* - \lambda^*
\end{align*}
Since $s^{\prime}(0)<0$ and $s$ is non-increasing, there exists $t^0>t^*=0>-t^0$, and $s(-t^0)>s(0)\geq \hat{\lambda}^*>s(t^0)>\lambda^*$. Let
\begin{align*} 
B&=\mathbb{E}\Bigl[[s(Yt^0)-\lambda^*]_+ - [s(Yt^0)-\hat{\lambda}^*]_+ \big|  X= x \Bigr]\\
&=\Big([s(t^0)-\lambda^*]_+ - [s(t^0)-\hat{\lambda}^*]_+ \Big) \eta(x) +  \Big([s(-t^0)-\lambda^*]_+ - [s(-t^0)-\hat{\lambda}^*]_+ \Big) \Big(1-\eta(x)\Big)\\
&=\Big([s(-t^0)-\lambda^*]_+ - [s(-t^0)-\hat{\lambda}^*]_+ \Big)\\
&+\Bigl[\Big([s(t^0)-\lambda^*]_+ - [s(t^0)-\hat{\lambda}^*]_+ \Big)-\Big([s(-t^0)-\lambda^*]_+ - [s(-t^0)-\hat{\lambda}^*]_+ \Big)\Bigr]\eta(x)\\
&=\hat{\lambda}^* - \lambda^*+\Bigl[s(t^0)-\lambda^*-(\hat{\lambda}^* - \lambda^*)\Bigr]\eta(x)
\end{align*}
Then
\begin{align*}
B-A=(s(t^0)-\hat{\lambda}^*)\eta(x)<0
\end{align*}
Therefore, $t=0$ is not a minimizer.

2. If $0\leq \lambda^* \leq 1 < \hat{\lambda}^*$ and $s(0)=1$, then
\begin{align*}
&\frac{d}{dt}\mathbb{E}[[s(Yt)-\lambda^*]_+ - [s(Yt)-\hat{\lambda}^*]_+]|_{t=0} \\
&=\frac{d}{dt}[\eta(x)([s(t)-\lambda^*]_+ - [s(t)-\hat{\lambda}^*]_+)+(1-\eta(x))([s(-t)-\lambda^*]_+ - [s(-t)-\hat{\lambda}^*]_+)]|_{t=0}\\
&=\frac{d}{dt}[\eta(x)(s(t)-\lambda^*)+(1-\eta(x))(s(-t)-\lambda^*)]|_{t=0}\\
&=[\eta(x)s^{\prime}(t)-(1-\eta(x))s^{\prime}(-t)]|_{t=0}\\
&=(2\eta(x)-1)s^{\prime}(0)<0
\end{align*}
Thus $t = 0$ is not a minimizer. 
\end{proof}

\subsection{Proof of Proposition 1}
\label{appendix:proof_proposition1}
\begin{proof}
We just need to prove that $\max_{y \not \in Y}\theta_{y}^\top x \ge \theta_{[|Y|+1]}^\top x$. If this is not the case, then for any label $y \not \in Y$, then its rank in the ranked list is no more than $|Y|+2$, then the sum of total number of such labels is not larger than $l - (|Y|+2) + 1 = l - |Y|-1$. And the total number of labels will be $|Y|+|\{y\not\in Y\}| \le l-1 \not = l$, which is a contradiction. 
\end{proof}

\section{Additional Experimental Details}

\subsection{Source Code}
For the purpose of review, the source code and datasets are accessible at supplementary file.

\subsection{Computing Infrastructure Description}
All algorithms are implemented in Python 3.6 and trained and tested on an Intel(R) Xeon(R) CPU W5590 @3.33GHz with 48GB of RAM.

\subsection{Time Complexity Analyze}
We consider the average case in the time complexity analyze. For a given outer loop size $|t|$, a inner loop size $|l|$, and training sample size $n$, the complexity of our MSoRR Algorithm \ref{Alg0} is $O(|t|(n\log n+|l|))$.

\subsection{Training Settings on Toy Examples for Aggregate Loss}
To reproduce the experimental results of \aorr~on synthetic data, we provide the details about the settings when we are training the model in Table \ref{tab:aggregate_synthetic_datasets_settings}. For example, the learning rate, the number of epochs for the outer loop, and the number of epochs for the inner loop. 

\begin{table}[h]
\captionsetup{font=footnotesize}
\centering
\begin{tabular}{c|c|c|c|c||c|c|c}
\hline
\multirow{2}{*}{Datasets} & \multirow{2}{*}{Outliers} & \multicolumn{3}{c||}{Logistic loss} & \multicolumn{3}{c}{Hinge loss} \\ \cline{3-8} 
                 & &LR    &   \# OE   &   \# IE    &   LR    &   \# OE    &    \# IE    \\ \hline
\multirow{7}{*}{\makecell{Multi-modal data}} &1 &  0.01     &  100     &   1000    &     0.01  &    5   &  1000     \\ \cline{2-8} 
                  &   2 &0.01&  100     &  1000     &   0.01    &   5    &  1000     \\ \cline{2-8} 
                  &   3 &0.01&   100    &  1000     &  0.01     &   5    & 1000      \\ \cline{2-8} 
                  &   4 &0.01&   100    &  1000     &   0.01    &  5     &   1000    \\ \cline{2-8} 
                  &   5 &0.01&   100    &   1000    &   0.01    &  5     &  1000     \\ \cline{2-8} 
                  &   10&0.01&   100    &  1000     &  0.01     &   5    &  1000     \\ \cline{2-8} 
                  &  20 &0.01&   100    &  1000     &   0.01    &  5     &  1000     \\ \hline
 \makecell{Imbalanced  data} &1 &   0.01    &   100    &   1000    &  0.01     & 5      &  1000     \\ \hline
\end{tabular}
\begin{tablenotes}\scriptsize
\centering
\item[*] LR: Learning Rate, OE: Outer Epochs, IE: Inner epochs
\end{tablenotes}
\vspace{1mm}
\caption{\small \em \aorr~ settings on toy experiments.}
\label{tab:aggregate_synthetic_datasets_settings}
\vspace{-5mm}
\end{table}

\subsection{Description of Datasets for Aggregate Loss}
\label{aggregate_loss_datasets}
In aggregate loss experiments, for real-world datasets, we use five benchmark datasets from the UCI and the KEEL data repositories. The details of these datasets are shown in Table \ref{tab:aggregate_real_datasets}.

\begin{table}[h]
\centering
\begin{tabular}{c|cccc}
\hline
        Datasets   &  \#Classes   &    \#Samples &  \#Features   & Class Ratio  \\ \hline
        Monk        &  2   & 432 & 6 & 1.12\\ 
        Australian  &  2   & 690 & 14 & 1.25\\ 
        Phoneme     &  2   & 5,404 & 5 & 2.41\\ 
        Titanic     &  2   & 2,201 & 3 & 2.10\\ 
        Splice      &  2   & 3,175 & 60 & 1.08\\ \hline
\end{tabular}
\vspace{1mm}
\caption{\small \em Statistical information of each dataset for aggregate loss.}
\label{tab:aggregate_real_datasets}
\vspace{-5mm}
\end{table}

\subsection{Training Settings on Real Datasets for Aggregate Loss}
We provide a reference for setting parameters to reproduce our \aorr~ experiments on real datasets. Table \ref{tab:aggregate_real_datasets_logistic_settings} contains the settings for individual logistic loss. Table \ref{tab:aggregate_real_datasets_hinge_settings} is for individual hinge loss.
 
\begin{table}[H]
\centering
\begin{tabular}{c|cccccc}
\hline
        Datasets   &  $k$   &  $m$ & $C$ &  \# Outer epochs  & \# Inner epochs & Learning rate \\ \hline
        Monk        & 70    & 20 & $10^4$ & 5 & 2000 & 0.01\\ 
        Australian  & 80    & 3 & $10^4$ & 10 & 1000 & 0.01\\ 
        Phoneme     & 1400  & 100 & $10^4$& 10 & 1000 &0.01\\ 
        Titanic     & 500   & 10  &$10^4$ & 10 & 1000 &0.01\\ 
        Splice      & 450   & 50 &$10^4$ & 10 & 1000 &0.01\\ \hline
\end{tabular}
\vspace{1mm}
\caption{\small \em \aorr~ settings on real datasets for individual logistic loss.}
\label{tab:aggregate_real_datasets_logistic_settings}
\end{table}

\begin{table}[h]
\centering
\begin{tabular}{c|cccccc}
\hline
        Datasets   &  $k$   &  $m$ & $C$ &  \# Outer epochs  & \# Inner epochs & Learning rate \\ \hline
        Monk        & 70    & 45 & $10^4$ & 5 & 1000 & 0.01\\ 
        Australian  & 80    & 3 & $10^4$ & 5 & 1000 & 0.01\\ 
        Phoneme     & 1400  & 410 & $10^4$& 10 & 500 &0.01\\ 
        Titanic     & 500   & 10  &$10^4$ & 5 & 500 &0.01\\ 
        Splice      & 450   & 50 &$10^4$ & 10 & 1000 &0.01\\ \hline
\end{tabular}
\vspace{1mm}
\caption{\small \em \aorr~ settings on real datasets for individual hinge loss.}
\label{tab:aggregate_real_datasets_hinge_settings}
\vspace{-5mm}
\end{table}

\subsection{Description of Datasets for Multi-label Learning}
\label{Multi-label_datasets}
In multi-label learning experiments, we conduct experiments on three benchmark datasets (Emotions, Scene and Yeast) from the KEEL data repository. The details of them as described in Table \ref{tab:multi_label_datasets}.

\begin{table}[h]
\centering
\begin{tabular}{c|cccc}
\hline
        Datasets   &  \#Samples   &    \#Features &  \#Labels   & $\overline{c}$  \\ \hline
        Emotions   &  593   & 72 & 6 & 1.81\\ 
        Scene      &  2,407 & 294 & 6 & 1.06\\ 
        Yeast      &  2,417 & 103 & 14 & 4.22\\ \hline
\end{tabular}
\vspace{1mm}
\caption{\small \em Statistical information of each dataset for multi-label learning, where $\overline{c}$ represents the average number of positive labels per instance.}
\label{tab:multi_label_datasets}
\vspace{-5mm}
\end{table}

\subsection{Training Settings for Multi-label Learning}
\label{sec:multi_label_settings}
The settings for \tkml~ on three real datasets are shown in Table \ref{tab:multi_label_settings}. 

\begin{table}[h]
\centering
\begin{tabular}{c|cccc}
\hline
        Datasets   &  $C$   &    \#Outer epochs &  \#Inner epochs   & Learning rate  \\ \hline
        Emotions   &  $10^4$   & 20 & 1000 & 0.1\\ 
        Scene      &  $10^4$ & 20 & 1000 & 0.1\\ 
        Yeast      &  $10^4$ & 20 & 1000 & 0.1\\ \hline
\end{tabular}
\vspace{1mm}
\caption{\small \em \tkml~ settings on each dataset.}
\label{tab:multi_label_settings}
\vspace{-5mm}
\end{table}

\subsection{Training Settings for Multi-class Learning}
Training settings for the MNIST dataset in different noise level can be found in Table \ref{tab:multi_class_settings}.

\begin{table}[H]
\centering
\begin{tabular}{c|ccc}
\hline
        Noise level   & \#Outer epochs &  \#Inner epochs   & Learning rate  \\ \hline
        0.2   &  27 & 2000 & 0.1\\ 
        0.3      &  25 & 2000 & 0.1\\ 
        0.4      &  21 & 2000 & 0.1\\ \hline
\end{tabular}
\vspace{1mm}
\caption{\small \em \tkml~ settings on the MNIST dataset in different noise levels.}
\label{tab:multi_class_settings}
\vspace{-5mm}
\end{table}

\section{Additional Experimental Results}

\subsection{Toy Examples with More Outliers for Effects of Aggregate Losses}
\label{additional_exp_more_outliers}
\begin{figure*}[t]
\captionsetup[subfigure]{justification=centering}
\centering
        \begin{subfigure}[b]{0.24\textwidth}
                \includegraphics[width=\linewidth]{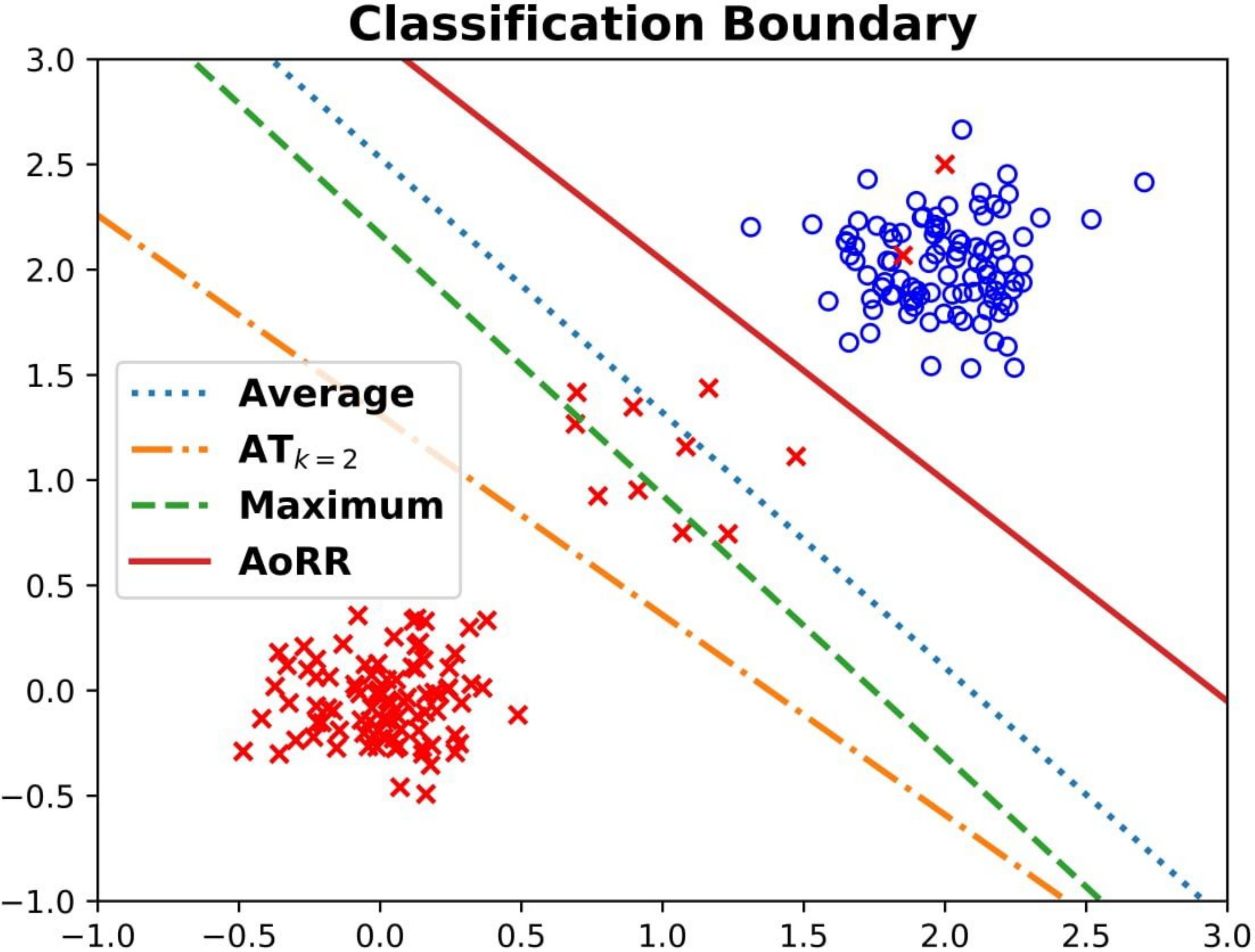}
                \caption{}
                \label{fig:LogisticRegression_data4_2}
        \end{subfigure}%
        \begin{subfigure}[b]{0.24\textwidth}
                \includegraphics[width=\linewidth]{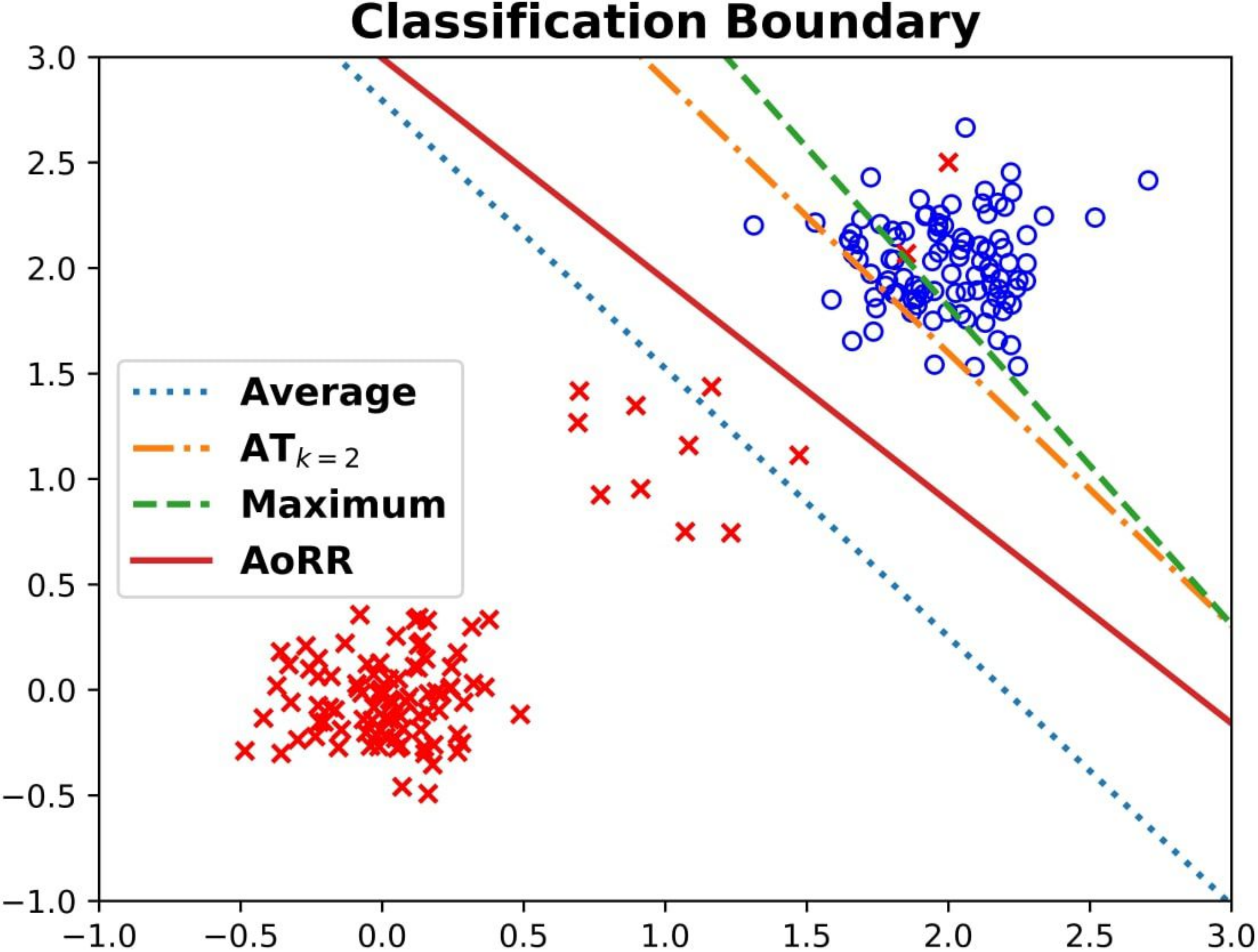}
                \caption{}
                \label{fig:Hinge_data4_2}
        \end{subfigure}%
        \rulesep
        \begin{subfigure}[b]{0.24\textwidth}
                \includegraphics[width=\linewidth]{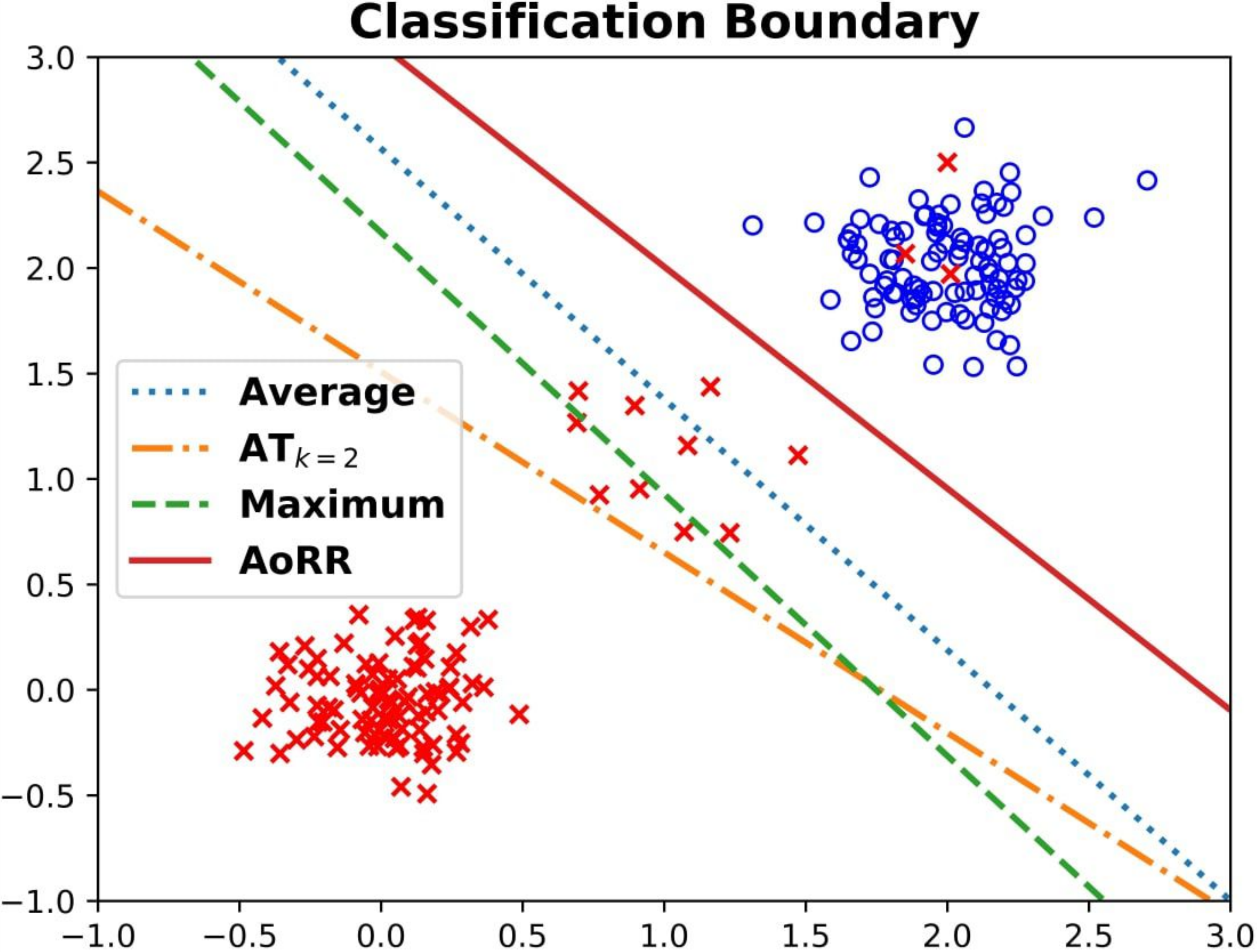}
                \caption{}
                \label{fig:LogisticRegression_data4_3}
        \end{subfigure}%
        \begin{subfigure}[b]{0.24\textwidth}
                \includegraphics[width=\linewidth]{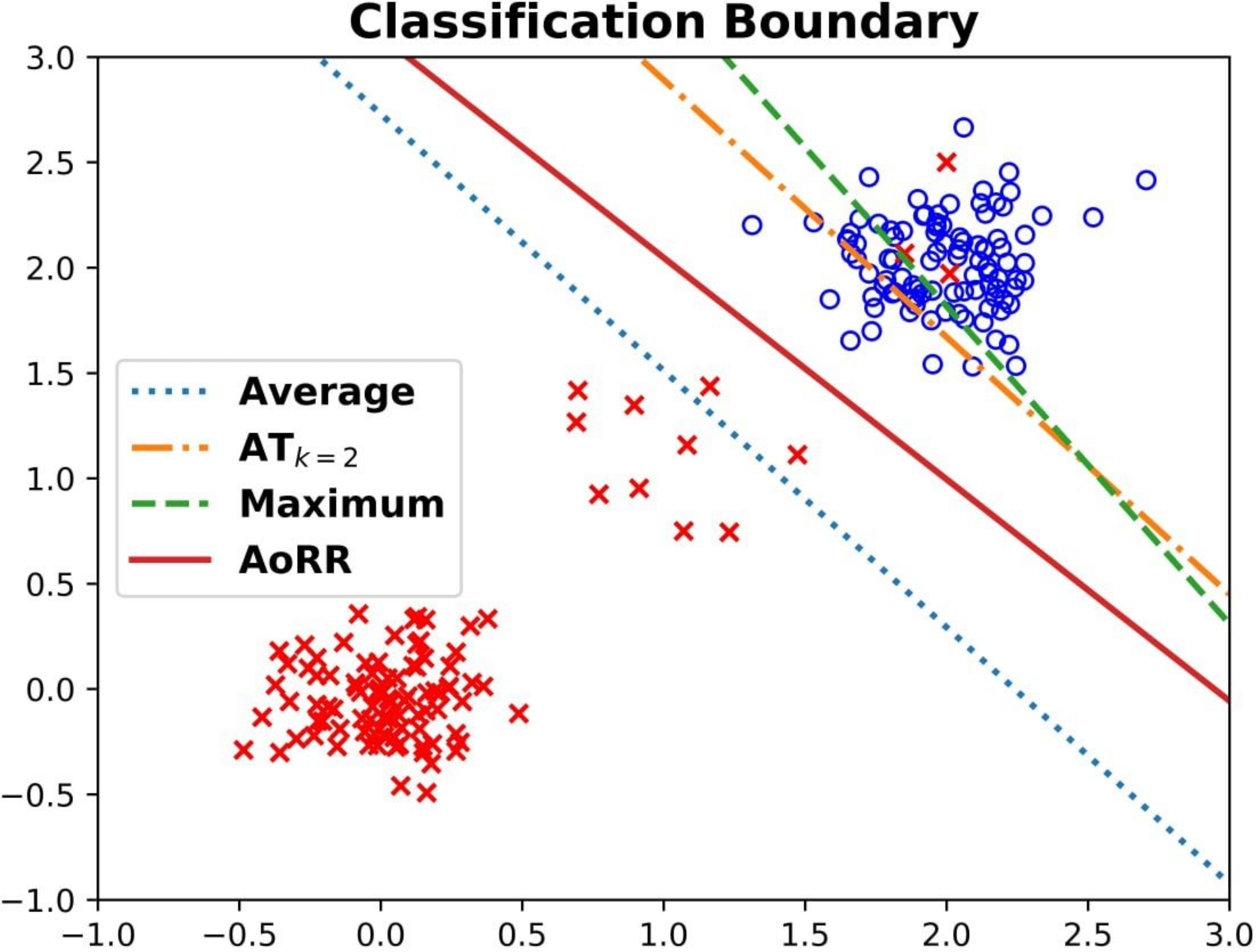}
                \caption{}
                \label{fig:Hinge_data4_3}
        \end{subfigure}
        \bigskip
        \begin{subfigure}[b]{0.24\textwidth}
                \includegraphics[width=\linewidth]{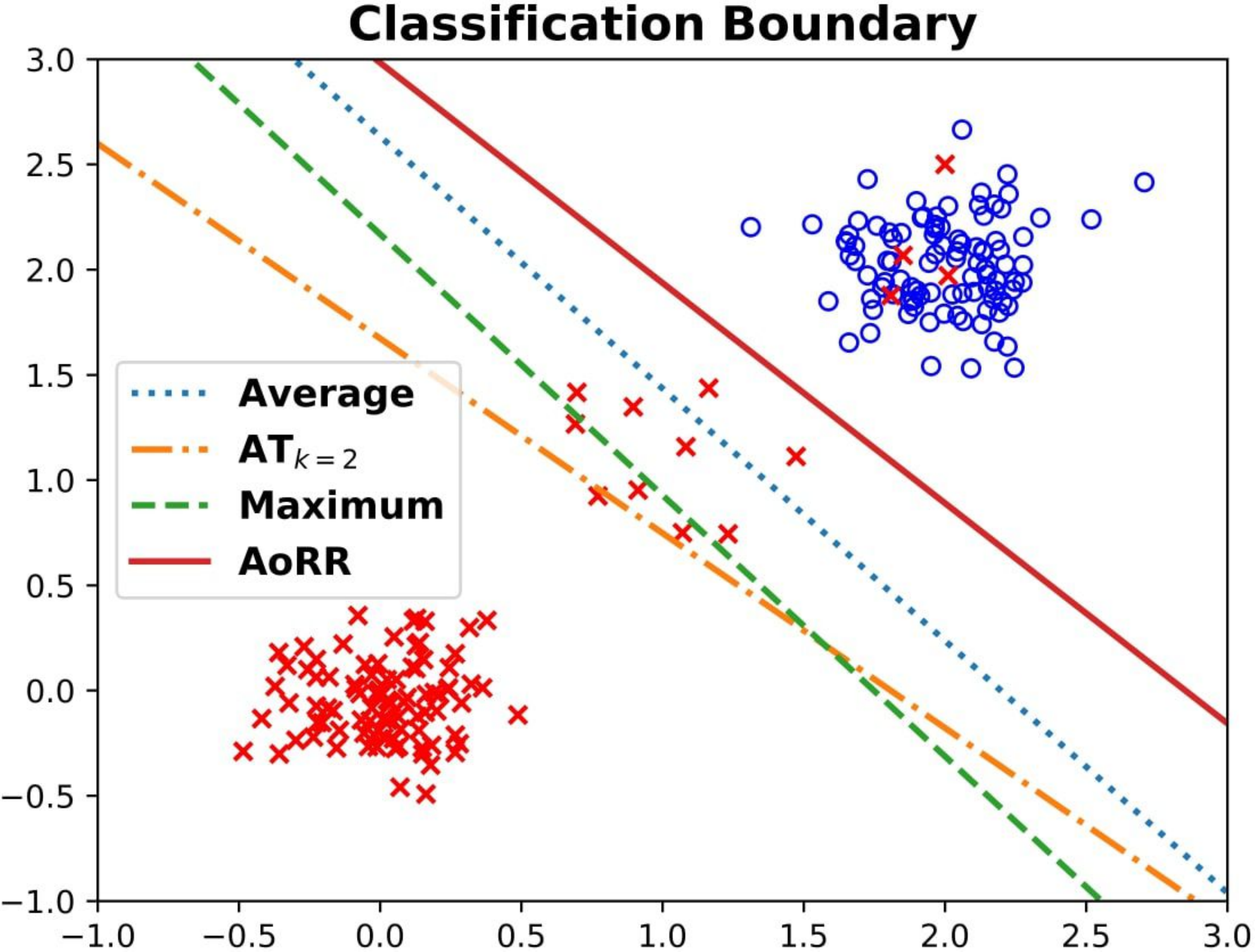}
                \caption{}
                \label{fig:LogisticRegression_data4_4}
        \end{subfigure}%
        \begin{subfigure}[b]{0.24\textwidth}
                \includegraphics[width=\linewidth]{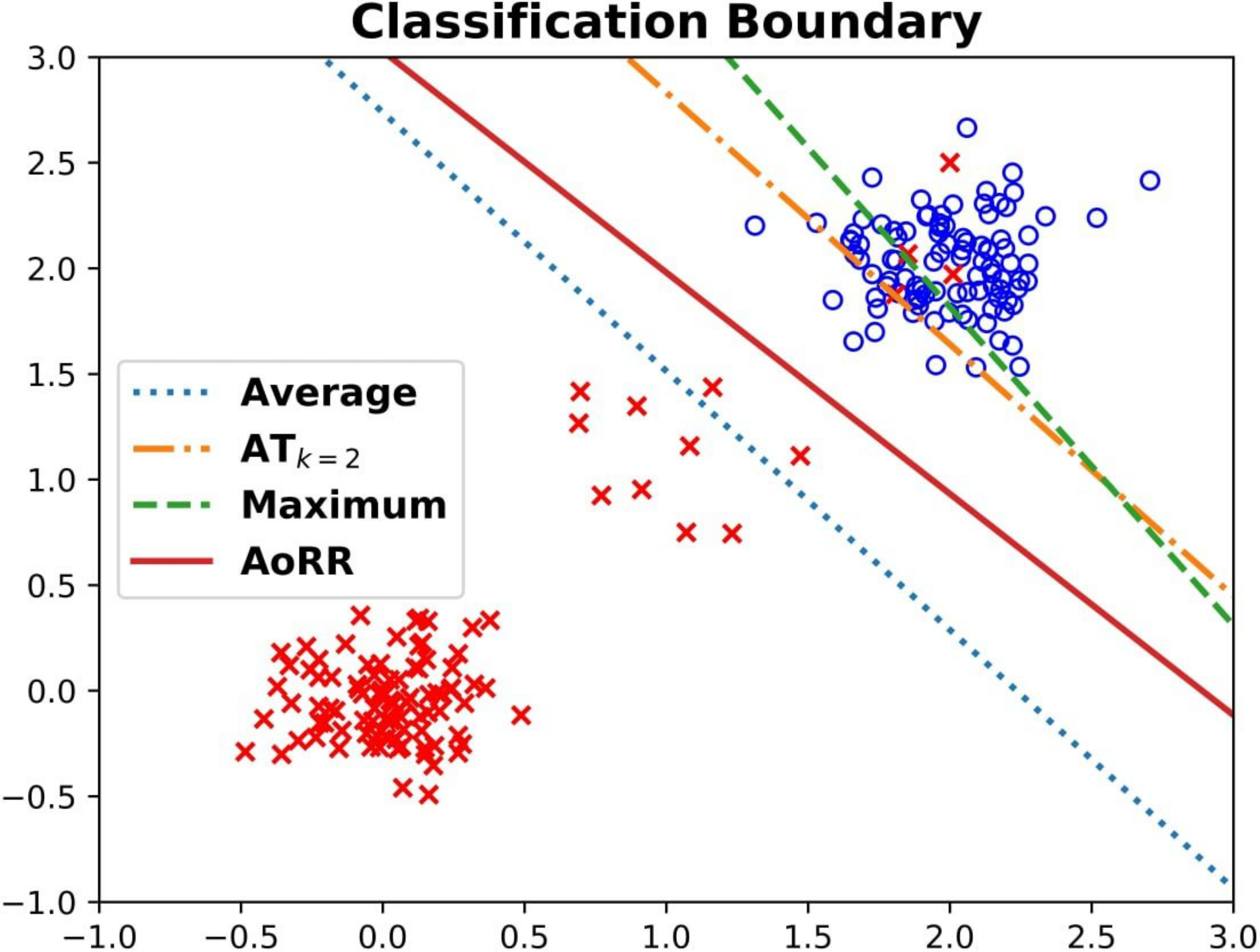}
                \caption{}
                \label{fig:Hinge_data4_4}
        \end{subfigure}%
        \rulesep
        \begin{subfigure}[b]{0.24\textwidth}
                \includegraphics[width=\linewidth]{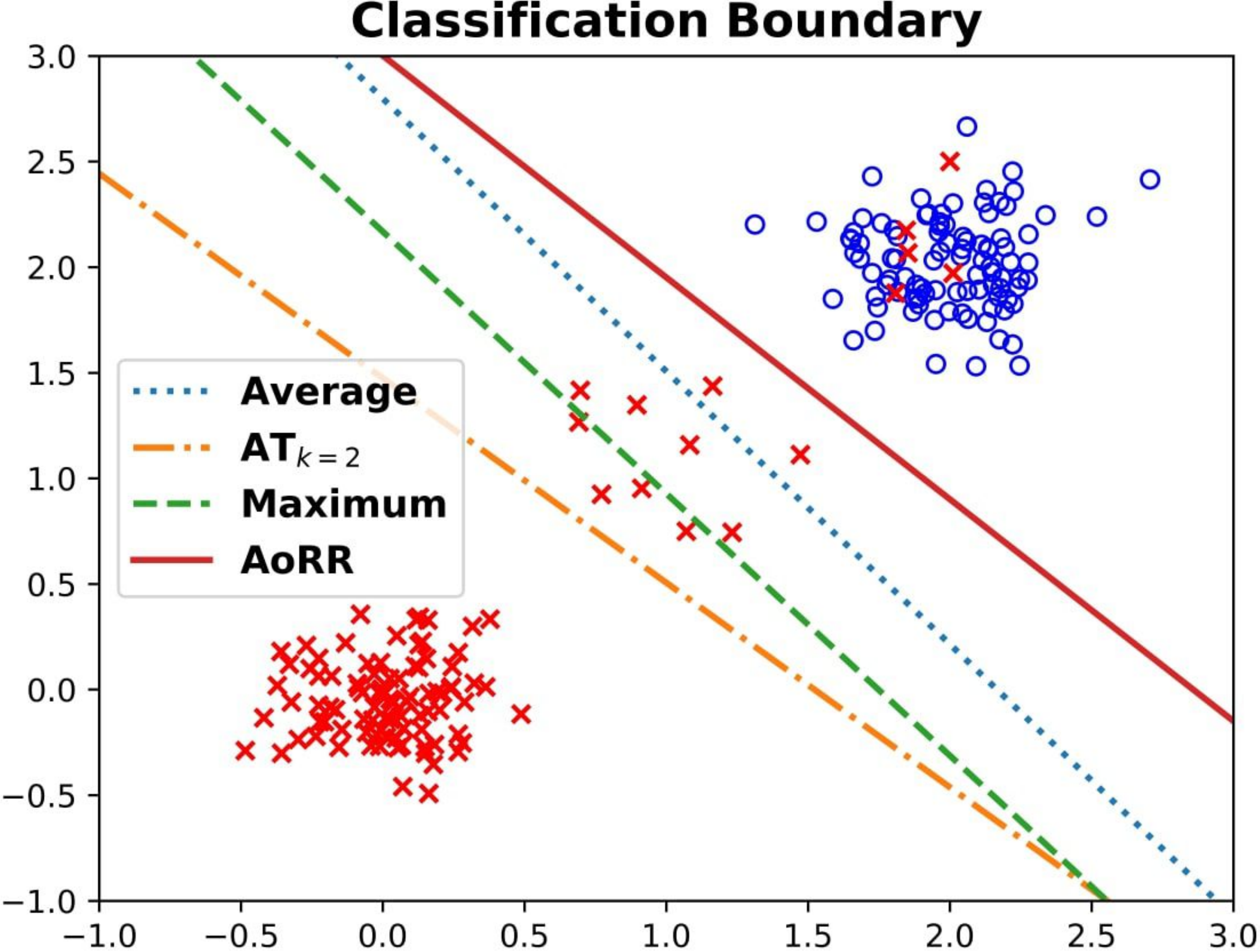}
                \caption{}
                \label{fig:LogisticRegression_data4_5}
        \end{subfigure}%
        \begin{subfigure}[b]{0.24\textwidth}
                \includegraphics[width=\linewidth]{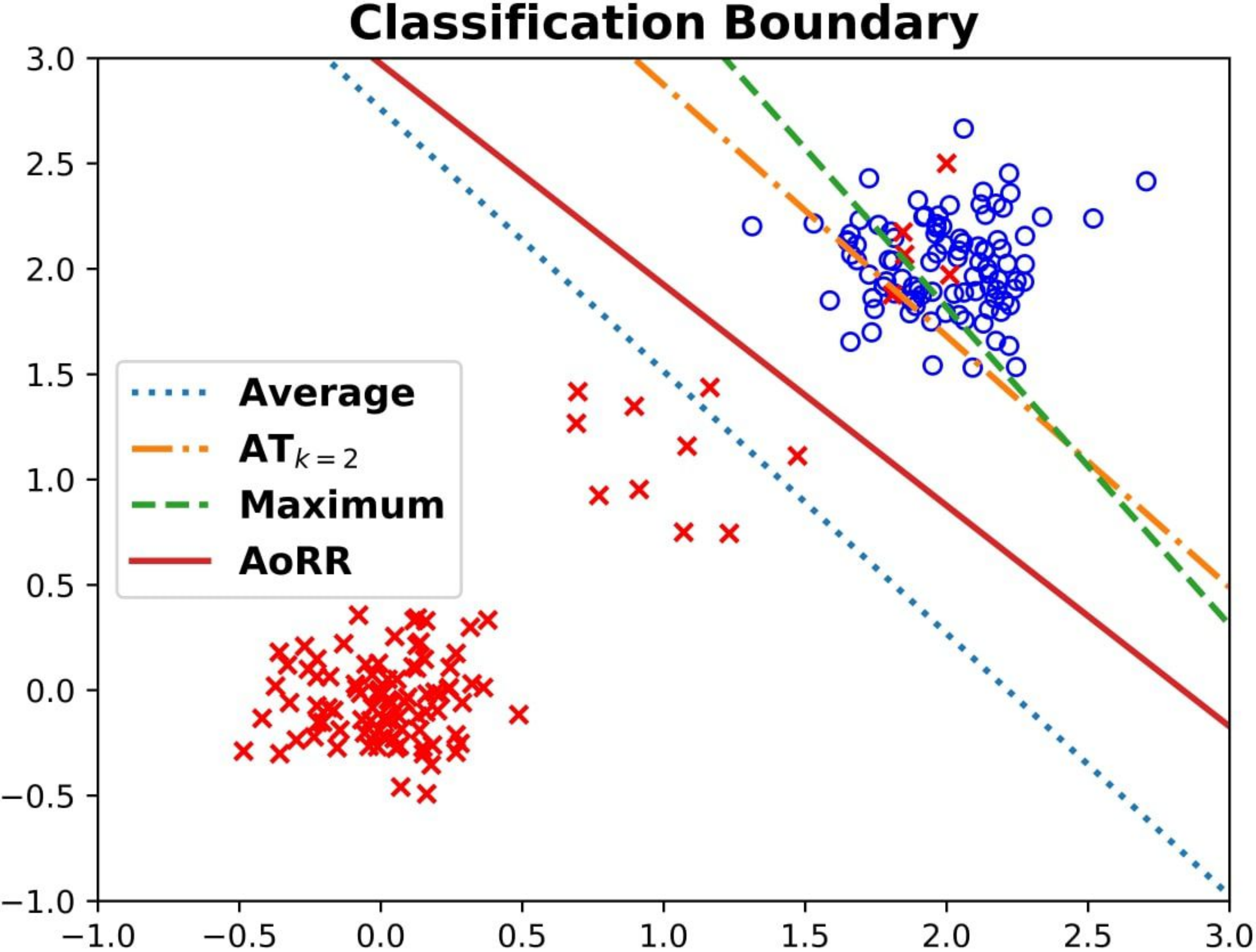}
                \caption{}
                \label{fig:Hinge_data4_5}
        \end{subfigure}
        \bigskip
        \begin{subfigure}[b]{0.24\textwidth}
                \includegraphics[width=\linewidth]{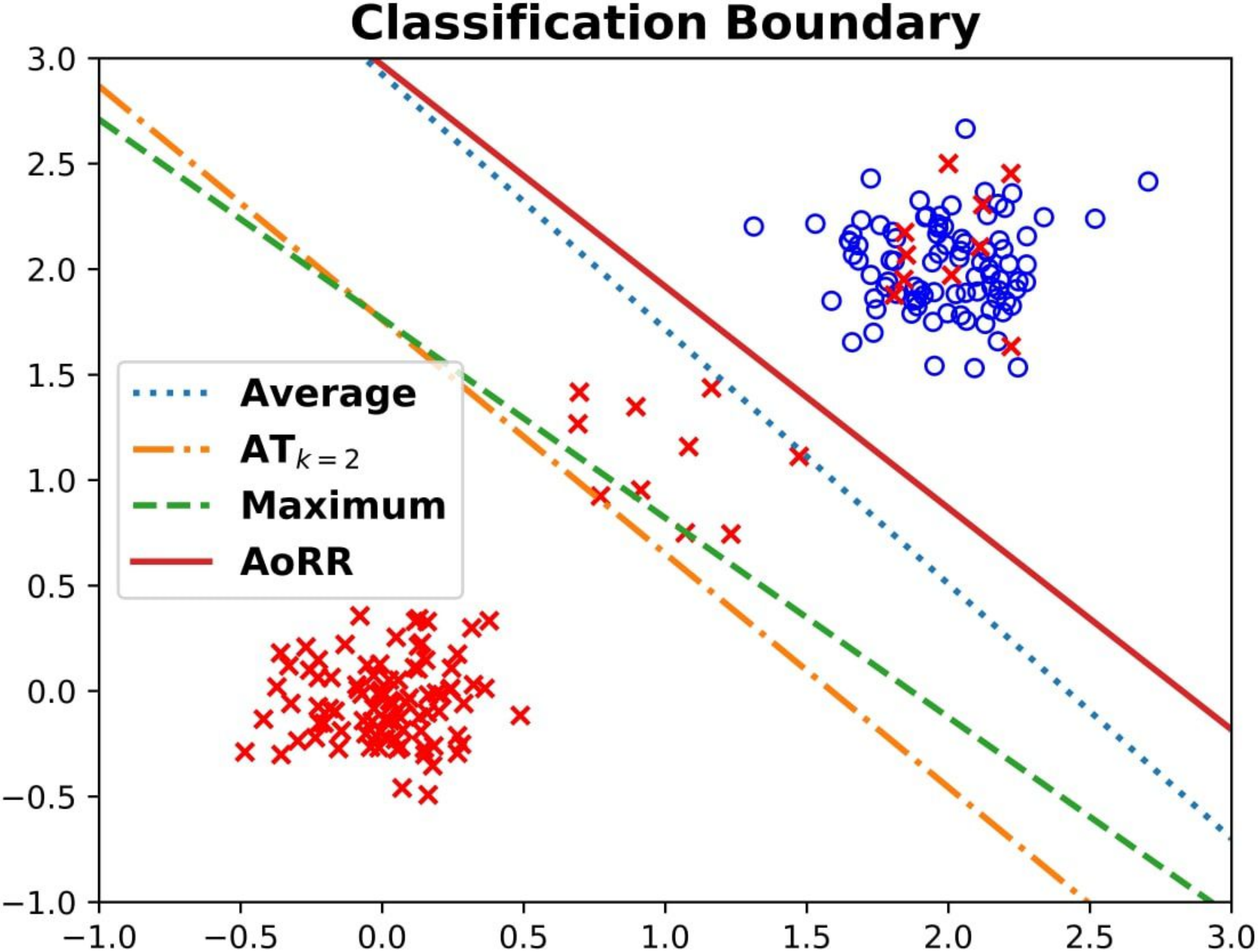}
                \caption{}
                \label{fig:LogisticRegression_data4_10}
        \end{subfigure}%
        \begin{subfigure}[b]{0.24\textwidth}
                \includegraphics[width=\linewidth]{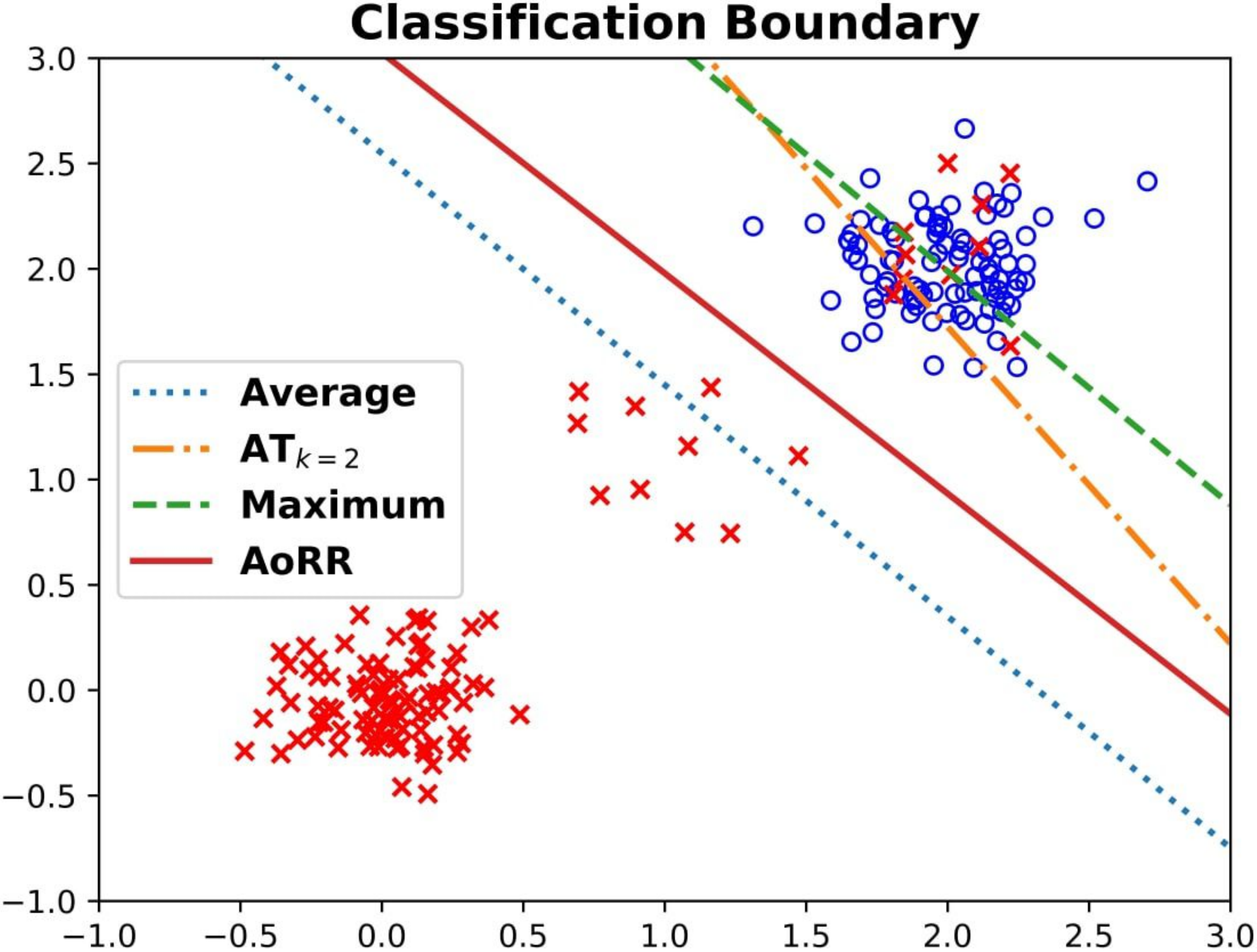}
                \caption{}
                \label{fig:Hinge_data4_10}
        \end{subfigure}%
        \rulesep
        \begin{subfigure}[b]{0.24\textwidth}
                \includegraphics[width=\linewidth]{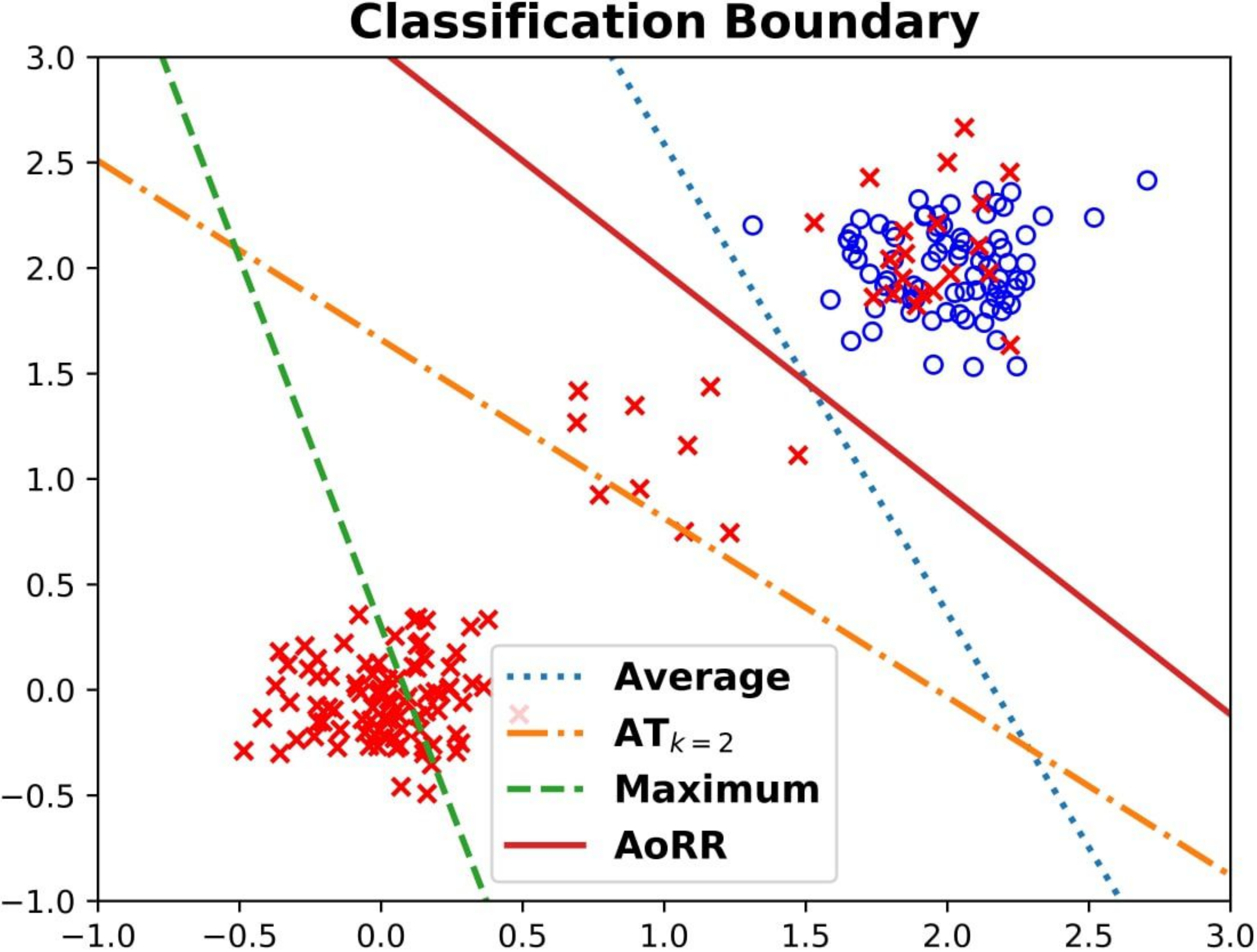}
                \caption{}
                \label{fig:LogisticRegression_data4_20}
        \end{subfigure}%
        \begin{subfigure}[b]{0.24\textwidth}
                \includegraphics[width=\linewidth]{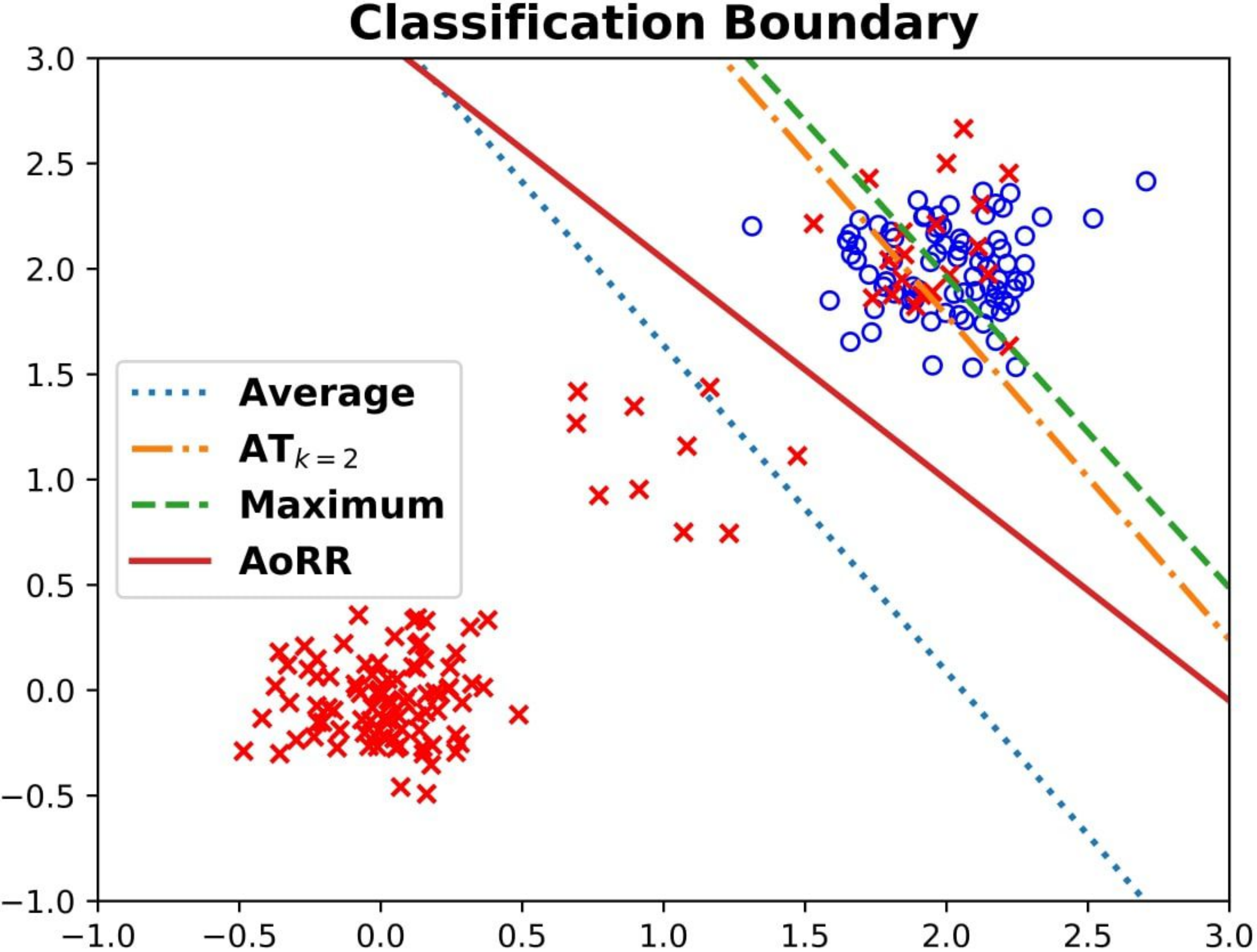}
                \caption{}
                \label{fig:Hinge_data4_20}
        \end{subfigure}
        \caption{\small \em Comparison of different aggregate losses on 2D synthetic data with 200 samples for binary classification with individual logistic loss (a, c, e, g, i, k) and individual hinge loss (b, d, f, h, j, l). Outliers are shown as $\times$ in blue class.}\label{fig: additional_toy_examples}
\vspace{-5mm}
\end{figure*}

In order to evaluate the effects of different aggregate losses on more than one outlier, we also conducted additional experiments on a multi-modal toy example with outliers. We use Gaussian distributions with the different mean and standard deviations to generate this dataset (Fig.\ref{fig: additional_toy_examples}). It contains 200 samples and is distributed in 2 classes (100 samples in red class and 100 samples in blue class). The red samples are sampled from two distributions (primary distribution and minor distribution). The blue samples are sampled from only one distribution. However, they can still be separated. A linear classifier is considered and different aggregate losses are evaluated in individual logistic loss (i.e., Fig.\ref{fig: additional_toy_examples} (a), (c), (e), (g), (i), (k)) and individual hinge loss (i.e., Fig.\ref{fig: additional_toy_examples} (b), (d), (f), (h), (j), (l)). Given a number $n$, we set outliers as replacing $n$ blue samples class with the opposite class. The outliers have been shown as $\times$ in blue class. For AT$_k$ and \aorr~losses, we let the value of $k$ be the same and equals to $n+1$. Let the value of $m$ equals to $n$ in \aorr~loss.  We consider six cases as follows, 

\textbf{Case 1 (2 outliers).} In Fig.\ref{fig: additional_toy_examples} (a) and (b), there exist two outliers. Let hyper-parameters $k=3$ and $m=2$.  

\textbf{Case 2 (3 outliers).} Fig.\ref{fig: additional_toy_examples} (c) and (d) contain three outliers. In this scenario, $k=4$ and $m=3$.  

\textbf{Case 3 (4 outliers).} Fig.\ref{fig: additional_toy_examples} (e) and (f) include four outliers and we set $k=5$ and $m=4$.   

\textbf{Case 4 (5 outliers).} There are five outliers in Fig.\ref{fig: additional_toy_examples} (g) and (h). We set $k=6$ and $m=5$.

\textbf{Case 5 (10 outliers).} Ten outliers have been included in Fig.\ref{fig: additional_toy_examples} (i) and (j). Let $k=11$ and $m=10$ in this case.

\textbf{Case 6 (20 outliers).} We create twenty outliers in Fig.\ref{fig: additional_toy_examples} (k) and (l) and make $k=21$ and $m=20$.

See from case 1, 2, 3, 4, the linear classifier learned from average aggregate loss cross some red samples from minor distribution even though the data is separable. The reason is that the samples close to the decision boundary are sacrificed to reduce the total loss over the whole dataset. 

Since the $k$ value is set to be $n+1$, the AT$_k$ loss select $k$ largest individual losses which contain many outliers to train the classifier. It leads to the instability of the learned classifier. This phenomenon can be found when we compare all cases. Similarly, the maximum aggregate loss cannot fit this data very well in all cases. This loss is very sensitive to outliers.

From cases 5 and 6, the average aggregate loss with individual logistic loss achieves better results than with individual hinge loss. A possible reason is that for correctly classified samples with a margin greater than 1, the penalty caused by hinge loss is 0. However, it is non-zero when using logistic loss. Since many outliers in blue class, to reduce the average loss, the decision boundary will close to blue class. Especially, when we compare (i) and (k), it is obvious that average loss can achieve a better result while the number of outliers is increasing. 

As we discussed, hinge loss has less penalty for correctly classified samples than logistic loss. This causes outliers to be more prominent than normal samples while using the individual hinge loss. This analysis can be verified in the experiment when we compare the individual logistic loss and the individual hinge loss. For example, (i) and (j), (k) and (l), etc.. We find the decision boundaries of maximum loss and AT$_k$ loss are close to outliers in the individual hinge loss scenario because both of them are sensitive to outliers in our cases.

\subsection{Additional Tendency Curves for Effects of Aggregate Losses}
\label{Additional Tendency Curves}
\begin{figure*}[ht]
\captionsetup[subfigure]{justification=centering}
\centering
        \begin{subfigure}[b]{0.256\textwidth}
                \includegraphics[width=\linewidth]{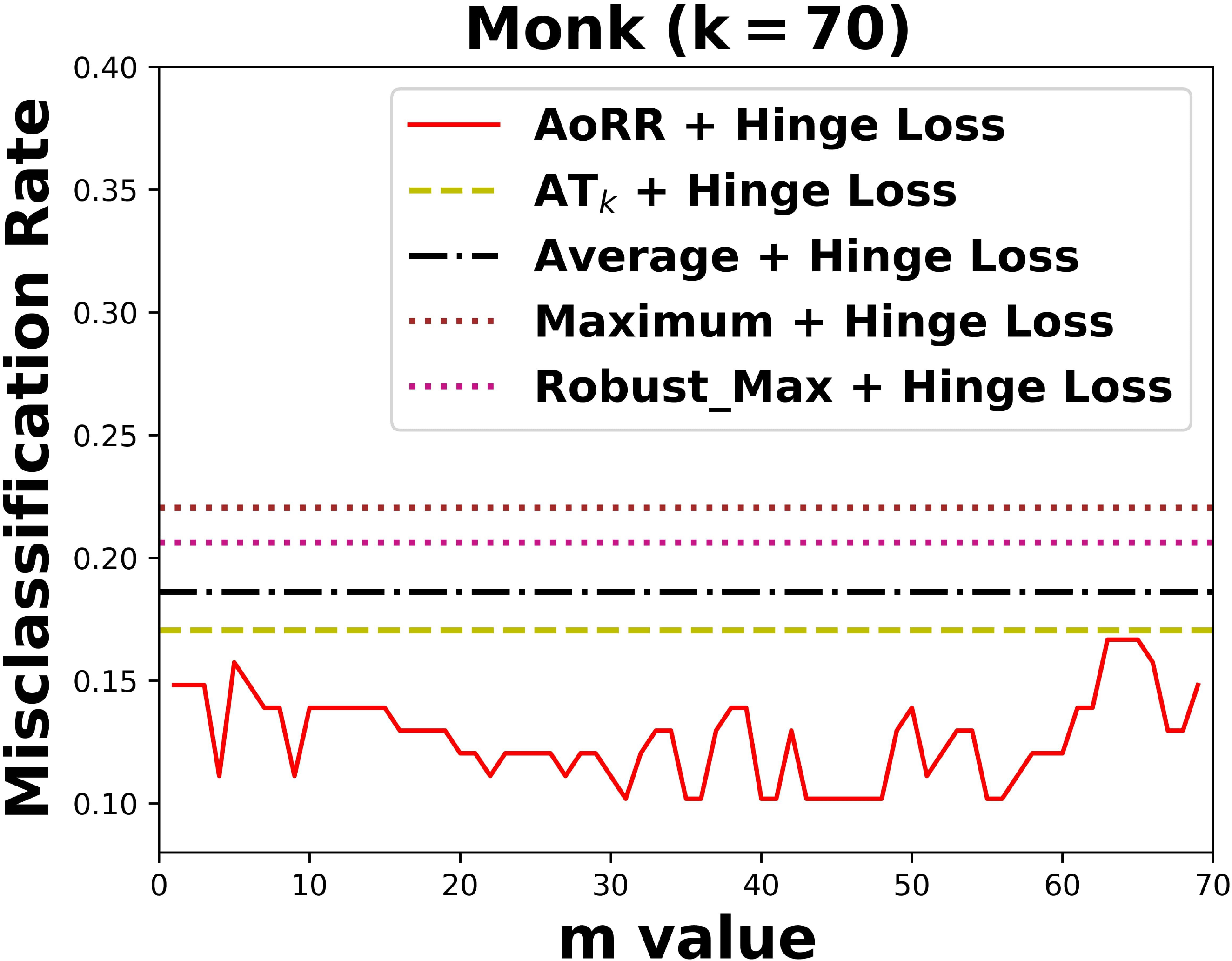}
        \end{subfigure}%
        \begin{subfigure}[b]{0.243\textwidth}
                \includegraphics[width=\linewidth]{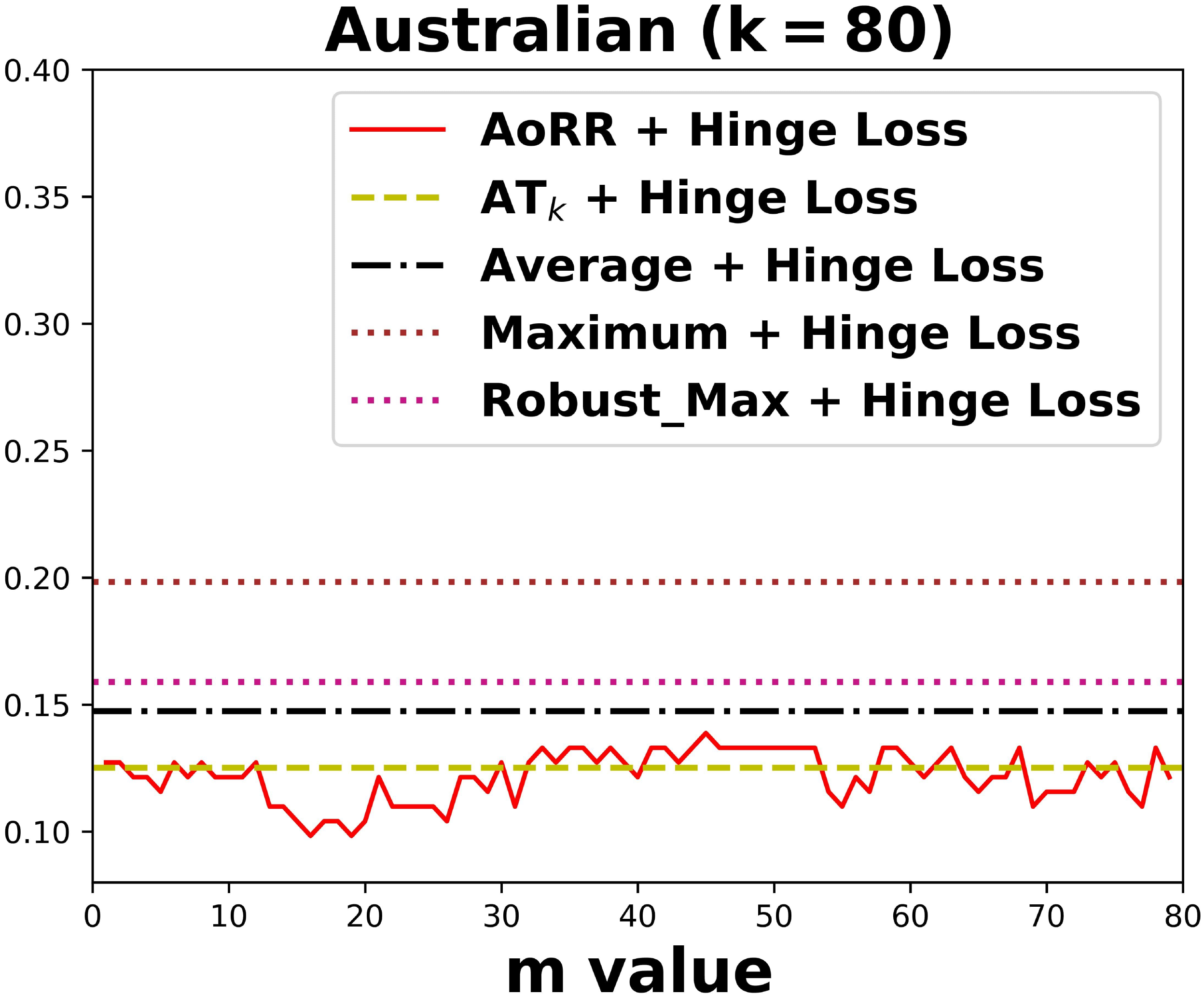}
        \end{subfigure}%
        \begin{subfigure}[b]{0.245\textwidth}
                \includegraphics[width=\linewidth]{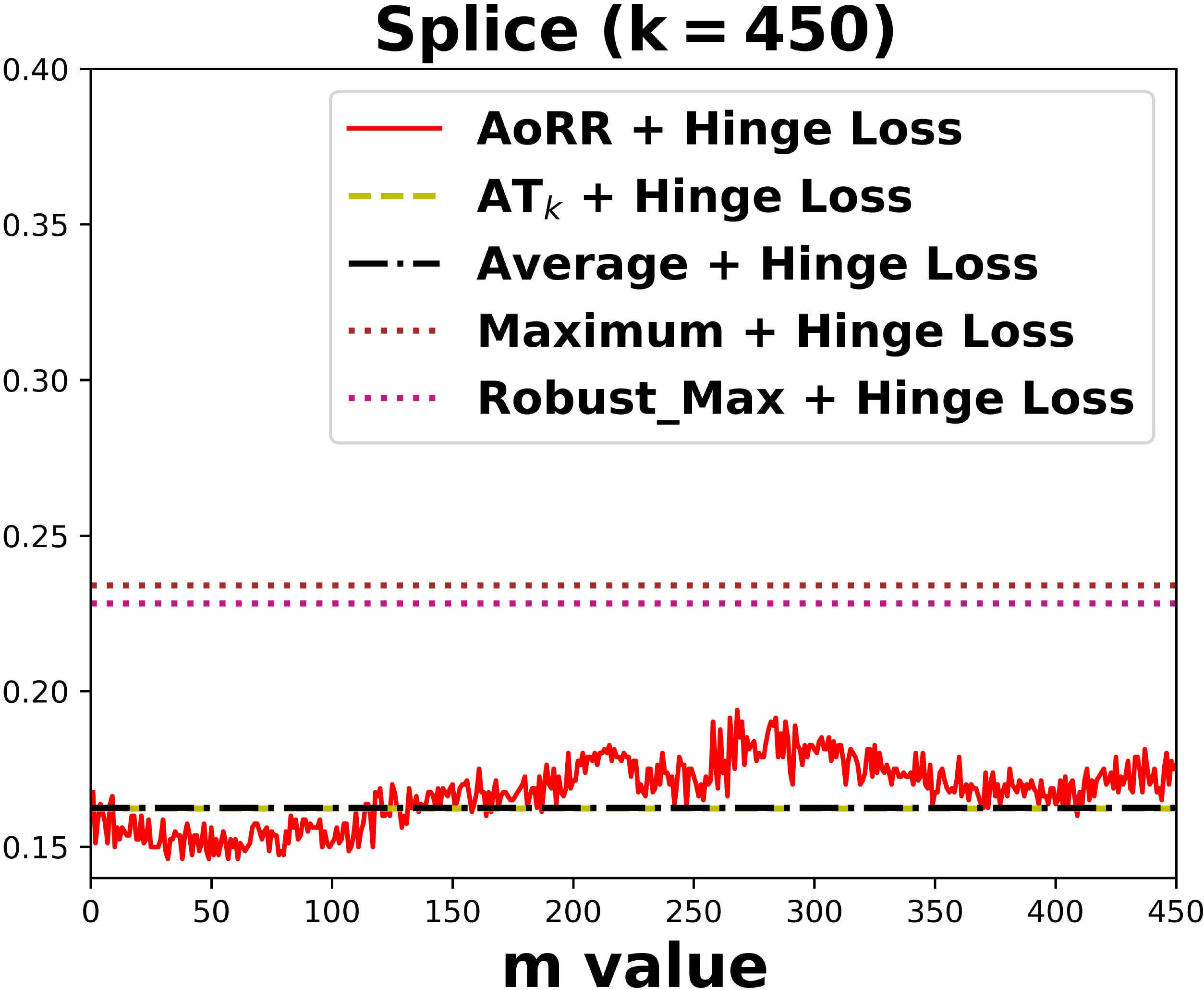}
        \end{subfigure}%
        \begin{subfigure}[b]{0.251\textwidth}
                \includegraphics[width=\linewidth]{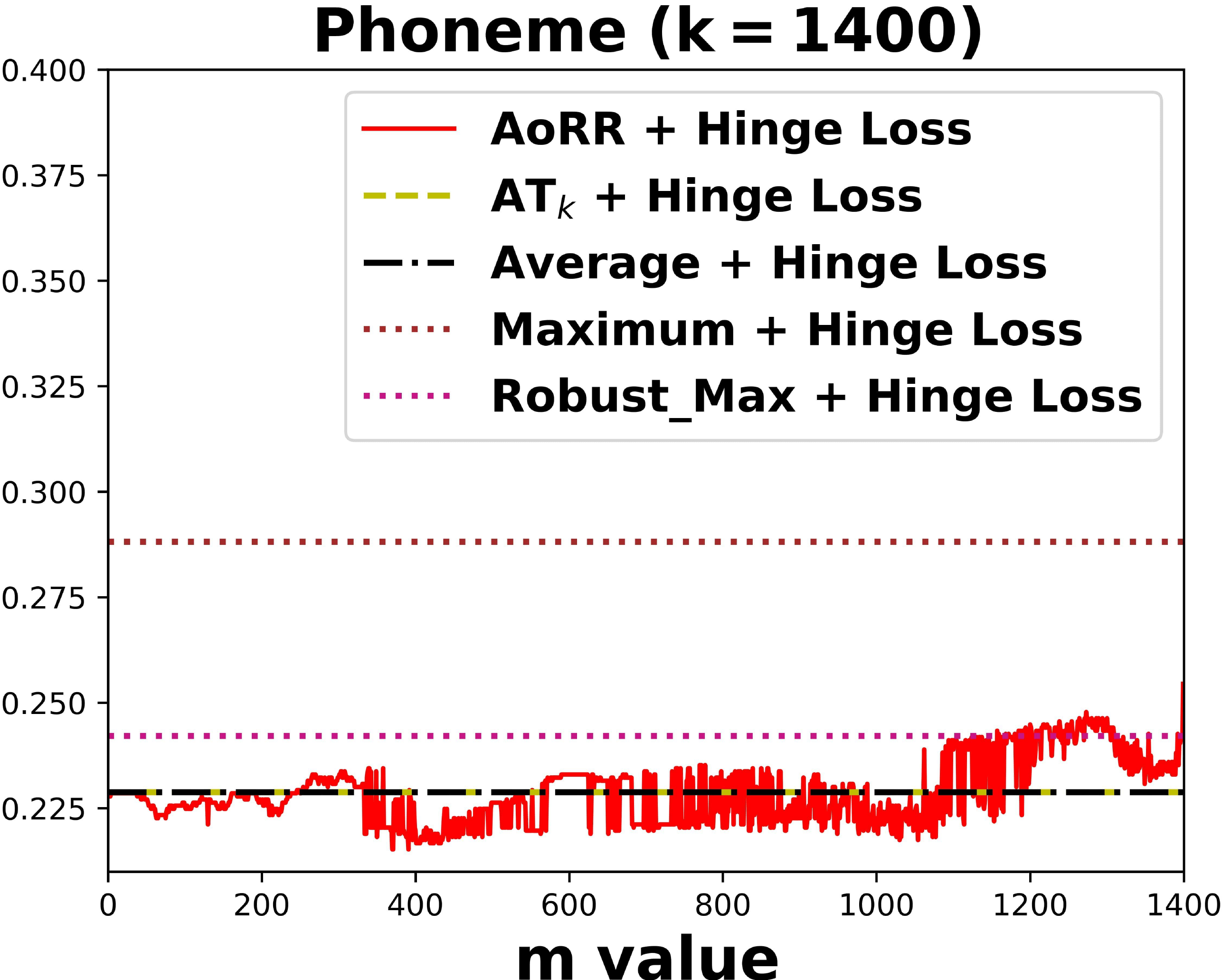}
        \end{subfigure}
        \caption{\small \em Tendency curves of error rate of learning \aorr~loss w.r.t. $m$ on four datasets.}\label{fig: aggregate_k_prime_value_hinge}
\end{figure*}

In this section, we use individual hinge loss as an example and plot tendency curves of the error rate w.r.t $m$ in Fig.\ref{fig: aggregate_k_prime_value_hinge} on 4 real-world datasets. From this figure, we get similar results as we discussed before.

\subsection{Performance of Additional Evaluation Metric on \tkml}
\label{sec:Additional_Evaluation_Metric}
We also adopt a widely used multi-label learning metric named average precision (AP) for performance evaluation. It is calculated as \cite{zhang2013review}

\[
AP=\frac{1}{n}\sum_{i=1}^n\frac{1}{|Y_i|}\sum_{j\in Y_i}\frac{|\{\tau \in Y_i| rank_f(x_i,\tau)<rank_f(x_i,j)\}|}{rank_f(x_i,j)}
\]

where $rank_f(x_i,j)$ returns the rank of $f_j(x_i)$ in descending according to $\{f_a(x_i)\}_{a=1}^l$.

From Table \ref{tab:ap_results}, we can find our \tkml~method outperforms the other two baseline approaches on all datasets. For the Emotions dataset, the AP score of \tkml~is 2.16\% higher than the LSEP method and near 10\% higher than the LR. The performance is also slightly improved on Scene and Yeast datasets.  These results demonstrate the effectiveness of our \tkml~method. 

\begin{table}[h]
\centering
\begin{tabular}{|c|c|c|c|}
\hline
\diagbox{Methods}{Datasets}& Emotions & Scene & Yeast \\ \hline
LR & 74.85 & 71.6 & 73.56 \\ \hline
LSEP & 82.66 & 85.43 & 74.26 \\ \hline
\tkml & \textbf{84.82} & \textbf{86.38} & \textbf{74.32} \\ \hline
\end{tabular}
\vspace{1mm}
\caption{\small \em AP (\%) results on three datasets. The best performance is shown in bold.}
\label{tab:ap_results}
\end{table}

\subsection{Performance on Each Class of the MNIST for Effects of \tkml}
\label{additional_exp_kgmc}
\begin{figure*}[t]
\captionsetup[subfigure]{justification=centering}
\centering
        \begin{subfigure}[b]{0.33\textwidth}
                \includegraphics[width=\linewidth]{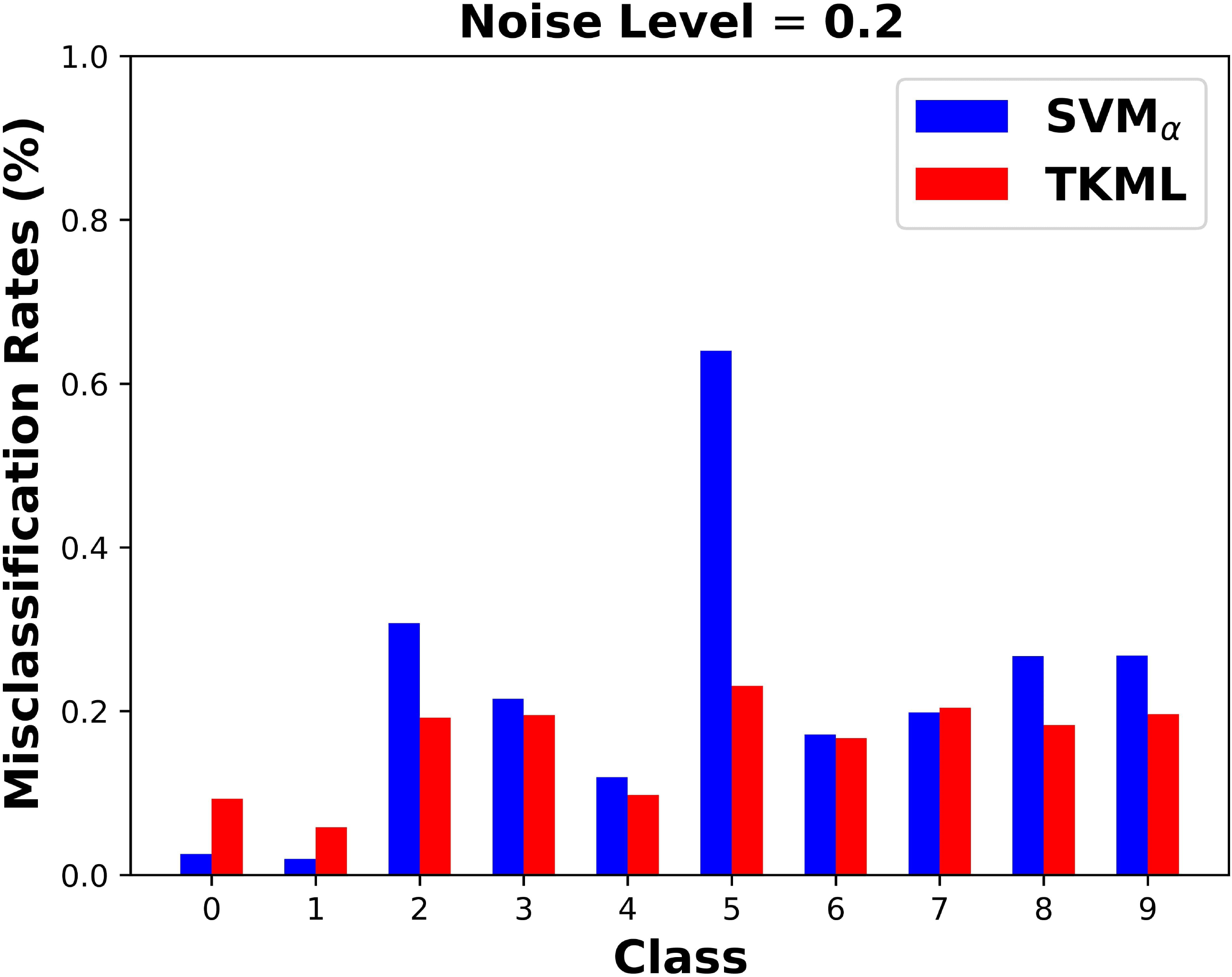}
                \caption{}
                \label{fig:mnist_noise_20}
        \end{subfigure}%
        \begin{subfigure}[b]{0.33\textwidth}
                \includegraphics[width=\linewidth]{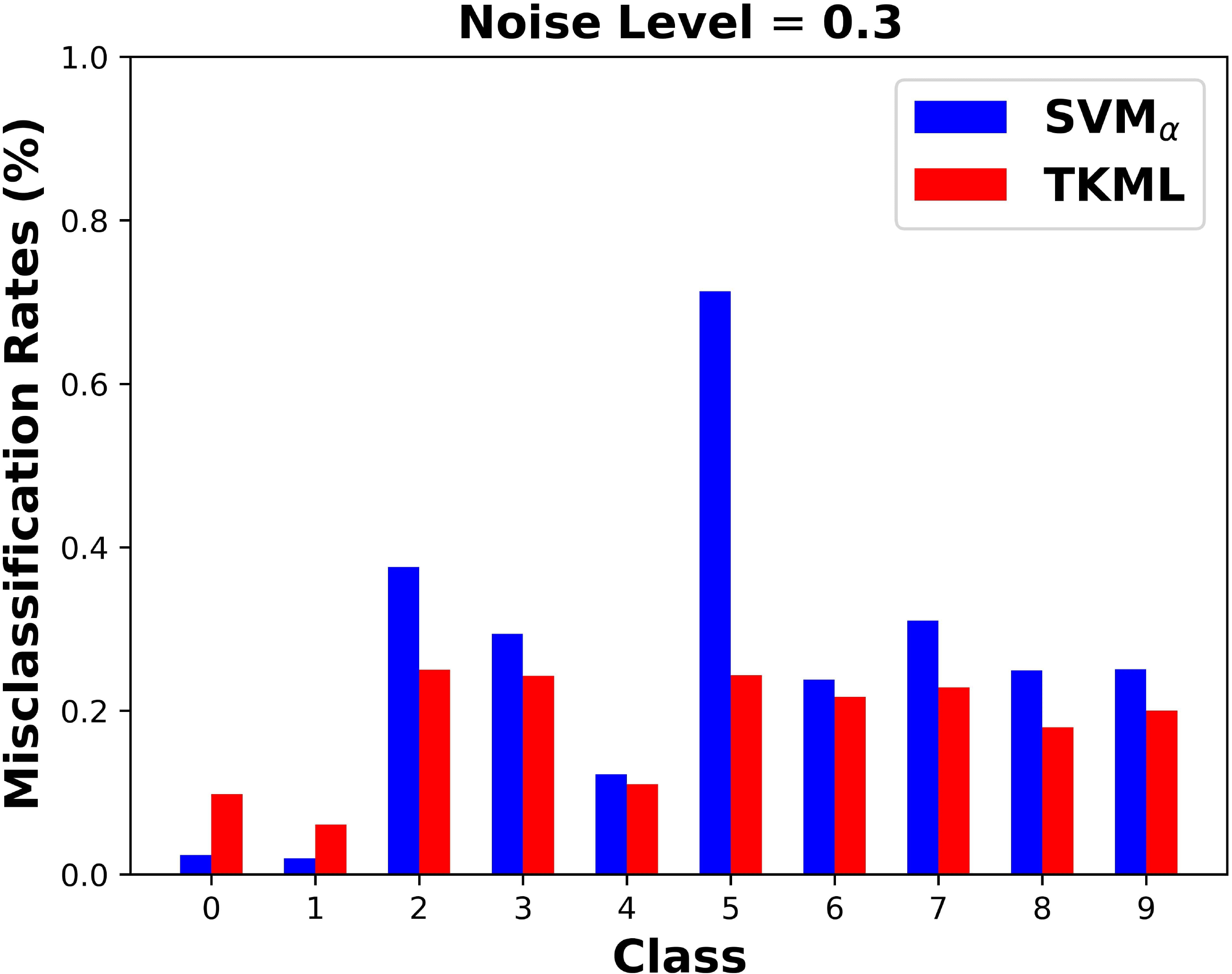}
                \caption{}
                \label{fig:mnist_noise_30}
        \end{subfigure}%
        \begin{subfigure}[b]{0.33\textwidth}
                \includegraphics[width=\linewidth]{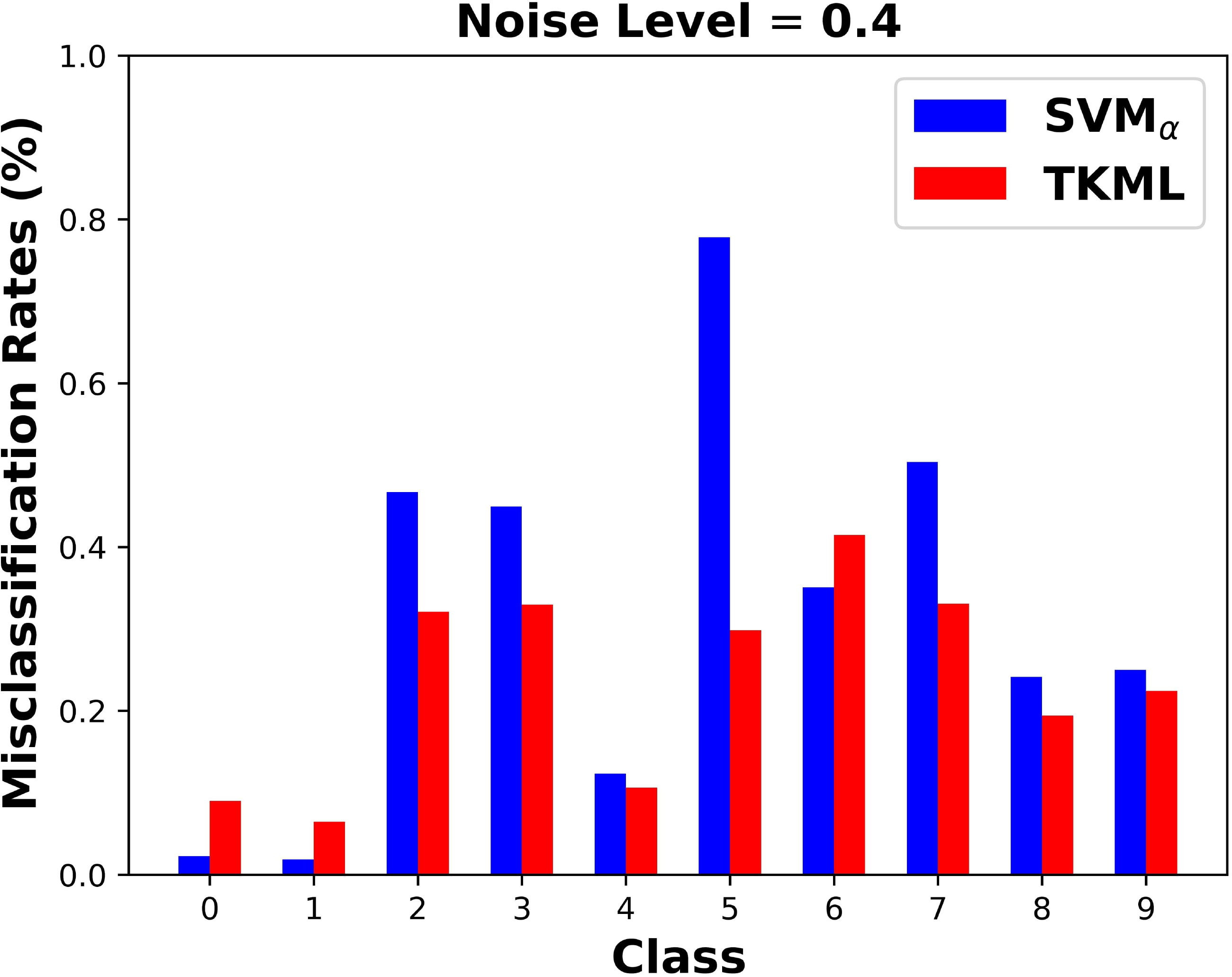}
                \caption{}
                \label{fig:mnist_noise_40}
        \end{subfigure}%
        \caption{\small \em The class-wise error rates of two methods with different noise level data.}\label{fig: MSVM_missclassification_error_by_class}
\end{figure*}

\textbf{Performance on each class.} To evaluate our method is better than SVM$_\alpha$ on the noisy data, we plot the class-wise error rate w.r.t different noise level data. As seen in Figure \ref{fig: MSVM_missclassification_error_by_class}, our method \tkml~outperforms SVM$_\alpha$ on the flipping classes such as 2 and 3, especially in class 5. As the noise level increases, the performance gap becomes more pronounced. For flipping class 7, the performance in this class is increased when the noise level increases from 0.3 to 0.4. The flipping class 6 also get good performance on the noise level 0.2 and 0.3.

\end{document}